\ifdefined\pdfsuppressptexinfo\pdfsuppressptexinfo=15\relax\fi
\ifdefined\pdfinfoomitdate\pdfinfoomitdate=1\relax\fi
\ifdefined\pdftrailerid\pdftrailerid{}\fi
\documentclass[11pt]{article}
\usepackage[T1]{fontenc}
\usepackage[utf8]{inputenc}

\usepackage[british]{babel}
\usepackage{manuscript}
\usepackage{colortbl,pifont,tabularx}

\usepackage{array}
\usepackage{needspace,wrapfig}
\newcommand{\finishwrap}{%
  \par
  \ifnum\value{WF@wrappedlines}>1
    \vspace{\dimexpr\value{WF@wrappedlines}\baselineskip-\baselineskip\relax}%
  \fi
  \WFclear
}
\usepackage[section]{placeins}
\usepackage{CJKutf8}
\graphicspath{{figures/}{cards/assets/}}

\newcommand{\greedyset}[1]{\ensuremath{J@#1}}


\definecolor{cardblue}{HTML}{2F6680}
\definecolor{cardblueback}{HTML}{EFF5F8}
\definecolor{rankbest}{HTML}{D9EFD3}
\definecolor{rankweak}{HTML}{FCE4B2}
\definecolor{factmark}{HTML}{CFDFEE}
\colorlet{cardlc}{black}
\tcbset{evidencebox/.style={
  enhanced,
  boxrule=0.6pt, arc=1.5pt,
  fonttitle=\bfseries\small, coltitle=white,
  left=5pt, right=5pt, top=4pt, bottom=4pt,
  before skip=4pt, after skip=4pt,
}}
\newtcolorbox[auto counter]{examplecard}[1]{%
  evidencebox, colframe=cardblue, colback=cardblueback,
  colbacktitle=cardblue,
  before upper={\colorlet{cardlc}{cardblue}},
  title={Example~\thetcbcounter\quad #1},
}

\makeatletter
\newcommand{\exhibitlabel}[1]{%
  \protected@edef\@currentlabel{\thetcbcounter}%
  \label{#1}}
\newcommand{\examplepartlabel}[2]{%
  \phantomsection
  \protected@edef\@currentlabel{\arabic{tcb@cnt@examplecard}#1}%
  \label{#2}}
\makeatother

\newcommand{\cardcue}[1]{%
  \smash{{\setlength{\fboxsep}{1pt}\colorbox{factmark}{\strut\textbf{#1}}}}}

\newsavebox{\papertitlebox}
\title{{\fontsize{17.28}{22}\selectfont Probing the ``Conscious'' Mind\thanks{Here, we use ``conscious'' to describe functional properties associated with conscious access in global workspace theory. We do not claim that these models possess consciousness or subjective experience.}:}\\[-0.25em]
\sbox{\papertitlebox}{\fontsize{14}{17}\selectfont Selective Access and Global Workspace in Vision-Language Models}%
\ifdim\wd\papertitlebox>\linewidth
\resizebox{\linewidth}{!}{\usebox{\papertitlebox}}%
\else\usebox{\papertitlebox}\fi}

\author{Jiahao Huang\,$^{1}$\quad Jun Wang\,$^{2\dagger}$}
\manuscriptaffiliations{%
  \mbox{$^{1}$ Imperial College London, UK\quad $^{2}$ University College London, UK}\\
  $^\dagger$ Corresponding author.
}

\hypersetup{hidelinks}

\DeclareCaptionFont{frontcaption}{\fontsize{9}{10.5}\selectfont}
\begin{document}

\newgeometry{left=1in,right=1in,top=0.60in,bottom=1in}

\begin{manuscriptfrontmatter}
Vision-language models (VLMs) integrate visual and textual information to perform a range of multimodal tasks, but their outputs do not reveal the models' internal concept representations.
In brain science, global workspace theory (GWT) proposes that information is selected from multiple parallel brain processes, integrated in a shared workspace, and made available to different specialised processes.
In this work, we investigate multimodal concept processing in VLMs in relation to these global workspace functions.
We introduce the Multimodal Concept Lens (MC-Lens), a multimodal Jacobian lens that combines vision encoder and language decoder lenses.
Both lenses map intermediate states into a shared semantic space defined by the decoder's output vocabulary, allowing us to compare concept accessibility across visual and language processing.
MC-Lens also supports internal concept interventions to examine the role of accessed concepts in downstream computation.
Our experiments address four related aspects.
\textbf{(i)~Accessibility and selective access.} Multimodal concept information is readable in intermediate states and can be selectively accessed according to task demands.
We extend the ``readout carriers'' from a single name token to a set of tokens, to better capture and express concept information and support subsequent counterfactual experiments.
\textbf{(ii)~Concepts in downstream computation.} Counterfactual experiments show that the concept information we access is not only readable, but also participates in downstream computation and reasoning.
\textbf{(iii)~Development of cross-modal alignment and selective access.} We develop two J-space based metrics to track cross-modal concept alignment.
In controlled VLM training, these metrics reflect its development more clearly than hidden state cosine.
The experiment also shows that reliable selective access develops later in training than concept accessibility.
\textbf{(iv)~Shared functions across models.} Across VLM families and sizes, we find shared patterns of concept accessibility, query-dependent selectivity and cross-task intervention effects, despite differences in internal organisation.
Together, these findings provide an experimental basis for studying GWT in multimodal artificial systems, and open a way to investigate how GWT can guide improvements in cross-modal alignment.

\end{manuscriptfrontmatter}

\begingroup
\setlength{\intextsep}{9pt}
\captionsetup{font=frontcaption,skip=4pt}
\begin{figure}[H]
    \centering
    \includegraphics[width=0.9\linewidth]{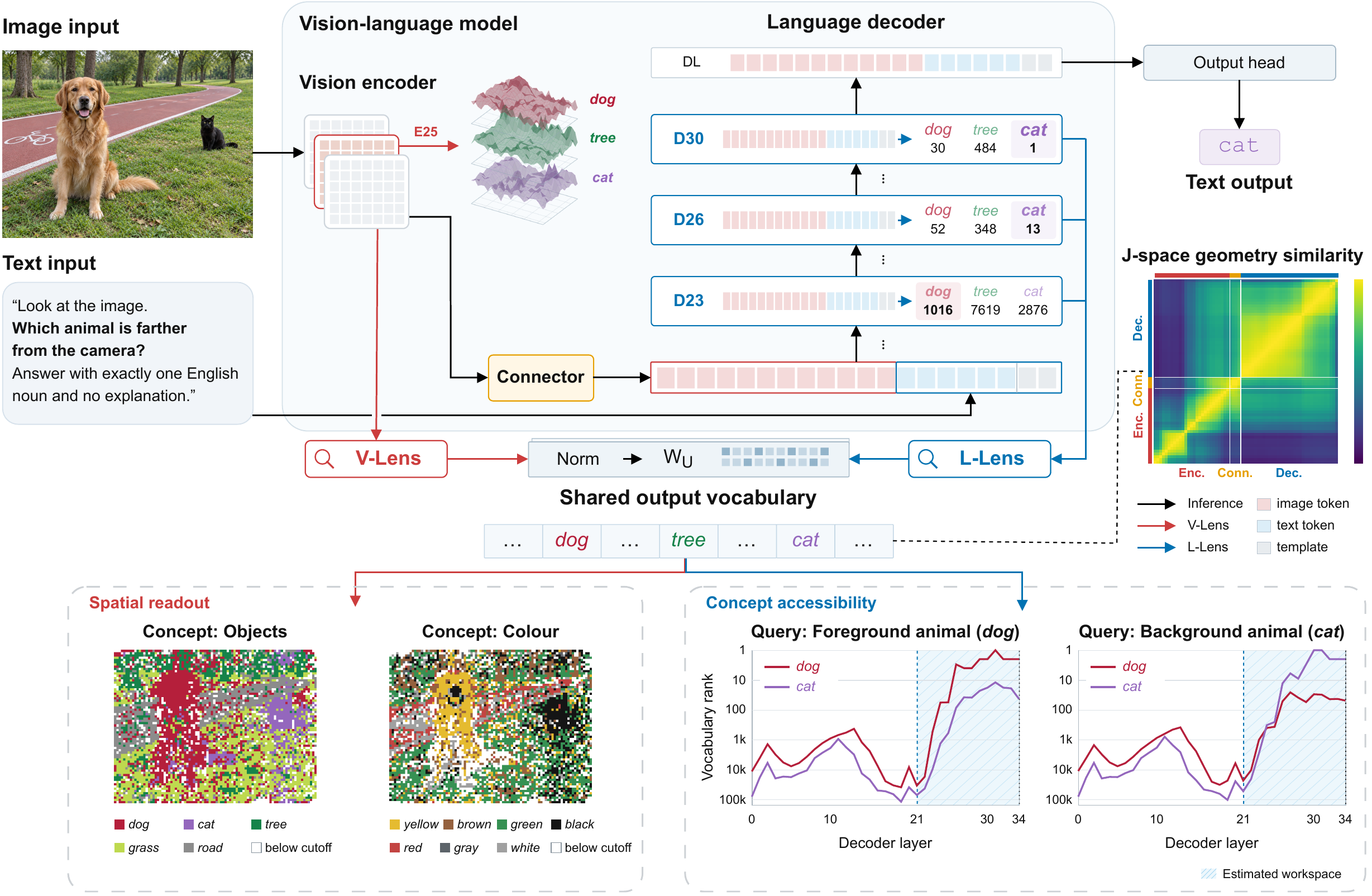}
    \caption{\textbf{Our proposed Multimodal Concept Lens (MC-Lens).}
    The vision encoder lens (V-Lens) reads object and colour concepts across image patches, while the language decoder lens (L-Lens) tracks how the question changes concept accessibility in the decoder.
    Both use a shared output vocabulary.
    Decoder numbers are vocabulary ranks, with smaller values indicating greater concept accessibility.}
    \label{fig:interface-overview}
\end{figure}
\endgroup

\restoregeometry
\section{Introduction}
\label{sec:introduction}

Vision-language models (VLMs) can effectively integrate visual and textual information to perform a range of multimodal tasks~\citep{alayrac2022flamingo,li2023blip2,liu2023visual,bai2025qwen3vl}.
For example, we can ask a VLM to identify an object in an image or describe its colour, and these questions require the model to extract different information from the same image and integrate it with the textual query.
However, although current VLMs can readily answer these questions, their outputs do not reveal their internal concept representations.
A VLM can also answer a visual question incorrectly even when the visual information needed to answer it remains readable from its intermediate states~\citep{fu2025hidden}.

Existing interpretability methods aim to identify information in internal states and examine its role in producing model outputs~\citep{alain2016understanding,neo2025visualprocessing,kaduri2025image}.
\citet{gurnee2026verbalizable} proposed the Jacobian Lens (J-Lens), a Jacobian-based method that provides vocabulary-based readouts of intermediate language-model states.
Specifically, it maps the model's intermediate hidden states into the space of final-layer decoder states via averaged Jacobian matrices, and provides concept directions for counterfactual intervention experiments.
Using J-Lens, \citet{gurnee2026verbalizable} identify functional and structural properties associated with a global workspace in large language models (LLMs).

In brain science, global workspace theory (GWT)~\citep{baars2005workspace,baars2007architectural,dehaene1998neuronal} is an influential theory of consciousness that explains how information from different sources is integrated to support flexible cognition.
The core hypothesis of this theory is that information is selected from the multiple parallel processes in the brain, integrated in a shared workspace, and made available to different specialised processes.
Building on the J-Lens study, we extend this analysis to concept information expressed through different modalities, and processed across the visual and language components of VLMs.

In this paper, we introduce the Multimodal Concept Lens (MC-Lens), a multimodal Jacobian lens that combines a vision encoder lens (V-Lens) and a language decoder lens (L-Lens).
V-Lens can access concept information from intermediate vision encoder states before visual information is combined with the text input in the decoder.
L-Lens can access concept information from decoder states during the integration of visual and textual inputs, allowing us to examine selective access under different text queries and during autoregressive generation.
Both V-Lens and L-Lens map intermediate states into a shared semantic space defined by the decoder's output vocabulary, so we can compare concept accessibility during multimodal information integration and processing.
Figure~\ref{fig:interface-overview} illustrates these readouts in a visual question answering (VQA) example.
We conduct counterfactual experiments using MC-Lens, where we intervene on intermediate state components associated with the accessed concepts and observe the resulting model outputs, to infer the role of these concepts in downstream computation.
In addition, we also investigate the functional properties and internal organisation of multimodal concept processing in VLMs in relation to GWT.
Specifically, our experiments address four related aspects.

\textbf{(i)~Accessibility and selective access.}
Our experiments show that multimodal concept information is readable in intermediate states and can be selectively accessed according to task demands.
Spatial readouts further reveal object identity and attributes such as colour and material at a fine spatial scale.
In addition, we extend the ``readout carriers'' used to express a concept, from a single name token to a set of tokens.
This extension helps us better capture and express concept information in intermediate states, and supports subsequent counterfactual experiments on its role in downstream computation.

\textbf{(ii)~Concepts in downstream computation.}
The counterfactual experiments indicate that the concept information we access is not merely shown on a ``display board'', but also participates in downstream computation and reasoning.
In causal reasoning VQA tasks, we perform counterfactual interventions on intermediate concept representations, and find that these interventions can redirect answers to questions requiring multiple steps of reasoning, or influence multiple aspects of downstream reasoning.
And we find that these effects generally become stronger as we expand the ``readout carriers'' using larger concept direction sets.

\textbf{(iii)~Development of cross-modal alignment and selective access.}
We develop two J-space based metrics to track correspondence between visual states and language-defined concepts.
We also construct a controlled VLM training setting for vision-language alignment, and evaluate these metrics across checkpoints at different stages of training.
Our experiments show that our proposed metrics reflect the development of cross-modal concept alignment more clearly than the hidden state cosine baseline.
Concept readouts from the same checkpoints further show that concept accessibility improves early in alignment training, whereas reliable selective access develops later.

\textbf{(iv)~Shared functions across models.}
We extend our experiments to VLMs of different sizes from multiple model families~\citep{bai2025qwen3vl,wu2024deepseekvl2,gemmateam2025gemma3}.
Across these models, we find shared patterns of concept accessibility, query-dependent selectivity and cross-task intervention effects, despite differences in their internal workspace organisation.

Together, these findings provide an experimental basis for studying GWT in multimodal artificial systems, and open a way to investigate how GWT can guide improvements in cross-modal alignment.

\section{MC-Lens: A Jacobian Lens for Multimodal Concepts}
\label{sec:mc-lens}

In general, a VLM's output provides only a limited view of its internal processing.
For example, a VLM may correctly identify a bicycle in a photograph, but the answer itself does not tell us where the concept of \emph{bicycle} appears inside the model.

Therefore, we develop MC-Lens, a Jacobian-based method inspired by J-Lens~\citep{gurnee2026verbalizable}, to access concept information from intermediate states in the VLM and to investigate the role of the accessed concepts in downstream computation and reasoning.
In this section, we focus on the mathematical formulation of MC-Lens.

\subsection{Multimodal Jacobian Lenses}
\label{sec:multimodal-lenses}

In this part, we introduce MC-Lens, a multimodal Jacobian lens that combines V-Lens and L-Lens.
V-Lens and L-Lens both apply an averaged Jacobian matrix to measure the effects of intermediate hidden states on the final-layer decoder hidden states in the VLM.
V-Lens measures the effect of an intermediate vision encoder state on the final-layer decoder hidden state, while L-Lens measures the corresponding effect of an intermediate language decoder state.

For a multimodal input \(x\), let \(h_{\ell,p}^{m}(x)\in\mathbb R^{d^m}\) denote the intermediate state at layer \(\ell\) and position \(p\), where \(m\in\{E,D,C\}\) marks the vision encoder, the language decoder and the connector, respectively.
Here, \(p\) indexes positions in the corresponding patch or token sequence.
The connector maps the encoder patches into decoder tokens.
The final state \(h_{L,q}^{D}(x)\in\mathbb R^{d^D}\) is the decoder state after the final layer \(L\) at position \(q\), before the final normalisation.
We take the partial derivatives of the final state \(h_{L,q}^{D}(x)\) with respect to the intermediate state \(h_{\ell,p}^{m}(x)\) to measure the effect of a small change \(\delta h_{\ell,p}^{m}\) on the final state.
These partial derivatives form the Jacobian matrix \(J_{\ell,p\rightarrow q}^{m}(x) \in\mathbb R^{d^D\times d^m}\) and give a first-order Taylor approximation of the effect:
\begin{align}
	J_{\ell,p\rightarrow q}^{m}(x)
	&=
	\frac{\partial h_{L,q}^{D}(x)}
	{\partial h_{\ell,p}^{m}(x)},
	\qquad m\in\{E,D,C\},
	\label{eq:local-jacobian}\\
	\delta h_{L,q}^{D}
	&\approx
	J_{\ell,p\rightarrow q}^{m}(x)\,\delta h_{\ell,p}^{m}.
	\label{eq:local-jacobian-action}
\end{align}

\Needspace{22\baselineskip}
\begingroup
\setlength{\columnsep}{0.02\textwidth}
\setlength{\intextsep}{0pt}
\begin{wrapfigure}{r}{0.49\textwidth}
\captionsetup{type=figure,font=small,justification=raggedright,singlelinecheck=false,skip=0pt}
\centering
\includegraphics[width=\linewidth]{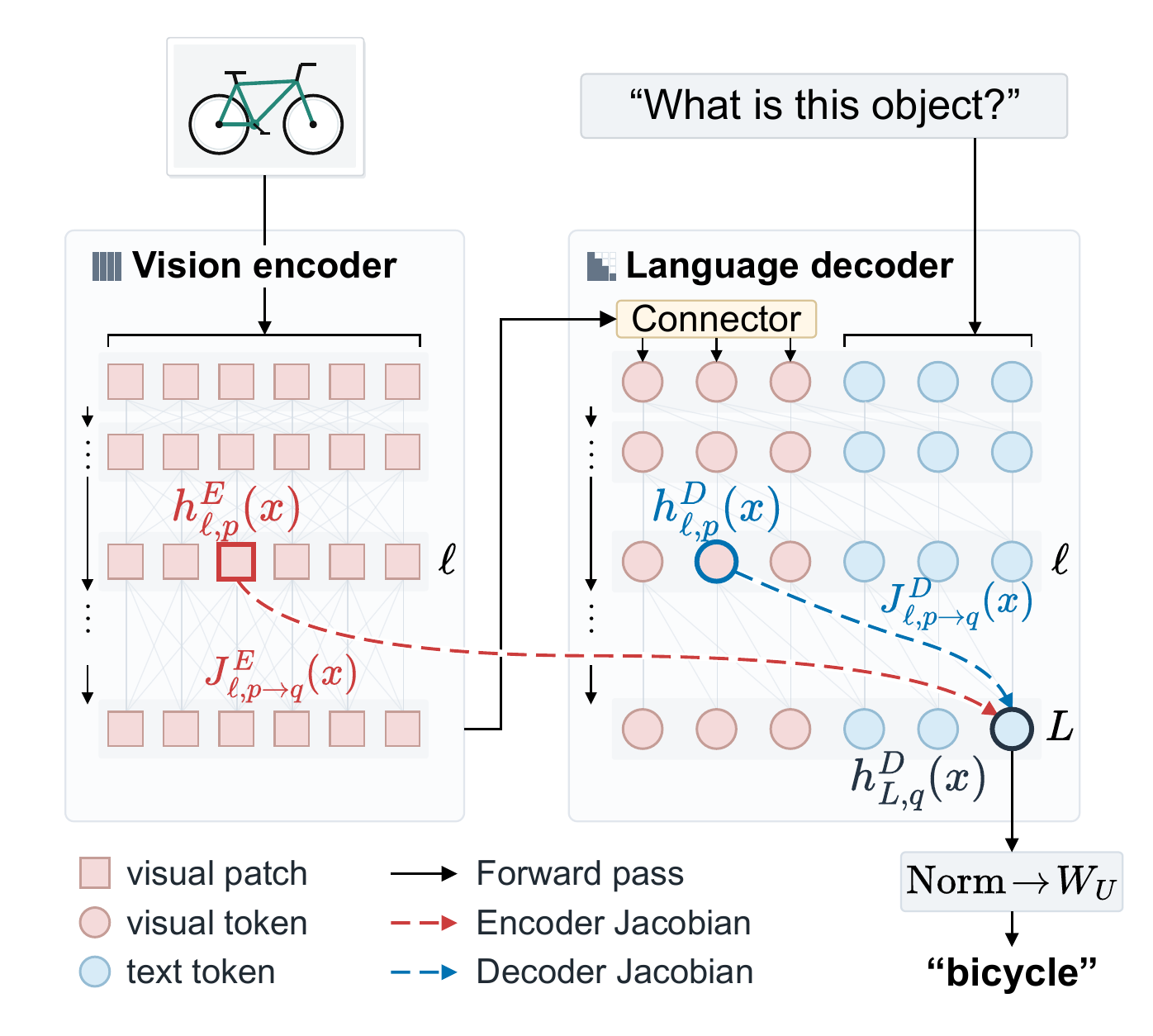}
\caption{\textbf{Local Jacobians in a VLM.} Squares and circles denote hidden states at patch or token positions. Red and blue dashed arrows show local Jacobians from encoder and decoder states at \(p\) to the same final-layer decoder state at \(q\).}
\label{fig:local-jacobians-shared-vlm}
\end{wrapfigure}
Figure~\ref{fig:local-jacobians-shared-vlm} illustrates these local Jacobians for selected encoder and decoder states.

However, such a Jacobian matrix mixes two things: the particular use of the intermediate state in the current context, and the general effect it has on later outputs across contexts.
J-Lens~\citep{gurnee2026verbalizable} isolates the general effect by averaging the Jacobian matrices within and across inputs.
Specifically, within each input, it sums the Jacobian matrices over the final-layer positions \(q\) and averages over the intermediate positions \(p\), and it then averages across a corpus of text prompts.
In this way, for a given model, we obtain one averaged Jacobian matrix per layer, which serves as the lens for that layer.

This uniform averaging strategy in J-Lens~\citep{gurnee2026verbalizable} suits a language model, since all input positions belong to the same category of text tokens.
However, in VLMs, image tokens often greatly outnumber text tokens~\citep{yang2025visionzip}, while text tokens carry the question and other task information and can disproportionately influence the final output~\citep{chen2024fastv,deng2025words}.
The uniform averaging strategy over the whole sequence would therefore give image positions a larger total weight than text positions in the lens estimate.
In addition, in a VLM, the vision encoder processes a sequence of image patches, and the language decoder processes a mixed sequence of image, text and template tokens, so the two lenses need different position sets.
We therefore present a unified formulation for V-Lens and L-Lens, and then specify the settings for each lens.
\finishwrap
\endgroup

We estimate V-Lens and L-Lens from \(N\) VQA examples \(x_i\), each consisting of an image, a question and the corresponding answer.
For each VQA example \(i\), \(\mathcal P_i\) is the set of positions \(p\) of the intermediate states, and \(\mathcal Q_i\) is the set of positions \(q\) of the final-layer decoder states in Equation~\ref{eq:local-jacobian}.

The intermediate position set \(\mathcal P_i\) depends on the lens: for V-Lens it contains the encoder patch positions, and for L-Lens the decoder token positions.
We further group \(\mathcal P_i\) into categories \(\mathcal C_{i,c}\), indexed by a category set \(\mathcal K\).
For V-Lens, \(\mathcal K\) has a single category containing all patches, and for L-Lens it has three categories: image, text and template tokens.
We average within each category and then across categories.
This gives each category equal total weight, regardless of how many tokens it contains.

The final-layer position set \(\mathcal Q_i\) is the same for both lenses and contains only the text positions in the decoder.
We exclude the decoder image positions from \(\mathcal Q_i\) for two reasons:
1) The model produces no output at these image token positions, and we do not study the effects of intermediate states on these image token states.
2) Image tokens are numerous, and their terms, if included, would dominate the sum and hide the effects on the text positions that we care about.
Appendix~\ref{app:lens-fitting} lists further excluded positions.

\Needspace{10\baselineskip}
To sum up, we formulate the estimation of both lenses in one unified form:
\begin{align}
	J_{\ell,p}^{m}(x_i)
	&=
	\sum_{q\in\mathcal Q_i}
	J_{\ell,p\rightarrow q}^{m}(x_i),
	\qquad m\in\{E,D,C\},
	\label{eq:position-jacobian}\\\displaybreak[1]
	J_\ell^{m}(x_i)
	&=
	\frac{1}{|\mathcal K|}
	\sum_{c\in\mathcal K}
	\frac{1}{|\mathcal C_{i,c}|}
	\sum_{p\in\mathcal C_{i,c}}
	J_{\ell,p}^{m}(x_i),
	\qquad
	\mathcal P_i=\bigcup_{c\in\mathcal K}\mathcal C_{i,c},
	\label{eq:example-jacobian}\\
	J_\ell^{m}
	&=
	\frac{1}{N}
	\sum_{i=1}^{N}
	J_\ell^{m}(x_i),
	\label{eq:multimodal-jacobian}
\end{align}

\noindent
where Equation~\ref{eq:position-jacobian} collects the effect of position \(p\) on all text positions, Equation~\ref{eq:example-jacobian} averages these effects within an example with the category balance described above, and Equation~\ref{eq:multimodal-jacobian} averages over examples.
The lens \(J_\ell^{m}\in\mathbb R^{d^D\times d^m}\) is fixed after estimation and can be applied across inputs and positions.

\paragraph{L-Lens.}
\phantomsection
\begingroup
\makeatletter
\edef\@currentlabel{\thesubsection}
\label{sec:decoder-lens}
\makeatother
\endgroup
For L-Lens (\(m=D\)), we group the intermediate position set \(\mathcal P_i\) of the decoder into image tokens, text tokens and template tokens.
Image tokens originate from the connector outputs of the encoder patches, text tokens cover the question and the answer, and template tokens are the formatting and control tokens of the prompt template.

Since the decoder uses causal attention, the final state at position \(q\) does not depend on positions \(p>q\), so the Jacobian matrix is zero for \(q<p\), and the sum in Equation~\ref{eq:position-jacobian} runs over \(q\geq p\).
With these settings, Equations~\ref{eq:position-jacobian}--\ref{eq:multimodal-jacobian} combine into
\begin{equation}\label{eq:decoder-jacobian}
	J_\ell^{D}
	=
	\frac{1}{N}
	\sum_{i=1}^{N}
	\frac{1}{|\mathcal K|}
	\sum_{c\in\mathcal K}
	\frac{1}{|\mathcal C_{i,c}|}
	\sum_{p\in\mathcal C_{i,c}}
	\sum_{\substack{q\in\mathcal Q_i\\q\geq p}}
	\frac{\partial h_{L,q}^{D}(x_i)}
	     {\partial h_{\ell,p}^{D}(x_i)},
	\qquad
	\mathcal K=\{\mathrm{img},\mathrm{txt},\mathrm{tpl}\}.
\end{equation}
Compared with J-Lens~\citep{gurnee2026verbalizable}, L-Lens balances the three token categories over the intermediate positions \(p\), and restricts the final-layer positions \(q\) to the text positions.

\paragraph{V-Lens.}
\phantomsection
\begingroup
\makeatletter
\edef\@currentlabel{\thesubsection}
\label{sec:vision-lens}
\makeatother
\endgroup
For V-Lens (\(m=E\)), the Jacobian matrix is computed by backpropagation through the connector and the whole decoder.
The intermediate position set \(\mathcal P_i\) contains the encoder patch positions, which form a single category, \(\mathcal K=\{\mathrm{patch}\}\) with \(\mathcal C_{i,\mathrm{patch}}=\mathcal P_i\), so the category average in Equation~\ref{eq:example-jacobian} reduces to a plain average over all patches.
The final-layer position set \(\mathcal Q_i\) is the same text position set as for L-Lens.
We do not apply the causal condition \(q\geq p\), since encoder patches and decoder tokens belong to different sequences.
With these settings, Equations~\ref{eq:position-jacobian}--\ref{eq:multimodal-jacobian} combine into
\begin{equation}\label{eq:vision-jacobian}
	J_\ell^{E}
	=
	\frac{1}{N}
	\sum_{i=1}^{N}
	\frac{1}{|\mathcal P_i|}
	\sum_{p\in\mathcal P_i}
	\sum_{q\in\mathcal Q_i}
	\frac{\partial h_{L,q}^{D}(x_i)}
	     {\partial h_{\ell,p}^{E}(x_i)}.
\end{equation}
The same construction applies to intermediate connector states with \(m=C\), since the connector processes only the encoder patches (Appendix~\ref{app:lens-fitting}).

\subsection{Vocabulary-Based Concept Readout}
\label{sec:vocabulary-readout}

Suppose a VQA task asks the model to identify the object in an input image of a bicycle.
The model can easily answer ``bicycle''.
However, this answer alone does not tell us where the concept of \emph{bicycle} appears inside the model.
Here, we want to figure out at which layer this concept already appears in the model's internal states before the answer is produced.

\begin{figure}[tbp]
  \centering
  \includegraphics[width=\linewidth]{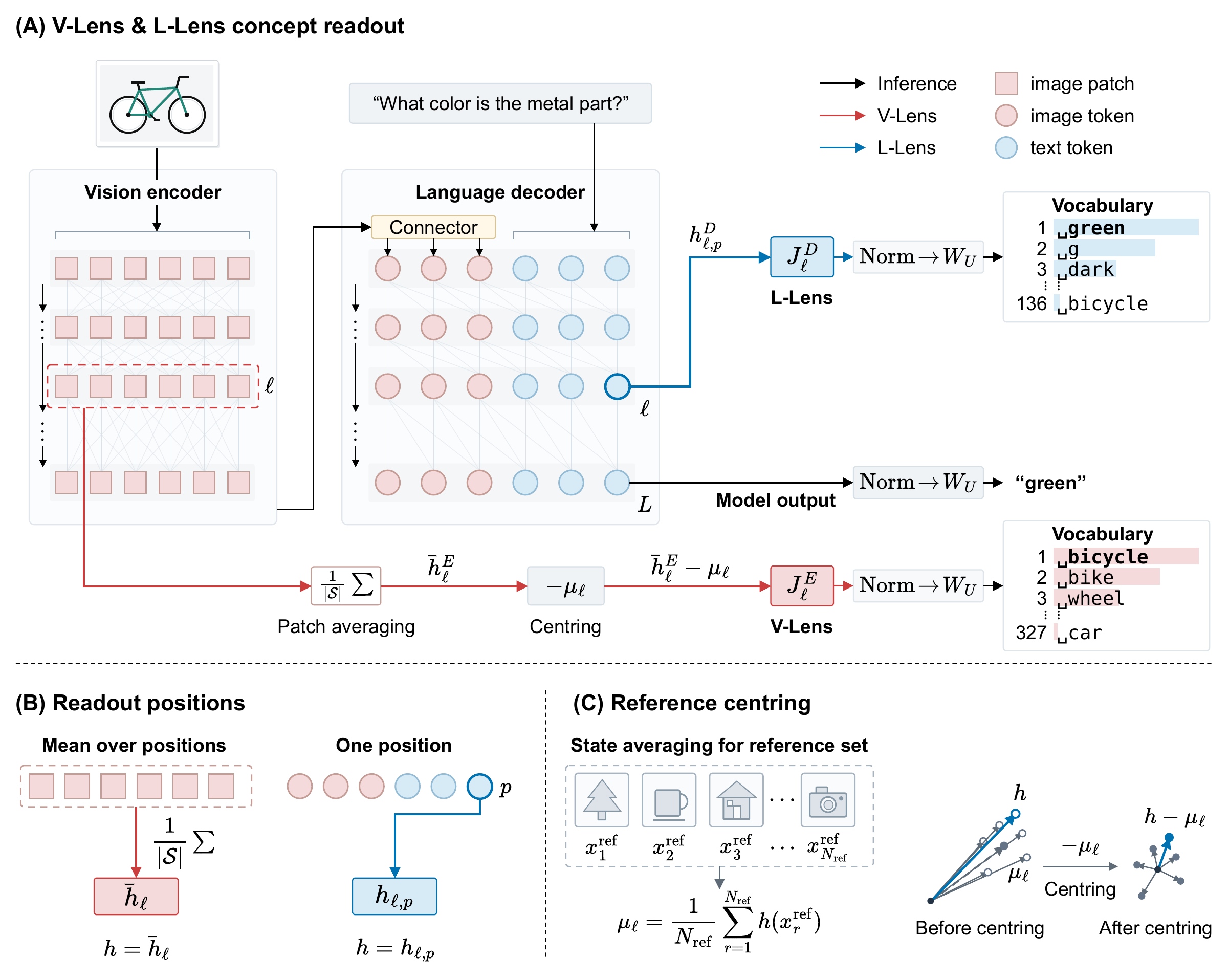}
\caption[Concept readout and reference centring.]{\textbf{Concept readout and reference centring.}
\textbf{(A)} V-Lens and L-Lens map intermediate states into the decoder's final-layer state space, then use its output head to rank vocabulary tokens.
\textbf{(B)} The readout state is taken from one position or averaged over a position set \(\mathcal S\).
\textbf{(C)} Reference centring subtracts \(\mu_\ell\), the mean state over reference inputs.}
\label{fig:vocabulary-readout-schematic}
\end{figure}

In this part, we use MC-Lens to read an intermediate state by projecting the state onto the decoder's output vocabulary (Figure~\ref{fig:vocabulary-readout-schematic}~(A)).
The readout gives a logit for every word in the vocabulary.
We then rank all words by these logits, and the ranks tell us which words the state is close to.

In this VQA example of identifying a bicycle, if an intermediate state ranks the word ``bicycle'' near the top, we regard the concept as readable in that state, and we use the rank of the concept name to measure how accessible the concept is.
The concept readout here corresponds to one token in the vocabulary, so we call such a concept a single-token concept.\footnote{A word can be tokenised in different ways. In the Qwen3-VL tokeniser, for example, \texttt{bicycle} without a leading space is split into the sub-word tokens \texttt{b} and \texttt{icycle}, while \texttt{\char32 bicycle} with its leading space is one token. Here we first discuss the latter case, where one token represents a concept.}

\begingroup
\clubpenalty=10000
We write \(h_{\ell,p}(x)\in\mathbb R^{d}\) for the intermediate state at layer \(\ell\) and position \(p\), and \(J_\ell\) for the estimated lens at layer \(\ell\).{\interfootnotelinepenalty=10000\footnote{These are the \(h_{\ell,p}^{m}(x)\), \(J_\ell^{m}\) and \(d^m\) of Section~\ref{sec:multimodal-lenses}. We omit the superscript \(m\) when the network is clear from context.}}
Specifically, we view the intermediate state to be read, \(h\in\mathbb R^{d}\), as a vector from the origin, and use the estimated lens \(J_\ell\) to map its direction into the final-layer decoder state space.
The resulting vector \(J_\ell h\) has the same form as a final-layer decoder hidden state, so we apply the decoder's final normalisation \(\operatorname{Norm}\) and unembedding matrix \(W_U\) to it, in the same way as the model computes its next-token logits.
Each row of the unembedding matrix \(W_U\) corresponds to one vocabulary token, \(W_U=[u_1\ \cdots\ u_w\ \cdots\ u_{|\mathcal V|}]^\top\in\mathbb R^{|\mathcal V|\times d^D}\), where \(|\mathcal V|\) is the vocabulary size and \(u_w\in\mathbb R^{d^D}\) is the transpose of row \(w\).
For state \(h\), we take the logit of token \(w\) as its score:
\begin{equation}\label{eq:vocabulary-readout}
	\operatorname{score}_\ell(w\mid h)
	=
	u_w^\top\operatorname{Norm}\!\left[J_\ell h\right],
\end{equation}
\endgroup
and softmax over the vocabulary gives the readout distribution:
\begin{equation}\label{eq:softmax-readout}
	\operatorname{lens}_\ell(h)
	=
	\operatorname{softmax}\!\left(W_U\operatorname{Norm}\!\left[J_\ell h\right]\right),
\end{equation}
which has the same form as the J-Lens readout~\citep{gurnee2026verbalizable}.

\paragraph{Readout positions.}
We use different position sets for V-Lens in the vision encoder and L-Lens in the language decoder to select the intermediate state \(h\).
In the vision encoder, V-Lens reads either the state of a single patch or the mean state of all image patches.
A single patch state tells us which concepts are readable at the corresponding spatial location, and the mean state over patches provides a readout for the whole image.

In the language decoder, L-Lens likewise reads either the state of a single token or the mean state over a set of tokens.
For a single state, we typically read the pre-answer state, the decoder state at the last token position before the answer.
This state tells us which concepts are readable after the model has combined the image with the question and right before it answers.
For a mean state, we average the decoder image-token states.
These tokens originate from the encoder patches through the connector, so their mean is the decoder counterpart of the mean patch state.

We write \(\bar h_\ell(x)\) for the mean state over a set \(\mathcal S\) of readout positions, so the state \(h\) in Equations~\ref{eq:vocabulary-readout} and~\ref{eq:softmax-readout} is either a single state or a mean state (Figure~\ref{fig:vocabulary-readout-schematic}~(B)):
\begin{equation}\label{eq:readout-state}
	h(x)=
	\begin{cases}
		h_{\ell,p}(x), & \text{a single position } p,\\[4pt]
		\bar h_\ell(x)=\dfrac{1}{|\mathcal S|}\displaystyle\sum_{p\in\mathcal S} h_{\ell,p}(x), & \text{the mean over the readout positions } \mathcal S.
	\end{cases}
\end{equation}

In the bicycle VQA example, reading a patch in the bicycle region via V-Lens tells us whether the concept \emph{bicycle} is readable at that spatial location, and reading the mean patch state via V-Lens tells us whether it is readable from the photograph as a whole.
Reading the pre-answer state via L-Lens tells us whether \emph{bicycle} is readable in the internal state after the model has collected the information from the image and the question, right before it answers.

\paragraph{Reference centring.}
The readout method described so far works well for the pre-answer state, but has a limitation for visual states (encoder patch states and decoder image-token states).
The same tokens tend to appear near the top for different images, and the name of the object in the image is buried below tokens that are uninformative about the object.

We assume that intermediate states contain a large component common to different images, and this component enters the scores in Equation~\ref{eq:vocabulary-readout} together with the image-specific part.
This assumption is also consistent with the findings of \citet{shukor2024implicit}, who observed that mean visual-token states from different images point in similar directions in the language decoder.
We therefore introduce reference centring (Figure~\ref{fig:vocabulary-readout-schematic}~(C)), which subtracts a reference state from \(h\) before the readout.
For the reference state, we compute the readout state in the same way as in Equation~\ref{eq:readout-state}, at the same layer, for each of \(N_{\mathrm{ref}}\) natural reference images \(x_r^{\mathrm{ref}}\) (ordinary photographs).
We then average these states:
\begin{equation}\label{eq:reference-means}
	\mu_\ell=
	\begin{cases}
		\dfrac{1}{N_{\mathrm{ref}}}\displaystyle\sum_{r=1}^{N_{\mathrm{ref}}} h(x_r^{\mathrm{ref}}), & \text{with reference centring},\\[10pt]
		0, & \text{without reference centring}.
	\end{cases}
\end{equation}
We subtract \(\mu_\ell\) from \(h\) to remove this common component and use \(\mu_\ell\) as a new origin for the readout, so the readout ranks tokens by how the state of this image differs from an ``average'' image.
Replacing \(h\) by \(h-\mu_\ell\) in Equations~\ref{eq:vocabulary-readout} and~\ref{eq:softmax-readout} gives the centred readout:
\begin{align}
	\operatorname{score}_\ell(w\mid h,\mu_\ell)
	&=
	u_w^\top\operatorname{Norm}\!\left[J_\ell\,(h-\mu_\ell)\right],
	\label{eq:centred-score}\\
	\operatorname{lens}_\ell(h,\mu_\ell)
	&=
	\operatorname{softmax}\!\left(W_U\operatorname{Norm}\!\left[J_\ell\,(h-\mu_\ell)\right]\right),
	\label{eq:centred-readout}
\end{align}
with \(h\) given by Equation~\ref{eq:readout-state} and \(\mu_\ell\) by Equation~\ref{eq:reference-means}.
Without centring, \(\mu_\ell=0\) and Equations~\ref{eq:centred-score} and~\ref{eq:centred-readout} reduce to Equations~\ref{eq:vocabulary-readout} and~\ref{eq:softmax-readout}.

In the bicycle VQA example, the pre-answer state read via L-Lens tends to rank \texttt{\char32 bicycle} near the top, while the mean patch state read via V-Lens and the mean image-token state read via L-Lens both tend to rank uninformative tokens near the top, such as \texttt{\char32 photo}, \texttt{\char32 image} and \texttt{**}, which appear for almost every image and say nothing about the bicycle.
After centring, the two visual readouts tend to rank \texttt{\char32 bicycle} much higher, and the uninformative tokens tend to drop out of the top ranks.

\subsection{Constructing Concept Direction Sets}
\label{sec:concept-direction-sets}

So far, we have used MC-Lens to access concept information from intermediate hidden states in a verbalised form.
However, we found that the hidden state usually carries more information about a concept than its name token expresses.
For example, in the bicycle VQA task, \texttt{\char32 bicycle} is not the only token that expresses the concept, for two reasons.
(1) Different word forms may express the same concept, such as \texttt{\char32 Bicycle}, \texttt{\char32 bike} and \texttt{\char32 bicycles}.
(2) Surrounding concepts can also express part of the meaning of this concept.
Some are concepts that are closely linked to a bicycle, such as \texttt{\char32 riders} and \texttt{\char32 road}, and some are parts of a bicycle or the category it belongs to, such as \texttt{\char32 handle} and \texttt{\char32 transport}.\footnote{Here we mean concepts associated with a single object, rather than other objects present in the image. Even when the image shows a bicycle against a blank background, the hidden state may still carry information associated with \texttt{\char32 riders}, \texttt{\char32 road} and \texttt{\char32 handle}.}

This phenomenon is especially pronounced in VLMs.
In an LLM, the word ``bicycle'' enters the hidden state as the token \texttt{\char32 bicycle} itself.
However, in a VLM, the concept enters from the image, which has no such token to anchor it, so we expect the concept to be spread over more vocabulary tokens in the hidden state.
The readout and counterfactual experiments in Sections~\ref{sec:readout-reconstruction} and~\ref{sec:counterfactual-interventions} support this view.

In this part, we therefore describe a concept with a set of tokens rather than one token.
J-Lens describes a single-token concept by the vocabulary direction of its token~\citep{gurnee2026verbalizable}.
Here, we select a set of tokens for each concept and use their vocabulary directions as its concept directions.
Together, these directions form the concept direction set.
We choose these vocabulary tokens in two ways: (1) word-form selection and (2) greedy selection via J-space reconstruction.
Below, we first review the definitions of vocabulary direction and J-space from J-Lens, and then describe the two selection methods.

To define vocabulary directions, we use the unnormalised score of token \(w\) for state \(h\):\footnote{The superscript \(*\) indicates that the decoder's final normalisation \(\operatorname{Norm}\) in Equation~\ref{eq:vocabulary-readout} is omitted.}
\begin{equation}\label{eq:score-inner-product}
	\operatorname{score}^{*}_\ell(w\mid h)
	=
	u_w^\top J_\ell h
	=
	(J_\ell^\top u_w)^\top h.
\end{equation}
Moving the state along \(J_\ell^\top u_w\) raises this score faster than moving along any other direction, so this vector can be regarded as the direction of token \(w\) in the intermediate state space.
Since we care about the direction of the vector rather than its length, we normalise the vector to unit length and call the resulting vector the vocabulary direction of \(w\), following J-Lens~\citep{gurnee2026verbalizable}.
We write the raw vector as \(\phi_{\ell,w}\) and the vocabulary direction as \(\bar\phi_{\ell,w}\):
\begin{equation}\label{eq:raw-token-direction}
	\phi_{\ell,w}=J_\ell^\top u_w\in\mathbb R^{d},
	\qquad
	\bar\phi_{\ell,w}
	=\frac{\phi_{\ell,w}}{\|\phi_{\ell,w}\|_2}\in\mathbb R^{d}.
\end{equation}

The J-space can then be constructed from the vocabulary directions.
Here, \(k\) limits the number of vocabulary directions used to represent a state.
The J-space \(\mathcal{J}_{\ell,k}\) is defined as
\begin{equation}\label{eq:j-space}
	\mathcal{J}_{\ell,k}
	=
	\left\{
	\sum_{w\in\mathcal{V}}
	a_w\bar\phi_{\ell,w}
	:
	a_w\geq0,
	\;
	\|a\|_0\leq k
	\right\}
	\subseteq
	\mathbb{R}^{d},
\end{equation}

\noindent
where \(a=(a_w)_{w\in\mathcal{V}}\) is the vector of coefficients and \(\|a\|_0\) counts how many coefficients are nonzero.

\begin{figure}[tbp]
    \centering
    \includegraphics[width=\linewidth]{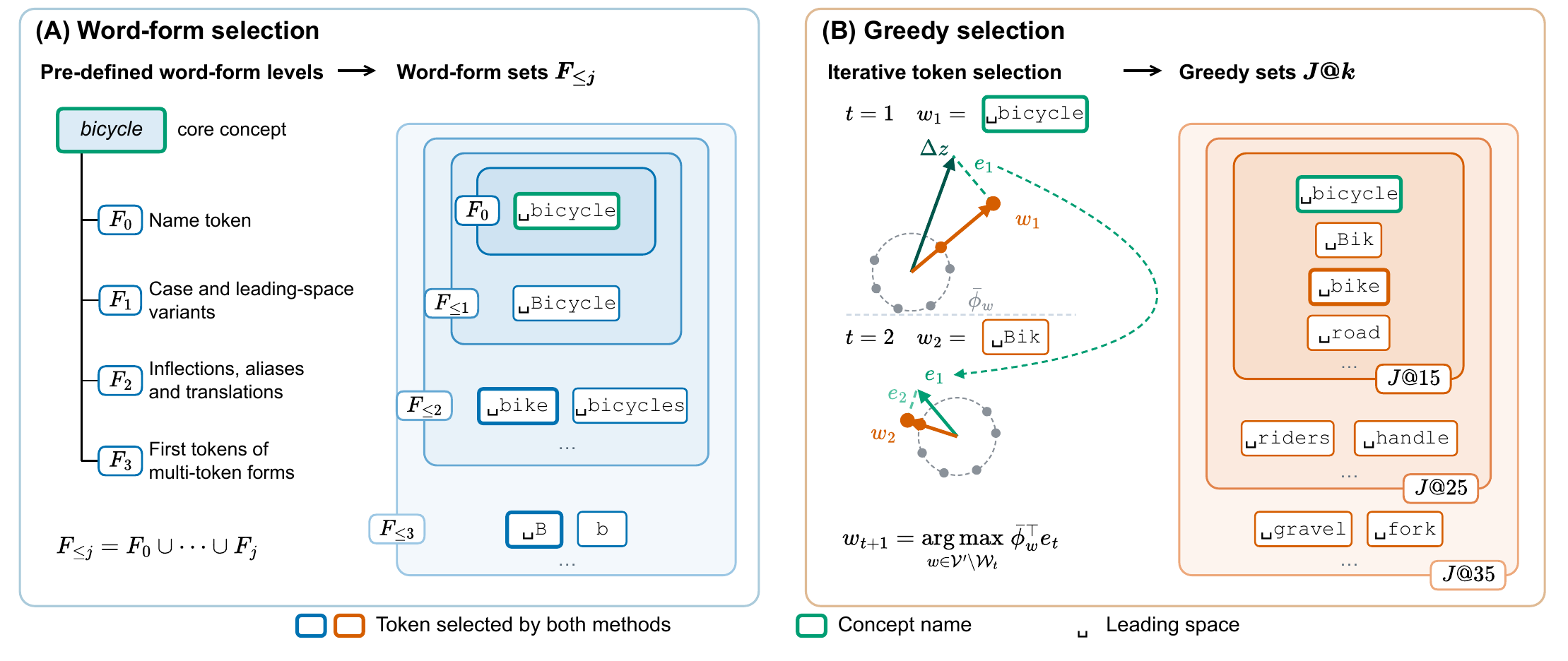}
    \caption{\textbf{Two ways to construct concept direction sets for \emph{bicycle}.}
    \textbf{(A)} Word-form selection starts with the name-token set \(F_0\) and adds tokens for further word forms to obtain the cumulative word-form sets \(F_{\leq j}\).
    \textbf{(B)} Greedy selection reconstructs the state difference \(\Delta z\) using vocabulary directions, producing the greedy sets \(\greedyset{k}\) for different budgets \(k\).}
    \label{fig:concept-direction-sets}
\end{figure}

We select the tokens of the concept direction sets in two ways, word-form selection and greedy selection, as illustrated for \emph{bicycle} in Figure~\ref{fig:concept-direction-sets}.

\paragraph{Word-form selection.}
\phantomsection
\begingroup
\makeatletter
\edef\@currentlabel{\thesubsection}
\label{sec:word-form-directions}
\makeatother
\endgroup
Word-form selection picks the tokens that name the concept, including its case forms, inflections, aliases and translations.
For each concept \(c\), we first fix its name, such as ``bicycle''.
The four levels below collect the tokens derived from this name and from related words, as Figure~\ref{fig:concept-direction-sets}~(A) illustrates for \emph{bicycle}.
In the examples, \texttt{\char32} marks a leading space.
\begin{itemize}
    \item \(F_0(c)\): the name token, the concept name with a leading space, if it is a single token (e.g.\ \texttt{\char32 bicycle}).
    \item \(F_1(c)\): other single-token forms of the concept name, differing in capitalisation or leading space (e.g.\ \texttt{\char32 Bicycle}).
    \item \(F_2(c)\): other single-token words for the concept, including inflections, aliases and translations (e.g.\ \texttt{\char32 bike} and \texttt{\char32 bicycles}).
    \item \(F_3(c)\): the first tokens of multi-token forms of the concept name (e.g.\ \texttt{b} from \texttt{bicycle} and \texttt{\char32 B} from \texttt{\char32 BICYCLE}).
\end{itemize}
We combine the levels into cumulative word-form sets:
\begin{equation}\label{eq:word-form-direction-set}
	F_{\leq j}(c)
	=
	\bigcup_{i=0}^{j}F_i(c)
	\subseteq\mathcal{V},
	\qquad
	j\in\{0,1,2,3\}.
\end{equation}
We call \(F_0(c)\) the name-token set, since it contains only the name token.

\paragraph{Greedy selection.}
\phantomsection
\begingroup
\makeatletter
\edef\@currentlabel{\thesubsection}
\label{sec:greedy-direction-selection}
\makeatother
\endgroup
The word-form selection we described above is based on pre-defined tokens, which does not guarantee that the hidden state actually uses these vocabulary directions for expressing the concept.
Here we introduce greedy selection to reconstruct the concept-specific part of the hidden state, using vocabulary directions chosen from the whole vocabulary.
J-Lens uses a related sparse decomposition to analyse which vocabulary directions are active in a state~\citep{gurnee2026verbalizable}.
We instead use the selected tokens as the direction set of a concept, for both readout and intervention.

Let \(h_\ell(x)\in\mathbb R^d\) be the hidden state for an input \(x\) expressing concept \(c\), and let \(\mathcal R(c)\) be the set of other concepts used for comparison under the same question.
For each comparison concept \(c'\), we average its states over its \(n_{c'}\) inputs \(x_{i,c'}\).
The comparison mean \(\mu_{\ell,c}^{\mathrm{comp}}\) and the state difference \(\Delta z_{\ell,c}\) are
\begin{align}
	\mu_{\ell,c}^{\mathrm{comp}}
	&=\frac{1}{|\mathcal R(c)|}
	\sum_{c'\in\mathcal R(c)}
	\left(
	\frac{1}{n_{c'}}\sum_{i=1}^{n_{c'}}h_\ell(x_{i,c'})
	\right),
	\label{eq:comparison-mean}\\
	\Delta z_{\ell,c}
	&=h_\ell(x)-\mu_{\ell,c}^{\mathrm{comp}}.
	\label{eq:readout-contrast}
\end{align}
We subtract this mean \(\mu_{\ell,c}^{\mathrm{comp}}\) to reduce information shared across the inputs.
We then use the centred state \(\Delta z_{\ell,c}\) as the reconstruction target.

\noindent\begin{minipage}{\linewidth}
\begin{minipage}[t]{0.49\linewidth}
\vspace{0pt}
We want to find a set of at most \(k\) vocabulary directions whose non-negative combination is closest to \(\Delta z_{\ell,c}\).
However, it is not feasible to search over all choices of \(k\) tokens, so we build the set one token at a time with a greedy procedure based on orthogonal matching pursuit~\citep{mallat1993matching,pati1993orthogonal}.

Let \(\mathcal V'\subseteq\mathcal V\) be the eligible tokens, whose text contains at least one letter.
Let \(\mathcal W_t\) be the tokens selected after \(t\) iterations and \(B_t\) be the matrix whose columns are the unit vocabulary directions of the tokens in \(\mathcal W_t\).
\(e_t\) is the remaining difference.
We start with \(\mathcal W_0=\varnothing\), an empty \(B_0\) and \(e_0=\Delta z_{\ell,c}\).
Each iteration has three steps, illustrated in Figure~\ref{fig:greedy-selection-iteration}.
\end{minipage}\hfill
\begin{minipage}[t]{0.48\linewidth}
\vspace{0pt}
\centering
\includegraphics[width=\linewidth]{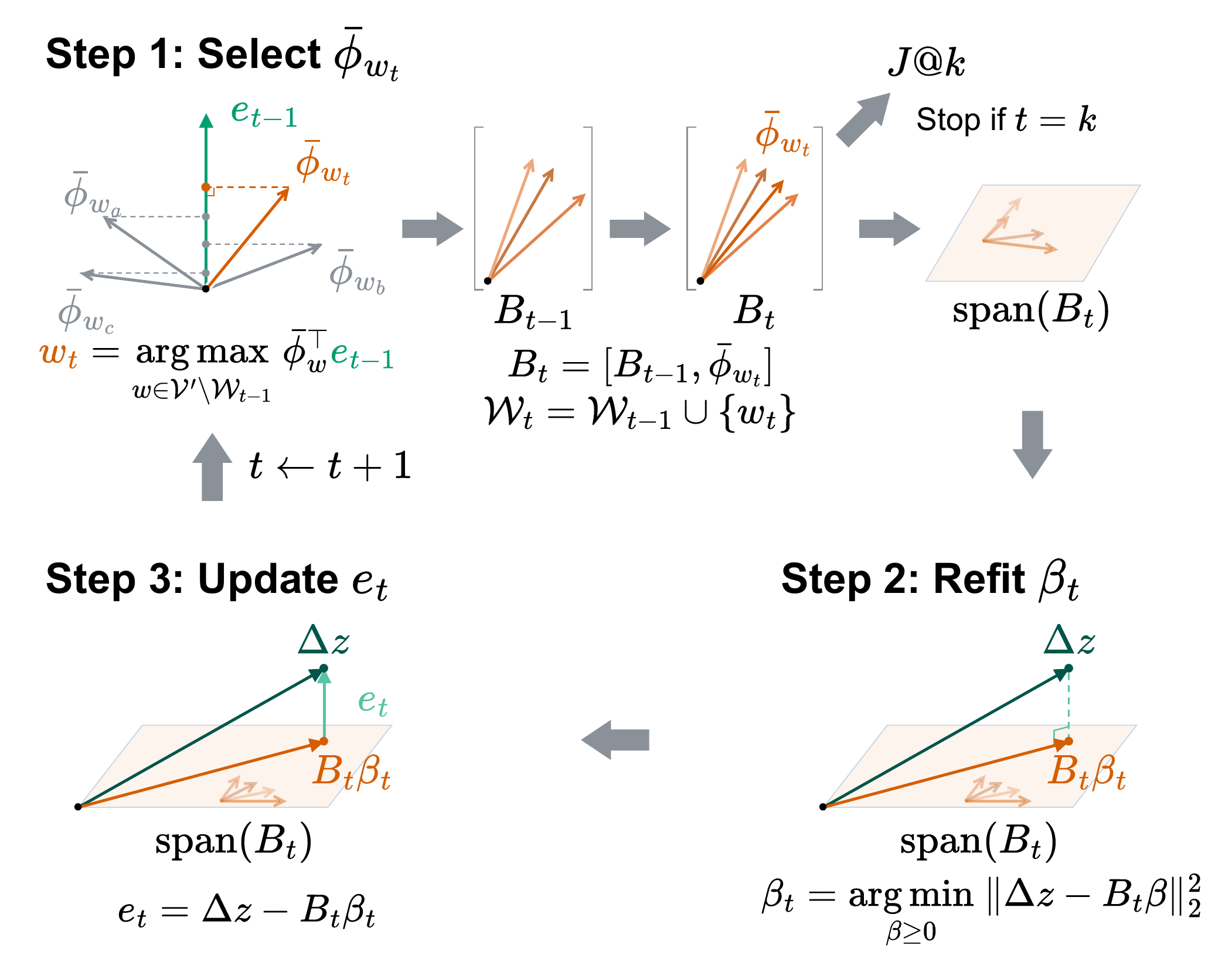}
\captionsetup{type=figure,font=small,justification=raggedright,singlelinecheck=false}
\caption{\textbf{Greedy selection.}
One iteration selects a direction, refits all coefficients by non-negative least squares (NNLS), and updates the remaining difference.
Layer and concept subscripts are omitted.}
\label{fig:greedy-selection-iteration}
\end{minipage}
\end{minipage}
\par\medskip
Step 1: we pick the token whose vocabulary direction has the largest inner product with the remaining difference.
We then add it to the set and append its unit direction to the matrix:
\begin{equation}\label{eq:pursuit-selection}
	w_{t}
	=\operatorname*{arg\,max}_{w\in\mathcal V'\setminus\mathcal W_{t-1}}
	\bar\phi_{\ell,w}^{\top}e_{t-1}.
\end{equation}
\begin{equation}\label{eq:pursuit-set-update}
	\begin{aligned}
	\mathcal W_{t}&=\mathcal W_{t-1}\cup\{w_{t}\},\\
	B_{t}&=
	\begin{bmatrix}
	B_{t-1},&\bar\phi_{\ell,w_{t}}
	\end{bmatrix}
	\in\mathbb R^{d\times t}.
	\end{aligned}
\end{equation}

Step 2: we refit the coefficients of all selected directions by non-negative least squares (NNLS)~\citep{lawson1995solving}, which keeps them non-negative as the J-space requires:
\begin{equation}\label{eq:pursuit-refit}
	\beta_{t}
	=\operatorname*{arg\,min}_{\beta\in\mathbb R^{t},\;\beta\geq0}
	\left\|\Delta z_{\ell,c}-B_{t}\beta\right\|_2^2.
\end{equation}

Step 3: we subtract the fitted combination from \(\Delta z_{\ell,c}\) to obtain the new remaining difference:
\begin{equation}\label{eq:pursuit-error}
	e_{t}=\Delta z_{\ell,c}-B_{t}\beta_{t}.
\end{equation}

We repeat the iteration until \(k\) tokens have been selected, or earlier if no remaining direction has a positive inner product with \(e_t\).
We call the set of tokens selected with budget \(k\) the greedy set \(\greedyset{k}\).
A larger budget collects more tokens, so the greedy set can reach related concepts beyond the word forms.

In the bicycle VQA example illustrated in Figure~\ref{fig:concept-direction-sets}~(B), \(\greedyset{15}\) contains \texttt{\char32 bicycle}, \texttt{\char32 Bik}, \texttt{\char32 bike} and \texttt{\char32 road}, \(\greedyset{25}\) adds \texttt{\char32 riders} and \texttt{\char32 handle}, and \(\greedyset{35}\) adds \texttt{\char32 gravel} and \texttt{\char32 fork}, among other tokens.
The tokens \texttt{\char32 bicycle} and \texttt{\char32 bike} are also in the word-form sets, and the other tokens lie outside them.
Most of these name related concepts, such as \texttt{\char32 road} and \texttt{\char32 riders}, which the word forms do not cover.

\subsection{Counterfactual Interventions on Concept Representations}
\label{sec:concept-interventions}
\suppressfloats[t]

Suppose that, in a VQA task, the input is an image of a Canadian flag and the question asks for the capital of the flag's country.
The model answers ``Ottawa''.
Section~\ref{sec:vocabulary-readout} tells us whether the concept \emph{Canada} is readable in an intermediate hidden state, and Section~\ref{sec:concept-direction-sets} gives the concept directions for \emph{Canada}.
Here we want to find out whether the concept of \emph{Canada} we access in the intermediate state is used to ``compute'' the answer about its capital, or is merely shown on a ``display board''.

\begin{figure}[tbp]
\centering
\includegraphics[width=\linewidth]{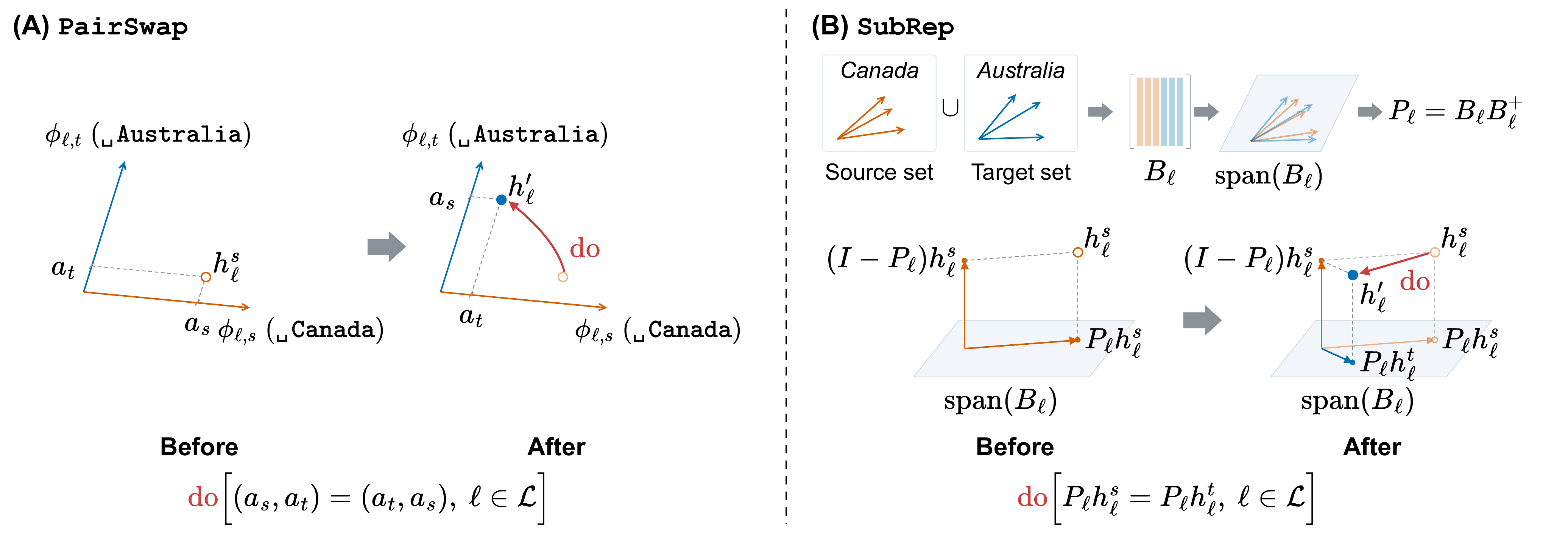}
\captionsetup{font=small,justification=raggedright,singlelinecheck=false}
\caption{\textbf{Concept interventions.}
\textbf{(A)} \texttt{PairSwap} swaps the source and target name-token coordinates within the source state.
\textbf{(B)} \texttt{SubRep} replaces the source state's projected component with that of the target state, while retaining the source state's orthogonal component.
Red arrows denote interventions.}
\label{fig:concept-interventions}
\end{figure}

We define the counterfactual intervention here as setting the hidden states of the model at chosen layers to chosen values during the forward pass, and letting the later layers compute from the edited states~\citep{meng2022locating,geiger2024das}.
We call the concept whose representation we want to change the \textbf{source} concept, and the concept we want to write into the state the \textbf{target} concept.
In our example illustrated in Figure~\ref{fig:concept-interventions}, the source and target concepts are \emph{Canada} and \emph{Australia}, respectively.

We first run the model on the source input \(x^{s}\) and the target input \(x^{t}\), each containing an image of its concept and the same question.
This gives the hidden states \(h_\ell^{s}=h_\ell(x^{s})\) and \(h_\ell^{t}=h_\ell(x^{t})\) at the same layer and position.
We then intervene on the source forward pass at a set of layers \(\mathcal L\), and \(\operatorname{do}[\,\cdot=\cdot\,]\) denotes an intervention that sets the variable on the left to the value on the right~\citep{pearl2009causality}.
The layers in \(\mathcal L\) are edited in order of depth, so the state at a deeper edited layer already reflects the edits at the shallower layers.

By comparing the model's answers before and after the intervention, we can investigate the role of the accessed concept in downstream reasoning.
In our example, if the answer changes from ``Ottawa'' to ``Canberra'', this shows that changing the country representation can influence the model's computation of the capital answer.

For specific intervention methods, \citet{gurnee2026verbalizable} intervene on a single-token concept with pairwise lens-coordinate swapping (\texttt{PairSwap}).
Specifically, let \(a_s\) and \(a_t\) be the coordinates of \(h_\ell^{s}\) along the raw vectors \(\phi_{\ell,s}\) and \(\phi_{\ell,t}\) for the source and target name tokens.
\texttt{PairSwap} exchanges the two coordinates and keeps the rest of the state unchanged (Figure~\ref{fig:concept-interventions}~(A)):
\begin{equation}\label{eq:do-pairswap}
	\operatorname{do}\bigl[(a_s,a_t)=(a_t,a_s),\ \ell\in\mathcal L\bigr].
\end{equation}
However, the \texttt{PairSwap} operator relies on pairing one token with another, so it does not directly apply to a whole concept direction set such as \(F_{\leq j}\) or \(\greedyset{k}\).

Therefore, we introduce subspace replacement (\texttt{SubRep}), a counterfactual intervention operator that writes the target component into the source state along a whole set of concept directions.
At layer \(\ell\), we take the concept directions of the source and target concepts from the word-form sets \(F_{\leq j}\) or the greedy sets \(\greedyset{k}\).\footnote{For computational efficiency, greedy selection for intervention uses a different coefficient update from the NNLS fit used for reconstruction. See Appendix~\ref{app:greedy-intervention-directions} for details.}
We combine the directions of the two concepts, and the resulting \(n\) unit directions lie in \(\mathbb R^{d^D}\) and form the columns of a direction matrix:
\begin{equation}\label{eq:intervention-direction-matrix}
	B_\ell=
	\begin{bmatrix}
	\bar\phi_{\ell,w_1}&\cdots&\bar\phi_{\ell,w_n}
	\end{bmatrix}
	\in\mathbb R^{d^D\times n}.
\end{equation}
The orthogonal projector onto the span of these directions is \(P_\ell=B_\ell B_\ell^+\), where \(B_\ell^+\) is the Moore-Penrose pseudoinverse.\footnote{We use the pseudoinverse because the concept directions need not be orthogonal or linearly independent.}
For a hidden state \(h\), \(P_\ell h\) is its component in the subspace, and \((I-P_\ell)h\) is the component orthogonal to the subspace, where \(I\) is the identity matrix.

Specifically, \texttt{SubRep} replaces \(P_\ell h_\ell^{s}\) with \(P_\ell h_\ell^{t}\), and keeps \((I-P_\ell)h_\ell^{s}\) unchanged, as in distributed interchange interventions~\citep{geiger2024das} (Figure~\ref{fig:concept-interventions}~(B)):
\begin{equation}\label{eq:do-subrep}
	\operatorname{do}\bigl[P_\ell h_\ell^{s}=P_\ell h_\ell^{t},\ \ell\in\mathcal L\bigr].
\end{equation}
The edited state is therefore
\begin{equation}\label{eq:subrep}
	h_\ell'
	=
	\underbrace{(I-P_\ell)h_\ell^s}_{\text{retained source contribution}}
	+
	\underbrace{P_\ell h_\ell^t}_{\text{added target contribution}}.
\end{equation}

Besides \texttt{SubRep} and \texttt{PairSwap}, we compare ablation and injection (\texttt{AbIn}) and full-state replacement (\texttt{FullRep}), illustrated in Figure~\ref{fig:intervention-overview}~(B):
\begin{equation}\label{eq:do-operators}
	\begin{aligned}
	\texttt{AbIn}:&\quad\operatorname{do}\bigl[h_\ell^{s}=(I-P_\ell^{s})h_\ell^{s}+\alpha r_\ell v_\ell^{t},\ \ell\in\mathcal L\bigr],\\
	\texttt{FullRep}:&\quad\operatorname{do}\bigl[h_\ell^{s}=h_\ell^{t},\ \ell\in\mathcal L\bigr].
	\end{aligned}
\end{equation}
\texttt{AbIn} removes the source component with the projector \(P_\ell^{s}\) onto the source concept directions and adds the mean target direction \(v_\ell^{t}\) at a chosen scale \(\alpha r_\ell\), as in directional ablation~\citep{arditi2024refusal} and activation addition~\citep{turner2023activation}, respectively.
\texttt{FullRep} sets the whole state to the target state, and we use it as a reference in Section~\ref{sec:cross-task-reuse}.
Appendix~\ref{app:intervention-operators} gives the complete definitions of all four intervention operators.

\Needspace{8\baselineskip}
\section{Multimodal Concept Accessibility}
\label{sec:multimodal-readout}

In brain science, GWT proposes that conscious access involves making information available to multiple specialised processes through a shared workspace~\citep{baars2005workspace}.

To investigate the potential relationship between GWT and multimodal information integration and processing in VLMs, we use our proposed MC-Lens to examine concept accessibility in intermediate states.
We first examine concept readability in intermediate states under different combinations of multimodal inputs.
We then extend the analysis to different ``readout carriers'', from a single name token to sets of tokens, comparing their ability to express the target concept and reconstruct the intermediate state.
We finally examine spatial-level concept accessibility in visual states across network layers and image locations.
Together, these experiments provide a preliminary evaluation of MC-Lens and these ``readout carriers'' for studying multimodal information in intermediate states.

\subsection{Concept Accessibility across Evidence Conditions}
\label{sec:multimodal-readout-selection}

In this part, we investigate concept accessibility in intermediate states with MC-Lens, under different conditions of model input.
Specifically, we use VQA as a task setting to compare concept accessibility when relevant task evidence is provided by an image or by text.
As illustrated in Example~\ref{ex:multimodal-readout}, we construct four evidence conditions for each question: 1) an image containing evidence for the target concept, with no extra text evidence, 2) a sentence explicitly naming the target concept, with a blank image, 3) both, or 4) neither.
We collect 100 VQA cases from A-OKVQA~\citep{schwenk2022aokvqa}, each with a single-word answer, such as the name of an object or a colour.
For each case, we use the ground-truth answer as the target concept for our experiment, and the target concept is fixed across the four evidence conditions.

We apply the proposed L-Lens to the pre-answer position in decoder layer D32 of Qwen3-VL-8B in this experiment, and then analyse the rank of the target concept token among all vocabulary tokens based on the readout scores.
We measure \(\operatorname{Pass}@K\) (where \(K\) is a rank threshold), the fraction of cases where the target token appears in the top-\(K\) ranking, and higher values indicate better concept accessibility.

\noindent\begin{minipage}{\linewidth}
\begin{minipage}[t]{0.49\linewidth}
\vspace{0pt}
As illustrated in Figure~\ref{fig:evidence-readout}, both image and text evidence conditions are able to raise the readability of the target concept under the three \(\operatorname{Pass}@K\) thresholds (\(K=1,5,10\)).
Text explicitly naming the target concept tends to give higher readout rates than image evidence only, while providing both sources gives similar rates to text alone.
Example~\ref{ex:multimodal-readout} illustrates this comparison with a VQA task about a skateboard move.
Without either form of evidence, the target concept token \texttt{\char32 grind} ranks 9\textsuperscript{th}.
Providing the skateboard image, the sentence naming \emph{grind}, or both raises the target concept token's readout rank to 1\textsuperscript{st} place.

This experiment shows that both image and text evidence promote concept readability in intermediate states.
\end{minipage}\hfill
\begin{minipage}[t]{0.48\linewidth}
\vspace{0pt}
\centering
\includegraphics[width=\linewidth]{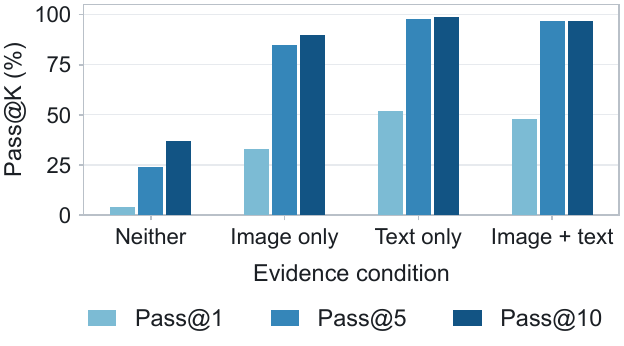}
\captionsetup{type=figure,font=small,justification=raggedright,singlelinecheck=false}
\caption{\textbf{Concept accessibility across evidence conditions.}
    Bars show the percentage of cases in which the target token appears among the top 1, 5 or 10 readout tokens.
    The question and target concept remain fixed across the four evidence conditions.}
\label{fig:evidence-readout}
\end{minipage}
\end{minipage}
\par\medskip
\input{cards/multimodal-readout}

\subsection{Extending Concept Readout Carriers}
\label{sec:readout-reconstruction}

In Section~\ref{sec:multimodal-readout-selection}, we used a single name token as the ``readout carrier'' for a concept, following J-Lens~\citep{gurnee2026verbalizable}.
This token gives one way to name the concept (e.g., \texttt{\char32 bicycle} \(\rightarrow\) the concept of a \emph{bicycle}), but the same concept can be expressed through different tokens (e.g., the concept of a \emph{bicycle} \(\rightarrow\) \texttt{\char32 bicycle}, \texttt{\char32 bike}, \texttt{\char32 Bicycle}, etc.).

Here, we extend the analysis to sets of tokens as ``readout carriers''.
Specifically, we compare two types of concept direction sets: the word-form sets \(F_{\leq j}\) and the greedy sets \(\greedyset{k}\), as illustrated in Figure~\ref{fig:concept-direction-sets}.
We want to examine the readability of these sets and their ability to reconstruct the state.

We collect 100 VQA cases from A-OKVQA~\citep{schwenk2022aokvqa} and build 50 VQA cases for country name identification using national-flag images from Noto Color Emoji.\footnote{Noto Color Emoji: \url{https://github.com/googlefonts/noto-emoji}. The actual number of flag images included in the quantitative analysis is 46, since four countries lack a complete single-token name for \(F_0\).}
For each case, we use the ground-truth answer as the target concept and compare the sets on the same intermediate state.
We include the name-token set \(F_0\), the cumulative word-form sets \(F_{\leq1}\), \(F_{\leq2}\) and \(F_{\leq3}\), and the greedy sets \(\greedyset{15}\), \(\greedyset{25}\) and \(\greedyset{35}\) for comparison.
For readout, we apply L-Lens at decoder layer D32 of Qwen3-VL-8B to the pre-answer state.

To measure the overall readability of a set, we use \(\operatorname{Coverage}@10\), the proportion of tokens in the top-ten readout that belong to the set.
Higher values mean that the target concept is readable through more tokens in the set.
We use the reconstruction score (Recon. Score) to measure a set's ability to reconstruct the centred intermediate state, and higher values indicate that the selected concept directions better capture the information in the state.

\begin{figure}[tbp]
    \centering
    \includegraphics[width=\linewidth]{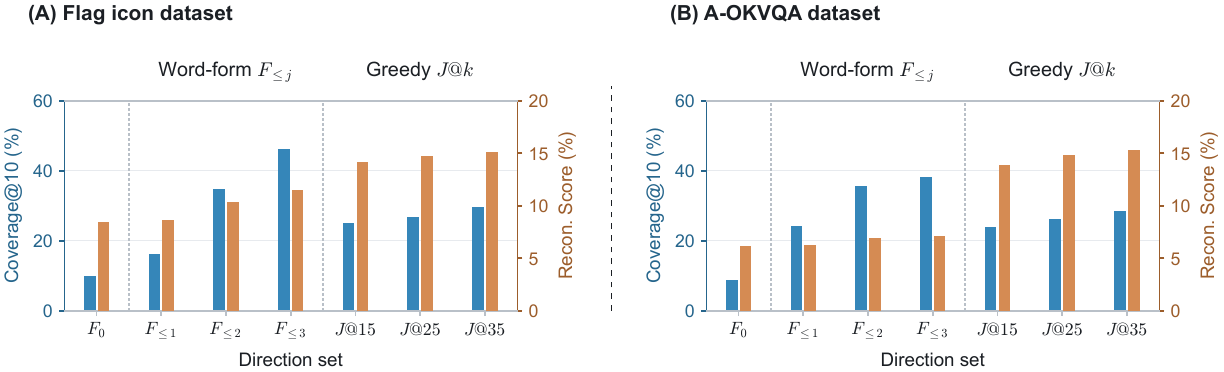}
    \caption{\textbf{Concept readout and state reconstruction.}
    Bars compare word-form sets \(F_{\leq j}\) and greedy sets \(\greedyset{k}\) for \textbf{(A)} flag icons and \textbf{(B)} natural images from A-OKVQA.
    Blue bars show \(\operatorname{Coverage}@10\) (left axis), and orange bars show Recon. Score (right axis), both as percentages.}
    \label{fig:readout-reconstruction}
\end{figure}

As shown by the blue bars in Figure~\ref{fig:readout-reconstruction} (A,B), both the expanded word-form sets \(F_{\leq j}\) and the greedy sets \(\greedyset{k}\) have higher \(\operatorname{Coverage}@10\) than the name-token set \(F_0\).
For \(F_{\leq j}\), \(\operatorname{Coverage}@10\) increases substantially as \(j\) increases.
The expanded sets cover more readable word forms of the same concept.
For \(\greedyset{k}\), \(\operatorname{Coverage}@10\) increases only slightly as \(k\) increases from 15 to 35, showing that few of the additional tokens are in the top ten.

For the reconstruction score, both the expanded word-form sets \(F_{\leq j}\) and the greedy sets \(\greedyset{k}\) outperform \(F_0\), as shown by the orange bars in Figure~\ref{fig:readout-reconstruction} (A,B).
The greedy sets \(\greedyset{k}\) give substantially higher reconstruction scores than the word-form sets \(F_{\leq j}\), consistent with their selection based on reconstruction quality.
However, their \(\operatorname{Coverage}@10\) is lower than that of \(F_{\leq3}\), showing that higher top-ten coverage does not necessarily give better state reconstruction.

\input{cards/word-form-readout}
\input{cards/concept-reconstruction}

Example~\ref{ex:word-form-readout} shows that the concept of \emph{Canada} can be accessed through multiple word forms, such as ``Canada'', ``Canadian'' and ``\begin{CJK}{UTF8}{gbsn}加拿大\end{CJK}'', with different levels of readability.
Example~\ref{ex:concept-reconstruction} shows that the greedy set \(\greedyset{25}\) for \emph{Canada} includes tokens outside \(F_{\leq2}\) with relatively large reconstruction weights, despite their lower readout ranks.
For example, \texttt{\char32 maple} ranks 116\textsuperscript{th} in the readout but is among the five tokens with the largest reconstruction weights.

\begin{samepage}
Overall, expanding \(F_{\leq j}\) mainly increases coverage of the top-ten readout, whereas the greedy sets \(\greedyset{k}\) reach beyond those highly readable words and give better state reconstruction than the word-form sets.
\par
\end{samepage}

\subsection{Concept Accessibility in Visual States}
\label{sec:visual-readout}

In Sections~\ref{sec:multimodal-readout-selection} and~\ref{sec:readout-reconstruction}, we examined concept accessibility in the pre-answer state in the language decoder using L-Lens, and here we extend our analysis to intermediate states at visual token positions.
We first examine concept accessibility across network sub-modules and depths, and then examine the correspondence between patch-level concept readouts and the locations of objects and attributes in the image.

\paragraph{Visual concept readout across network depth.}

In this experiment, we examine visual concept accessibility across network sub-modules and depths.

We select 32 object concepts, with four images for each concept (128 images in total) from THINGS \citep{hebart2019things}.
For each image, we use the main object shown as the target concept.
Since we want to evaluate concept accessibility in visual states without the influence of a text question, we use only the image as the user input in this experiment.
This experiment is conducted on Qwen3-VL-8B, using both V-Lens and L-Lens.
For the visual encoder, we use V-Lens to read the mean encoder patch state, and for the language decoder, we use L-Lens to read the mean decoder image-token state.
For the connector, we read the mean visual state using the lenses fitted to its intermediate states (Appendix~\ref{app:lens-fitting}).
As described in Section~\ref{sec:vocabulary-readout}, we apply the centring strategy here to make the concept more readable (Appendix~\ref{app:visual-readout-aggregation-centring}).

For each image, we rank vocabulary tokens by their readout scores, and record the best rank among the complete single-token word forms of the target object's name.
For example, if \texttt{\char32 Bicycle} ranks 3\textsuperscript{rd} and \texttt{\char32 Bike} ranks 7\textsuperscript{th}, we record rank 3.
We report the median of these ranks across images as the median target rank.
We also measure \(\operatorname{Pass}@10\), the percentage of images for which at least one of these tokens appears in the top-ten readout.
A lower median target rank and a higher \(\operatorname{Pass}@10\) both indicate greater object-name readability.

\begin{figure}[tbp]
    \centering
    \includegraphics[width=\linewidth]{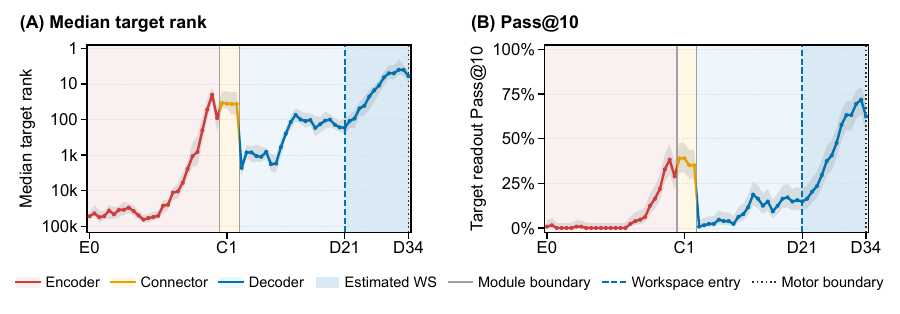}
    \caption{\textbf{Visual concept readout across network depth.}
    Object-name readability in the centred mean visual state at each layer, without a text question, measured by \textbf{(A)} median target rank and \textbf{(B)} \(\operatorname{Pass}@10\).
    Grey bands show 95\% confidence intervals.
    The darker blue region marks the estimated workspace (WS).}
    \label{fig:visual-concept-readout}
\end{figure}

\begin{figure}[tbp]
\begin{examplecard}{Spatial Concept Readout with V-Lens}
\exhibitlabel{ex:spatial-concept-readout}
\newcommand{\efourdivider}{%
  \par\vspace{5pt}\nointerlineskip
  \hbox to\linewidth{\color{black}%
    \leaders\hbox{\vrule width 2.5pt height 0.4pt depth 0pt\kern2.5pt}\hfill}%
  \vspace{5pt}\nointerlineskip
}
{\raggedright\footnotesize
\textbf{Input:} The kitchen and desk images are supplied without a question.\par
\textbf{Readout:} Each patch is coloured by the highest-scoring word within the object or attribute set.
White means no candidate ranks in the top 10,000.\par}
\efourdivider
\centering
\includegraphics[width=\linewidth]{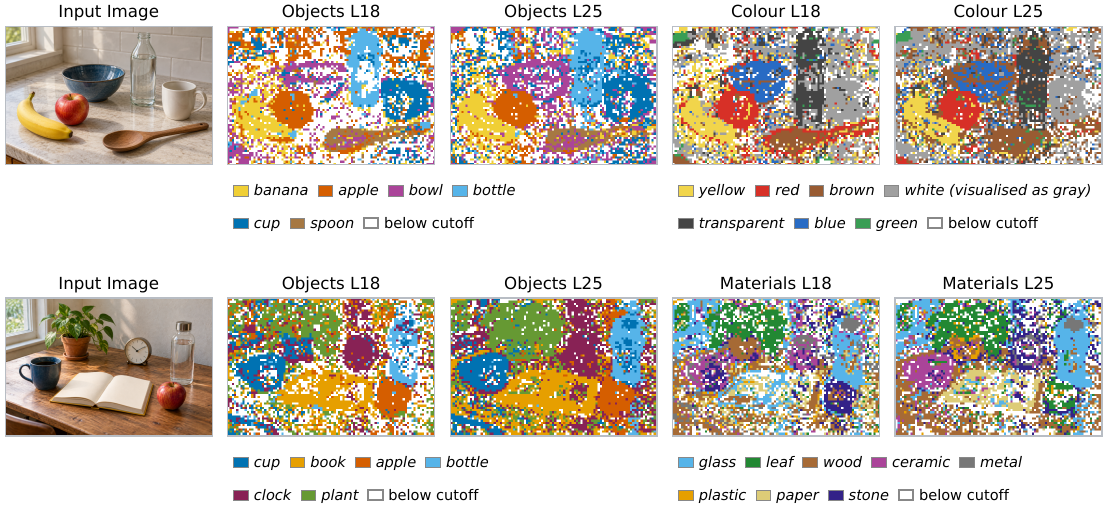}
\efourdivider
{\raggedright\footnotesize
\textbf{Finding:} From L18 to L25, object names are readable at more patches, and the maps more clearly show the arrangement of objects in both scenes.
These spatial readouts extend beyond object names to colours and materials, although material readouts remain more scattered.\par}
\end{examplecard}
\end{figure}

As shown in Figure~\ref{fig:visual-concept-readout}~(A,B), object-name readability improves towards the end of the encoder and remains at a similar level through the connector.
It drops at the start of the decoder and then improves again in later decoder layers.
The shaded region marks the estimated workspace, whose boundaries and interpretation are discussed in Section~\ref{sec:workspace-organisation}.
In the later layers of the estimated workspace, the median target rank is better and \(\operatorname{Pass}@10\) is higher than in the later encoder layers.
In the encoder, concepts in intermediate states are accessible using V-Lens and become more accessible with increasing depth.
In the decoder, even without a text question, we can still use L-Lens to access concepts from intermediate states at visual token positions, and accessibility increases substantially within the estimated workspace.

\paragraph{Spatial concept readout.}

V-Lens can provide a spatial-level concept map.
Specifically, we apply V-Lens to individual image patches at encoder L18 and L25, using only the image as the user input in this experiment.

Example~\ref{ex:spatial-concept-readout} illustrates two scenes. For each scene, we choose a set of object names and a separate set of attribute words: colours for the kitchen scene and material-related words for the desk scene.
We compare readout scores within each set and colour each patch by its highest-scoring word, producing separate object and attribute maps. Patches remain white if the winning word falls outside the top 10,000 tokens (cutoff).

At L18, several objects are already distinguishable, but many patches remain white.
At L25, more patches have readable object names, and the arrangement of objects is clearer, especially in the desk scene.
In the kitchen scene, \textit{banana} and \textit{apple} mark the fruit on the left, and \textit{cup} marks the mug on the right.
In the desk scene, \textit{book} highlights the open pages, and interestingly, the pot region is labelled as \textit{cup} at L18 but as \textit{plant} at L25.
This change suggests that the concept readout can better match the object in the image at a deeper layer.

The attribute maps show that these spatial readouts extend beyond object names.
In the kitchen scene, \textit{yellow} and \textit{red} coincide with the banana and apple, while \textit{blue} identifies the bowl.
In the desk scene, \textit{glass} highlights the bottle, and \textit{ceramic} marks the cup.
Material readouts remain more scattered than the object and colour maps, and some labels extend beyond the corresponding objects.

In this experiment, the concept maps generated by V-Lens allow us to examine the model's understanding of multimodal information in intermediate states at a fine spatial scale, and to analyse it from different concept dimensions.

\section{Selective Access to Multimodal Concepts}
\label{sec:selective-access}

An important feature described by GWT is that information generated simultaneously by different specialised processes competes for access to a shared workspace.
The selected information then becomes consciously accessible~\citep{baars2007architectural}.
In this section, we try to identify a similar selective access mechanism for multimodal concepts in the intermediate states of VLMs, supported by our proposed MC-Lens.

Specifically, we first examine how the text query affects concept accessibility in intermediate states, by changing the text query for a fixed image.
We then further track changes in concept accessibility during autoregressive generation.
Together, these experiments provide evidence for selective access to multimodal concepts in the intermediate states of VLMs.

\subsection{Query-Dependent Selective Access to Visual Concepts}
\label{sec:question-dependent-selection}

In this experiment, we investigate selective access to visual concepts through MC-Lens readouts and measure the relative accessibility of concepts in the same scene.

We collect 100 scene images from COCO~\citep{lin2014coco}.
For each image, we select three visible objects as the target concepts for our experiment and prepare one question for each object.
Each question identifies its target through colour, position, or a relation to other scene content, without naming it.
The VLM receives the same image and one question at a time, and we query each image three times using three different questions.
We apply L-Lens to the pre-answer position at each decoder layer of Qwen3-VL-8B, and record the ranks of all three target object-name tokens for comparison.

\input{cards/query-object-selection}

To measure ``selectivity'', we define successful selection for a question as the requested object's name token ranking above the name tokens of the other two objects.
Therefore, selective sensitivity can be defined as the percentage of images with successful selection under all three questions at the same decoder layer.
Higher values indicate more consistent query-dependent selection.

To account for potential concept information introduced by the text question itself, we repeat the same questions with a uniform grey image (blank) to measure the contribution of visual evidence to selection.

\Needspace{24\baselineskip}
\begin{wrapfigure}{r}{0.48\linewidth}
\centering
\includegraphics[width=\linewidth]{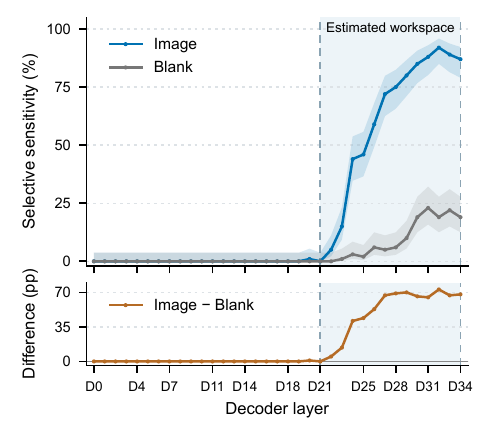}
\captionsetup{type=figure,font=small,justification=raggedright,singlelinecheck=false}
\caption{\textbf{Query-dependent selection of visual concepts.}
Top: selective sensitivity for original images (Image) and uniform grey images (Blank).
Bands show 95\% confidence intervals.
Bottom: Image minus Blank, in percentage points (pp).}
\label{fig:question-dependent-selection}
\end{wrapfigure}

As shown in the upper panel of Figure~\ref{fig:question-dependent-selection}, selective sensitivity remains close to zero in earlier decoder layers and rises sharply within the estimated workspace when the original image is provided.
With a grey image baseline, the increase is much smaller.
The lower panel shows the difference in selective sensitivity between Image and Blank, and this difference still becomes larger within the estimated workspace.
We find that the text query can substantially increase the relative accessibility of the corresponding concepts among the compared concepts in the same scene.
These results demonstrate that the text query can drive selective access.

We also provide three examples to examine selective access to objects, their associated attributes and other scene content.
In Example~\ref{ex:query-object-selection}, we examine query-dependent selection of objects and their associated colours.
We use an image containing a blue cup on the left and a yellow book on the right, and alternate between questions about the left and right objects.
We compare the readability of \textit{cup}, \textit{book}, \textit{blue} and \textit{yellow} under the left-object query and the right-object query.

We can find that before the estimated workspace, concept accessibility under the two different queries is similar, with little separation between the queries.
In contrast, within the estimated workspace, the left-object question makes \textit{cup} more readable than \textit{book}, while the right-object question reverses this order.
Across much of the estimated workspace, the colour concept \textit{blue} is also more readable under the left-object question and \textit{yellow} under the right-object question.
Since neither question requests a colour, this change suggests that the selective access mechanism can extend beyond the answer name to an associated attribute, and this selective pattern becomes pronounced within the estimated workspace.

\input{cards/scene-region-selection}

In Example~\ref{ex:scene-region-selection}, we analyse the relationship between this selectivity and spatial relations.
Specifically, the example image contains a foreground dog and a distant cat beside a tree, and we keep the image fixed and use a neutral query, a foreground query and a distance query.

Within the estimated workspace, neutral and foreground questions make \textit{dog} more readable than \textit{cat} and \textit{tree}, and the model answers ``dog''.
The distance question instead makes both \textit{cat} and \textit{tree} more readable within the estimated workspace.
\textit{Tree} becomes highly readable at earlier decoder layers than \textit{cat}, even though the model ultimately answers ``cat''.
Under a neutral query, the visually prominent \textit{dog} is more accessible.
When the text query specifies the distant region, it directs access towards \textit{cat} and \textit{tree}, even though the dog remains visually prominent in the image.
Similar to the previous example, this selective pattern becomes clearer within the estimated workspace.

\input{cards/readout-answer-dissociation}

Example~\ref{ex:readout-answer-dissociation} provides another example of selective access, where we use a specific instruction to reduce access to a concept.
Specifically, we use the same cup/book image as Example~\ref{ex:query-object-selection} and compare a direct query with an instruction to ignore the book and name the other object.
The model answers ``cup'' under both queries.

Within the estimated workspace, \textit{book} has substantially lower readability under the ignore query, even though it is not included in either answer.
This shows that selective access can also affect concepts in intermediate states, even though they are not selected for output.

\FloatBarrier
\subsection{Selective Access during Autoregressive Generation}
\label{sec:generation-concept-selection}

We have examined selective access to visual concepts at the pre-answer position in the previous section.
Here, we extend the analysis to concept accessibility during autoregressive generation.
Specifically, we build two case studies of autoregressive rollout: identifying objects in spatial order and naming objects with or without skipping specified items.

\begin{figure}[tbp]
\begin{examplecard}{Concept Readability Changes during Answer Generation}
\exhibitlabel{ex:generation-concept-readout}
\noindent
\begin{minipage}[t]{0.20\linewidth}
\vspace{0pt}\centering
{\setlength{\fboxsep}{0pt}\fbox{\includegraphics[width=0.98\linewidth]{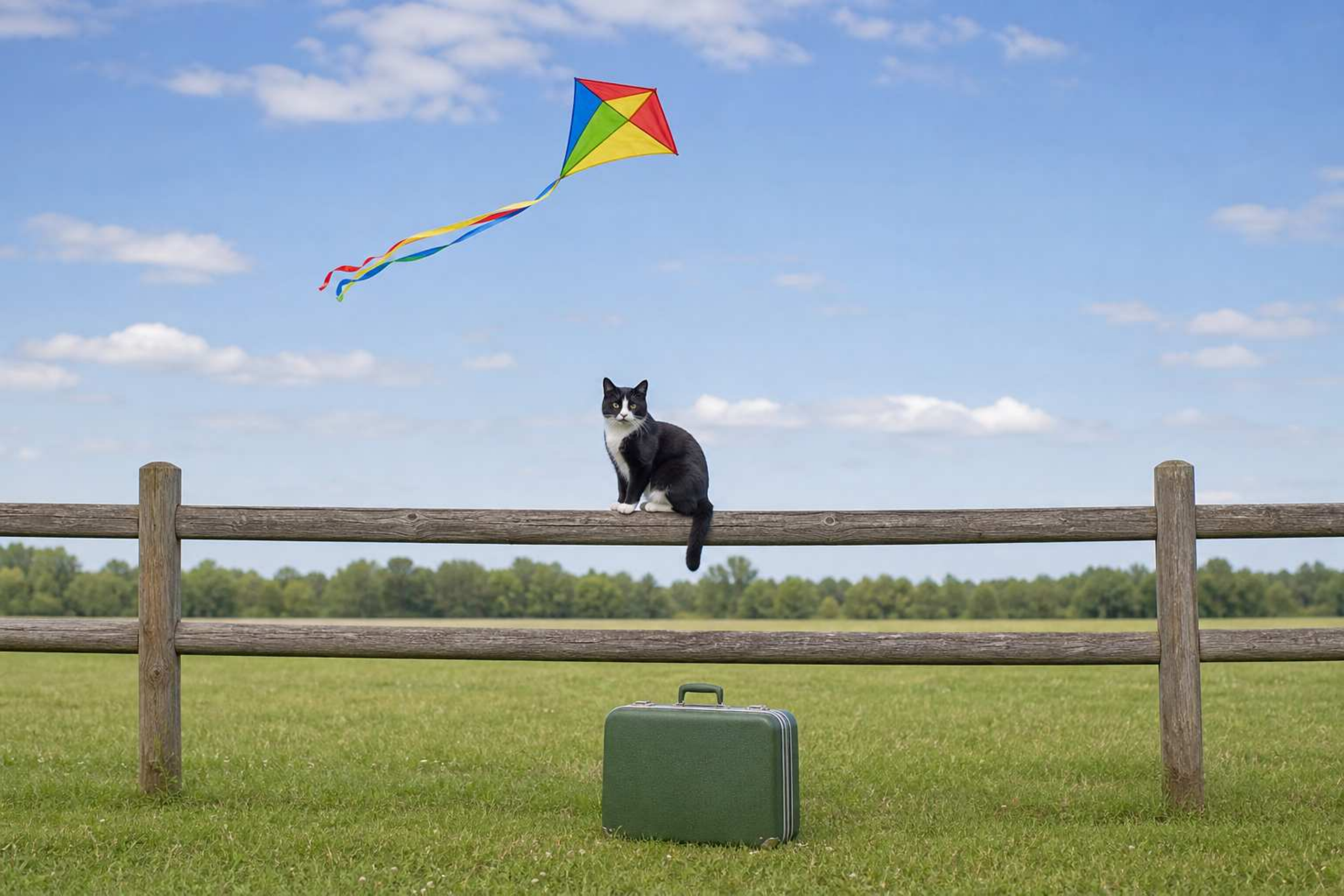}}}\par
\vspace{3pt}
{\footnotesize Input Image\par}
\end{minipage}\hfill
\begin{minipage}[t]{0.76\linewidth}
\vspace{0pt}\footnotesize\raggedright
\begin{tcolorbox}[colback=white,colframe=cardblue!55,boxrule=0.5pt,arc=2pt,
  left=6pt,right=6pt,top=5pt,bottom=5pt,before skip=0pt,after skip=0pt]
\textbf{Query:} ``Describe exactly the three main objects in this image, from \cardcue{top to bottom}, using one complete sentence.
Name each object once.
Do not use a list, numbering, or bullet points.''
\end{tcolorbox}
\vspace{5pt}
\textbf{Reading the heatmaps:} The ``vintage'' column shows the readout just before the model generates \cardcue{\texttt{\char32 suitcase}}.
\end{minipage}
\par\vspace{7pt}
\noindent\begin{tikzpicture}
\draw[black,line width=0.4pt,dash pattern=on 2.5pt off 2.5pt] (0,0) -- (\linewidth,0);
\end{tikzpicture}\par
\vspace{4pt}
\noindent\begin{minipage}[c]{0.21\linewidth}
{\footnotesize\bfseries L-Lens readout}
\end{minipage}\hfill
\begin{minipage}[c]{0.76\linewidth}
\includegraphics[width=\linewidth]{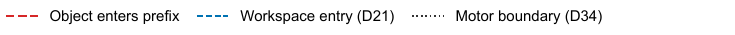}
\end{minipage}\par
\vspace{3pt}
\noindent\includegraphics[width=\linewidth]{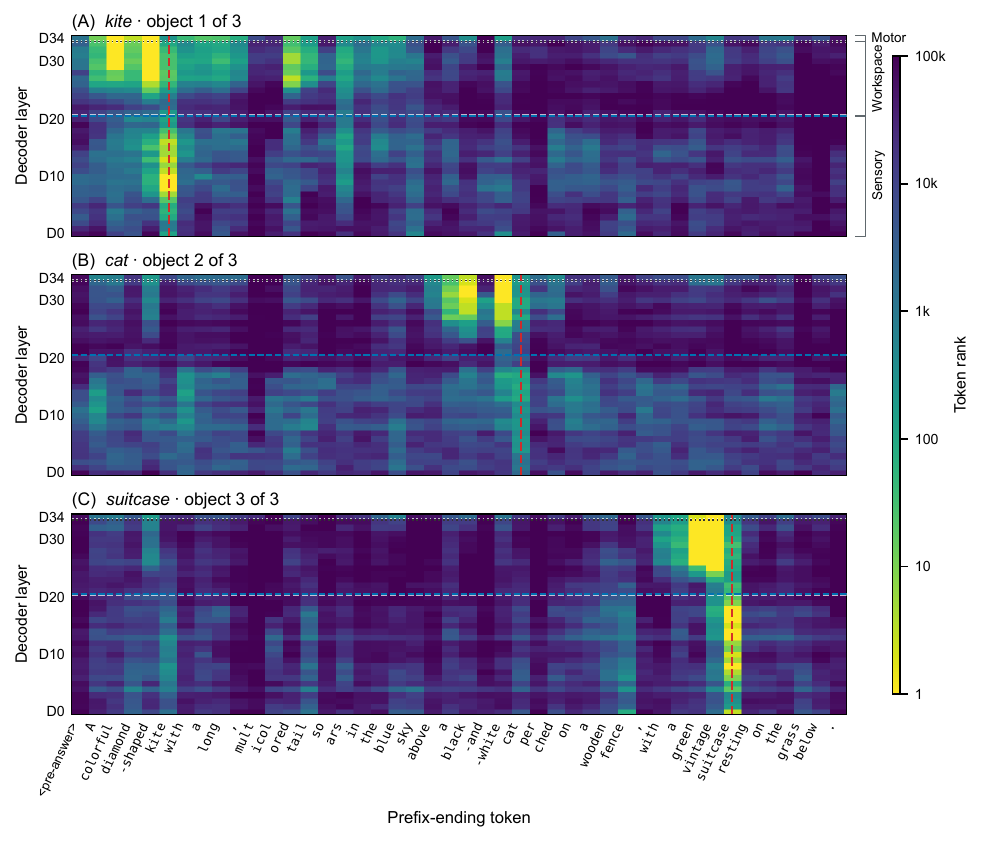}\par
\vspace{-4pt}
\noindent\begin{tikzpicture}
\draw[black,line width=0.4pt,dash pattern=on 2.5pt off 2.5pt] (0,0) -- (\linewidth,0);
\end{tikzpicture}\par
\vspace{3pt}
{\raggedright\footnotesize
\textbf{Finding:} Before an object's name is generated, its concept is usually more readable in the estimated workspace than in earlier decoder layers.
Immediately before generation, the name token reaches rank~1 in the motor layers.
Once the name enters the text prefix, the concept also becomes more readable in earlier layers.\par}
\end{examplecard}
\end{figure}

We first examine concept readability before and after an object's name is generated during autoregressive generation.
In Example~\ref{ex:generation-concept-readout}, the image contains a kite, a cat and a suitcase, and we instruct the model to describe these objects from top to bottom in a single sentence.

After the model generates its answer, we feed back successively longer prefixes of that answer, and use L-Lens to read the updated last token state at each decoder layer.
At each prefix, we read the same three target object-name tokens, i.e., \texttt{\char32 kite}, \texttt{\char32 cat} and \texttt{\char32 suitcase}.
Each heatmap presents the accessibility of one object-name token, where columns show the state after each prefix-ending token, rows show decoder layers, and colour indicates vocabulary rank.

We can find that the three observed target concepts show similar selective access patterns.
Before each object-name token is generated, it is usually more readable in the estimated workspace than in earlier decoder layers, and immediately before generation it reaches rank 1 in the output-preparation (motor) layers.
However, once a name enters the text prefix, it is also more readable in earlier decoder layers.
For example, \textit{suitcase} is already highly readable in the estimated workspace when the prefix ends in ``green'' and then ``vintage'', before the model names the object.
In contrast, after ``suitcase'' enters the text prefix, it becomes more readable in the pre-workspace layers.

This example is also consistent with the sensory, workspace and motor organisation discussed in the J-Lens study~\citep{gurnee2026verbalizable}.
In our example, a concept becomes more readable in earlier layers once its name enters the input, and in motor layers just before its name is generated.
We interpret its earlier accessibility within the workspace as consistent with the concept being available for subsequent computation.

\begin{figure}[tbp]
\begin{examplecard}{Concept Readability Changes with Output Instructions}
\exhibitlabel{ex:generation-order-selection}
\newcommand{\eNineSkipped}[1]{%
  \tikz[baseline=(word.base)]{%
    \node[inner sep=0pt,outer sep=0pt,text=black!60] (word) {\texttt{#1}};%
    \draw[black!70,line width=0.4pt]
      ($(word.base west)+(0,0.5ex)$) -- ($(word.base east)+(0,0.5ex)$);%
  }%
}
\noindent
\begin{minipage}[t]{0.20\linewidth}
\vspace{0pt}\centering
{\setlength{\fboxsep}{0pt}\fbox{\includegraphics[width=0.98\linewidth]{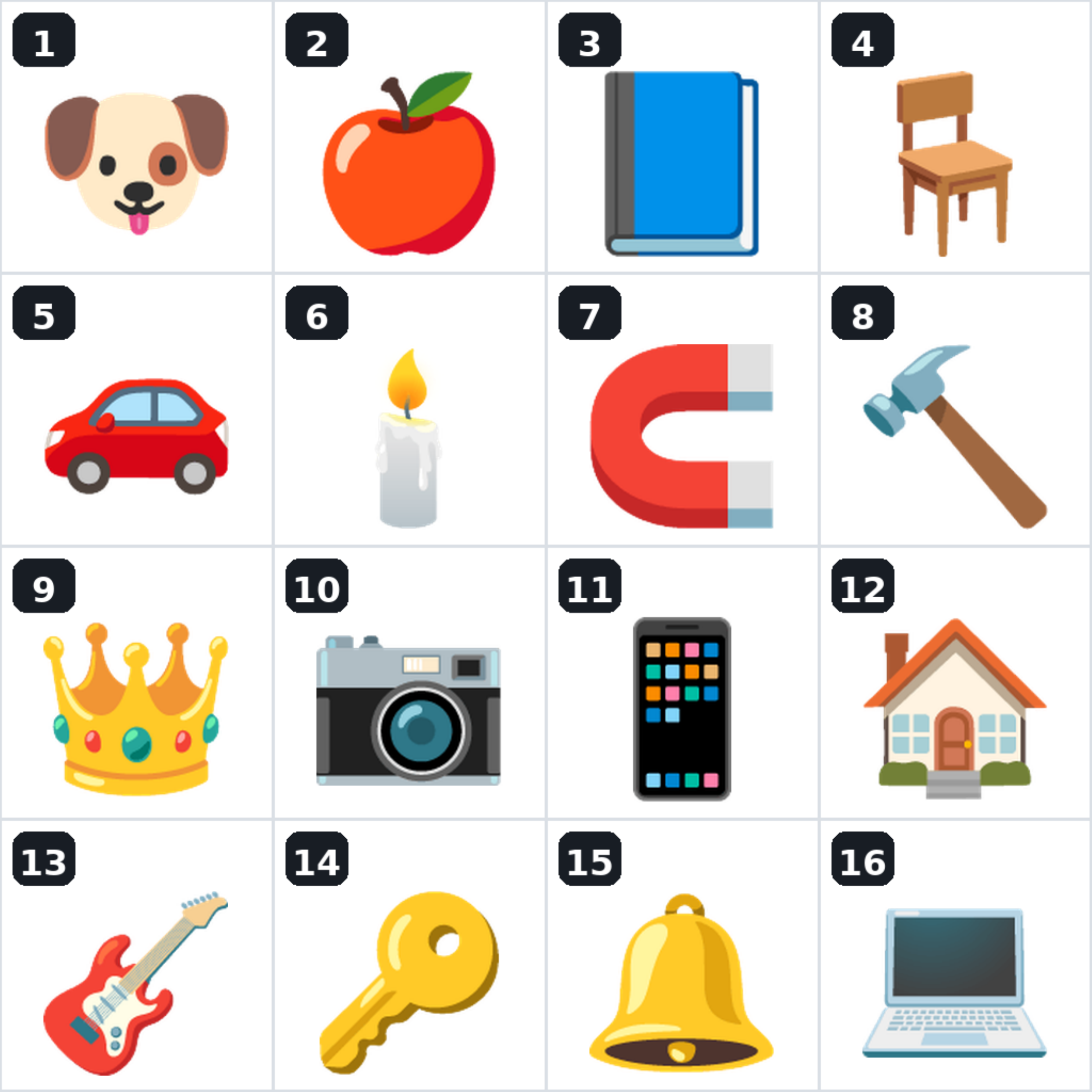}}}\par
\vspace{3pt}
{\footnotesize Input Image\par}
\end{minipage}\hfill
\begin{minipage}[t]{0.76\linewidth}
\vspace{0pt}\footnotesize\raggedright
\textbf{Task:} Name the objects in numerical order as a comma-separated list.\par
\vspace{5pt}
\textbf{Shared generated prefix:}\par
\texttt{list: dog, apple, book, chair, car, candle, magnet}\par
\vspace{3pt}
\textbf{Model continuation:}\par
\vspace{1pt}
\noindent
\begin{minipage}[t]{0.48\linewidth}
\begin{tcolorbox}[colback=white,colframe=cardblue!55,boxrule=0.5pt,arc=2pt,
  left=6pt,right=6pt,top=2.5pt,bottom=2.5pt,before skip=0pt,after skip=0pt]
\textbf{\strut Ascending order}\par\vspace{1pt}
\texttt{\strut hammer, crown, camera}\ldots
\end{tcolorbox}
\end{minipage}\hfill
\begin{minipage}[t]{0.48\linewidth}
\begin{tcolorbox}[colback=white,colframe=cardblue!55,boxrule=0.5pt,arc=2pt,
  left=6pt,right=6pt,top=2.5pt,bottom=2.5pt,before skip=0pt,after skip=0pt]
\textbf{\strut Skip 8 and 9}\par\vspace{1pt}
\strut\eNineSkipped{hammer}\texttt{, }\eNineSkipped{crown}\texttt{, camera}\ldots
\end{tcolorbox}
\end{minipage}
\end{minipage}
\par\vspace{7pt}
\noindent\begin{tikzpicture}
\draw[black,line width=0.4pt,dash pattern=on 2.5pt off 2.5pt] (0,0) -- (\linewidth,0);
\end{tikzpicture}\par
\vspace{4pt}
\noindent\begin{minipage}[c]{0.21\linewidth}
{\footnotesize\bfseries L-Lens readout}
\end{minipage}\hfill
\begin{minipage}[c]{0.74\linewidth}
\includegraphics[width=\linewidth]{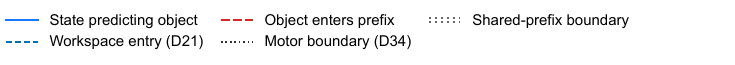}
\end{minipage}\par
\vspace{3pt}
\noindent\includegraphics[width=\linewidth]{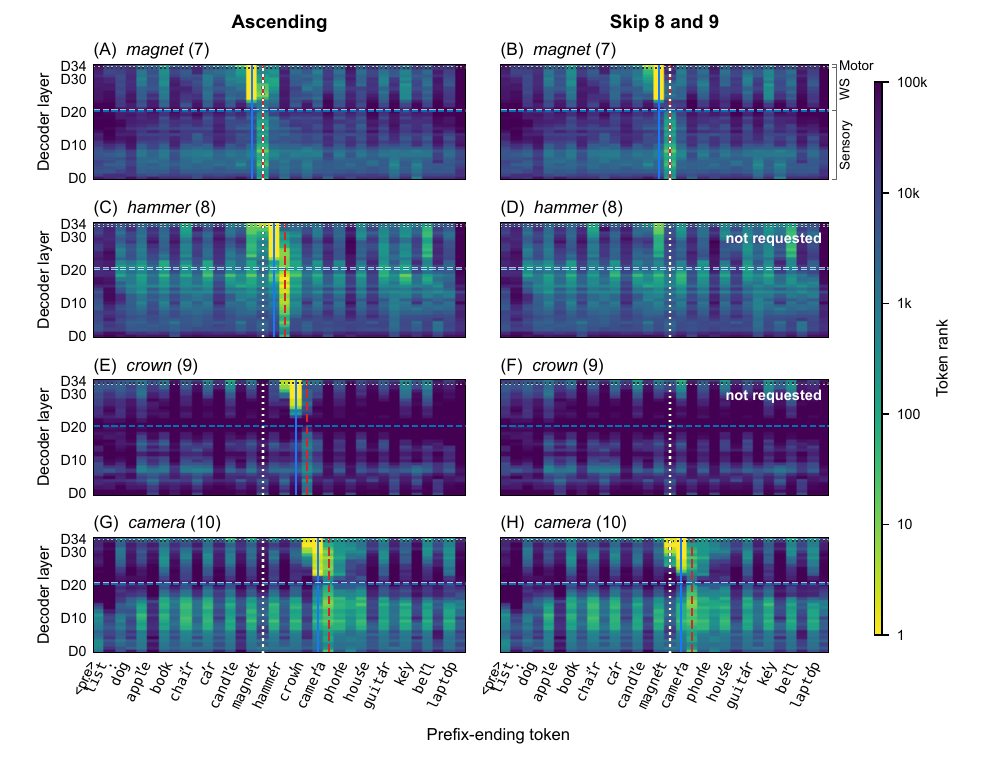}\par
\vspace{-4pt}
\noindent\begin{tikzpicture}
\draw[black,line width=0.4pt,dash pattern=on 2.5pt off 2.5pt] (0,0) -- (\linewidth,0);
\end{tikzpicture}\par
\vspace{3pt}
{\raggedright\footnotesize
\textbf{Finding:} Changing the output instructions changes concept accessibility within the estimated workspace, even when the image and the generated prefix are the same.\par}
\end{examplecard}
\end{figure}

We next examine the effect of the text instructions on concept accessibility during autoregressive generation.
In Example~\ref{ex:generation-order-selection}, we use an image with 16 objects in a specific order, and the query is to report these items following this order.
In a second condition, we ask the model to skip some of these objects, specifically ``hammer'' and ``crown''.
During autoregressive rollout, the generated prefixes are identical through the first seven object names, ending with ``magnet''.
And in the normal ascending run, the next required name is ``hammer'', while in the skip run it is ``camera''.

Similar to the previous example, concepts become more readable within the estimated workspace before their names are generated, and in earlier layers once their names enter the input.
This experiment shows that changing the text instructions also leads to the change in concept accessibility within the estimated workspace, even when the image and the generated prefix are the same.

\section{Causal Effects of Multimodal Concepts on Downstream Reasoning}
\label{sec:downstream-computation}

\begin{figure}[tbp]
    \centering
    \includegraphics[width=\linewidth]{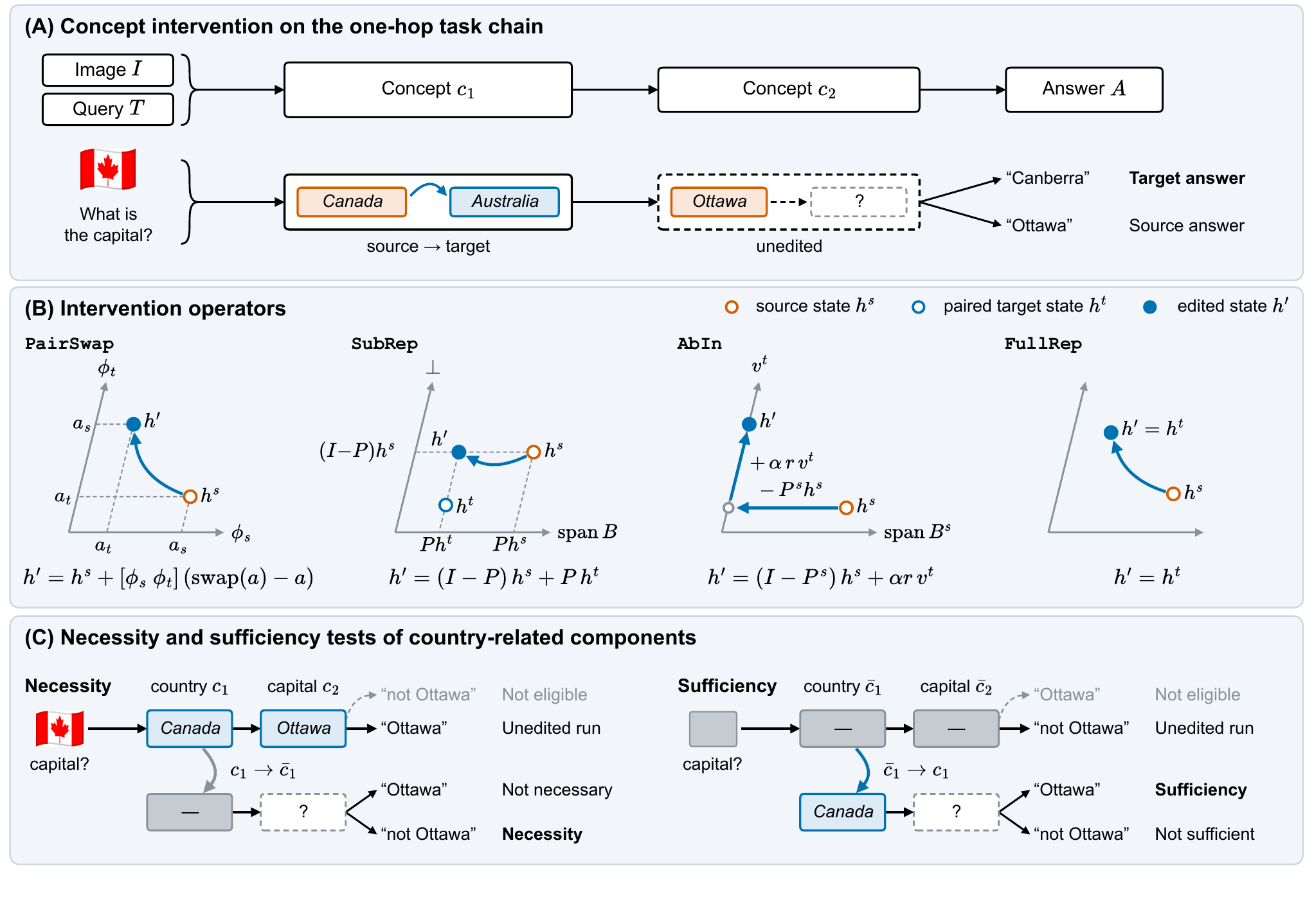}
    \caption{\textbf{Counterfactual interventions on concept representations.}
    \textbf{(A)} Replacing country-related components can change the answer to a capital question.
    \textbf{(B)} Four intervention operators and their state updates.
    \textbf{(C)} Necessity and sufficiency tests replace and restore components using original and grey-image states.}
    \label{fig:intervention-overview}
\end{figure}

The previous sections showed that concepts can be selectively accessed in intermediate states, but these observations do not explain how the model uses the information.
In GWT, the selected information in the shared workspace can be available to other specialised processes, which allows this information to influence more than one computation~\citep{baars2007architectural}.
In this section, we use the counterfactual interventions defined in Section~\ref{sec:concept-interventions} to examine the effects of concept-related state components on the final outputs of the model, as illustrated in Figure~\ref{fig:intervention-overview}.

\subsection{Counterfactual Concept Interventions Can Redirect Downstream Reasoning}
\label{sec:counterfactual-interventions}

In this experiment, we examine the effects of concept-related state components on model answers under the settings of direct and one-hop VQA tasks.
In this setting, the concept \(c_1\) is identified from the image, and \(c_2\) is an associated concept, as illustrated in Figure~\ref{fig:intervention-overview}~(A).
We intervene on the state components associated with \(c_1\) and compare a question about \(c_1\) (direct intervention) with a question about \(c_2\) (one-hop intervention).

We use the counterfactual interventions defined with \(\operatorname{do}\) in Section~\ref{sec:concept-interventions}.
The name-token set \(F_0\), cumulative word-form sets \(F_{\leq j}\) and greedy sets \(\greedyset{k}\) specify which concept directions are used for intervention.
We combine these sets with \texttt{SubRep} or \texttt{AbIn}, and include \(F_0+\texttt{PairSwap}\) as a baseline~\citep{gurnee2026verbalizable}.
These operators specify how the state is changed, as illustrated in Figure~\ref{fig:intervention-overview}~(B).

In Example~\ref{ex:answer-interventions}, \(c_1\) is the country represented by the national flag, and \(c_2\) is its capital.
We keep the Canadian flag fixed and change the country-related components from \textit{Canada} to \textit{Australia} under the direct and one-hop questions.
Both \(F_0+\texttt{PairSwap}\) and \(\greedyset{25}+\texttt{SubRep}\) change the country answer to ``Australia''.
For the capital question, however, \(F_0+\texttt{PairSwap}\) leaves the answer at ``Ottawa'', whereas \(\greedyset{25}+\texttt{SubRep}\) changes it to ``Canberra''.
Both interventions can therefore influence the model's answer in the direct task.
In the one-hop task, the \(\greedyset{25}+\texttt{SubRep}\) intervention also redirects the capital answer, showing that country-related information can influence downstream computation.

\input{cards/answer-interventions}

For the quantitative analysis, we build source-to-target pairs using the 50 national-flag images from Noto Color Emoji and follow the same intervention procedure as in Example~\ref{ex:answer-interventions}.
We use the same country and capital question settings and intervene on country-related components in Qwen3-VL-8B, editing all decoder layers in order of depth (Appendix~\ref{app:intervention-depth-windows} examines intervention depth and window width).
We compare the name-token set \(F_0\) with broader word-form sets and greedy sets with different direction budgets.
Each direction set is used with \texttt{SubRep} and \texttt{AbIn}, alongside the \(F_0+\texttt{PairSwap}\) baseline.
For the one-hop task, capital-answer tokens are excluded during greedy selection.

We measure target-answer rate, the fraction of eligible pairs producing a complete target answer, and higher values indicate more successful redirection.
We also compare the median ranks of the target and source answer tokens before and after intervention, measured at the final layer before generation.

\begin{figure}[tbp]
    \centering
    \includegraphics[width=\linewidth]{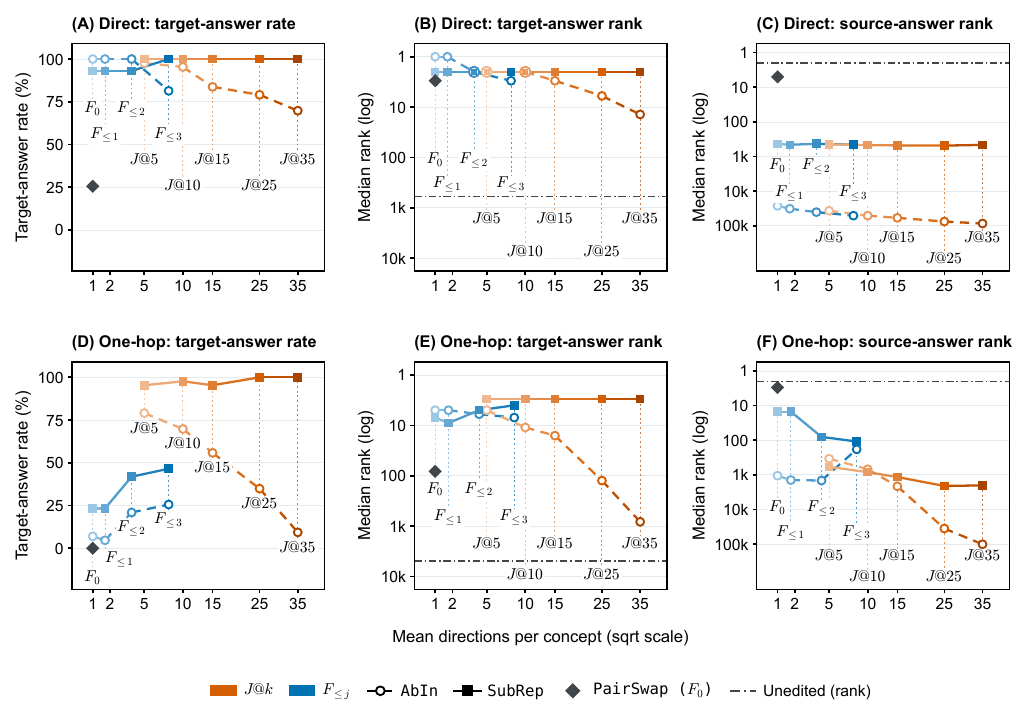}
    \caption{\textbf{Concept interventions change answers and vocabulary readouts.}
    Rows show direct (country) and one-hop (capital) questions.
    Columns show target-answer rate and the median ranks of the target and source answer tokens.}
    \label{fig:intervention-direction-operator}
\end{figure}

For both direct and one-hop tasks, different intervention operators and direction sets lead to diverse effects on model answers.
As shown in Figure~\ref{fig:intervention-direction-operator}~(A), direct interventions with \texttt{SubRep} redirect nearly every answer across all tested direction sets.
The name-token set \(F_0\) therefore already provides strong control in the direct task, and expanding this set brings only limited improvement.
In contrast, the success of one-hop interventions varies much more with the direction set, as shown in Figure~\ref{fig:intervention-direction-operator}~(D).
With \texttt{SubRep} fixed, broader word-form sets \(F_{\leq j}\) improve one-hop intervention success, but the greedy sets \(\greedyset{k}\) are more effective.
Even the smallest greedy set, \(\greedyset{5}\), redirects nearly every one-hop answer, whereas the broadest word-form set, \(F_{\leq3}\), redirects fewer than half.
\(\greedyset{5}\) gives stronger one-hop control with fewer directions per concept, using five compared with roughly eight for \(F_{\leq3}\).

We also find that answer control depends on the intervention operator.
With the same name-token set \(F_0\), \texttt{SubRep} and \texttt{AbIn} redirect more direct answers than \texttt{PairSwap}, as shown in Figure~\ref{fig:intervention-direction-operator}~(A).
\texttt{SubRep} maintains high success across the greedy budgets in both direct and one-hop tasks, while \texttt{AbIn}'s success falls as the set expands (Figure~\ref{fig:intervention-direction-operator}~(A,D)).
The vocabulary ranks help explain these different answer outcomes.
For one-hop interventions, target answers remain near the top of the vocabulary with \texttt{SubRep}, but become less readable as the greedy budget increases with \texttt{AbIn}, as shown in Figure~\ref{fig:intervention-direction-operator}~(E).
At the same time, \texttt{AbIn} suppresses source answers more strongly in Figure~\ref{fig:intervention-direction-operator}~(F).
Suppressing the source is therefore insufficient for reliable redirection when the target also loses accessibility.

Generally speaking, these experiments show that concept information identified through MC-Lens can contribute to downstream computation in both direct and one-hop tasks.
The effects depend on both the selected concept directions and the intervention operator.

\subsection{Necessity and Sufficiency Analysis of Concept Representations in Reasoning}
\label{sec:necessity-sufficiency}

Previous counterfactual experiments show that concepts read from intermediate states can influence downstream computation and model answers.
We extend the analysis to the necessity and sufficiency of these selected intermediate concept components for downstream tasks~\citep{pearl2009causality}.

As illustrated in Figure~\ref{fig:intervention-overview}~(C), necessity tests whether replacing the selected components disrupts the original answer, inspired by the counterfactual tests of \citet{yu2025causal}.
Here, we select a VQA example that the model answers correctly with the original image, and construct a control run with a uniform grey image under the same question.
We retain examples with an incorrect grey-image answer.
Let \(h_\ell\) and \(\bar h_\ell\) denote the states of the unedited original-image and grey-image runs, respectively.
Here, \(P_{\ell,c}h_\ell\) is the component along the selected directions for concept \(c\).
We replace the selected components of \(h_\ell\) in the original-image run using the \texttt{SubRep} intervention \(\operatorname{do}[P_{\ell,c}h_\ell=P_{\ell,c}\bar h_\ell,\ \ell\in\mathcal L]\).
If the resulting answer becomes incorrect, this indicates that the selected components were necessary for the original answer.

Sufficiency tests whether restoring the selected components recovers a correct answer, similar to the restoration step of causal tracing~\citep{meng2022locating,palit2023blip}.
For this test, we use the same original-image and grey-image runs.
We replace the selected components of \(\bar h_\ell\) in the grey-image run with those of \(h_\ell\) from the original-image run using the \texttt{SubRep} intervention \(\operatorname{do}[P_{\ell,c}\bar h_\ell=P_{\ell,c}h_\ell,\ \ell\in\mathcal L]\).
If the answer becomes correct, the selected components are sufficient to recover the answer in the grey-image run.

For a VQA example, let \(I\) and \(\bar I\) denote the original-image and grey-image inputs under the same question.
We use \(\mathrm{Y}(X)\in\{0,1\}\) to indicate whether the unedited model answers correctly under input \(X\), with 1 for a correct answer and 0 otherwise.
We write \(\mathrm{Y}(X;\operatorname{do}[\cdot])\) for the correctness of the answer after the specified intervention.
For eligible examples, \(\mathrm{Y}(I)=1\) and \(\mathrm{Y}(\bar I)=0\).

The per-example necessity and sufficiency outcomes are
\begin{align}
    N
    &=\mathrm{Y}(I)\left[1-\mathrm{Y}\!\left(I;\operatorname{do}[P_{\ell,c}h_\ell=P_{\ell,c}\bar h_\ell]\right)\right],
    \label{eq:necessity-outcome}\\[4pt]
    S
    &=\left[1-\mathrm{Y}(\bar I)\right]\mathrm{Y}\!\left(\bar I;\operatorname{do}[P_{\ell,c}\bar h_\ell=P_{\ell,c}h_\ell]\right),
    \label{eq:sufficiency-outcome}
\end{align}
where both interventions apply to \(\ell\in\mathcal L\), as described above.
Here, \(N=1\) and \(S=1\) indicate that the example passes the necessity and sufficiency tests, respectively.

We denote the common eligible set by \(\mathcal E\) and its size by \(|\mathcal E|\), and let \(N_i\) and \(S_i\) denote the outcomes for example \(i\).
We calculate the necessity rate (NR), sufficiency rate (SR) and joint necessity and sufficiency rate (NSR) over this set as
\begin{equation}
    \mathrm{NR}=\frac{1}{|\mathcal E|}\sum_{i\in\mathcal E}N_i,
    \qquad
    \mathrm{SR}=\frac{1}{|\mathcal E|}\sum_{i\in\mathcal E}S_i,
    \qquad
    \mathrm{NSR}=\frac{1}{|\mathcal E|}\sum_{i\in\mathcal E}N_iS_i.
    \label{eq:necessity-sufficiency-rates}
\end{equation}

Higher NR indicates that downstream reasoning depends more consistently on the accessed concept components.
Higher SR indicates that these components are more consistently sufficient to recover the correct answer.
Higher NSR indicates that the components are more consistently both necessary and sufficient.

For the quantitative analysis, we use the same 50 national-flag images from Noto Color Emoji and test Qwen3-VL-8B with the capital question fixed.
We apply \texttt{SubRep} using the name-token set \(F_0\), broader word-form sets \(F_{\leq j}\) and greedy sets \(\greedyset{k}\). For each direction set, we compare direct interventions on capital-related components with one-hop interventions on country-related components.

\begin{figure}[tbp]
    \centering
    \includegraphics[width=\linewidth]{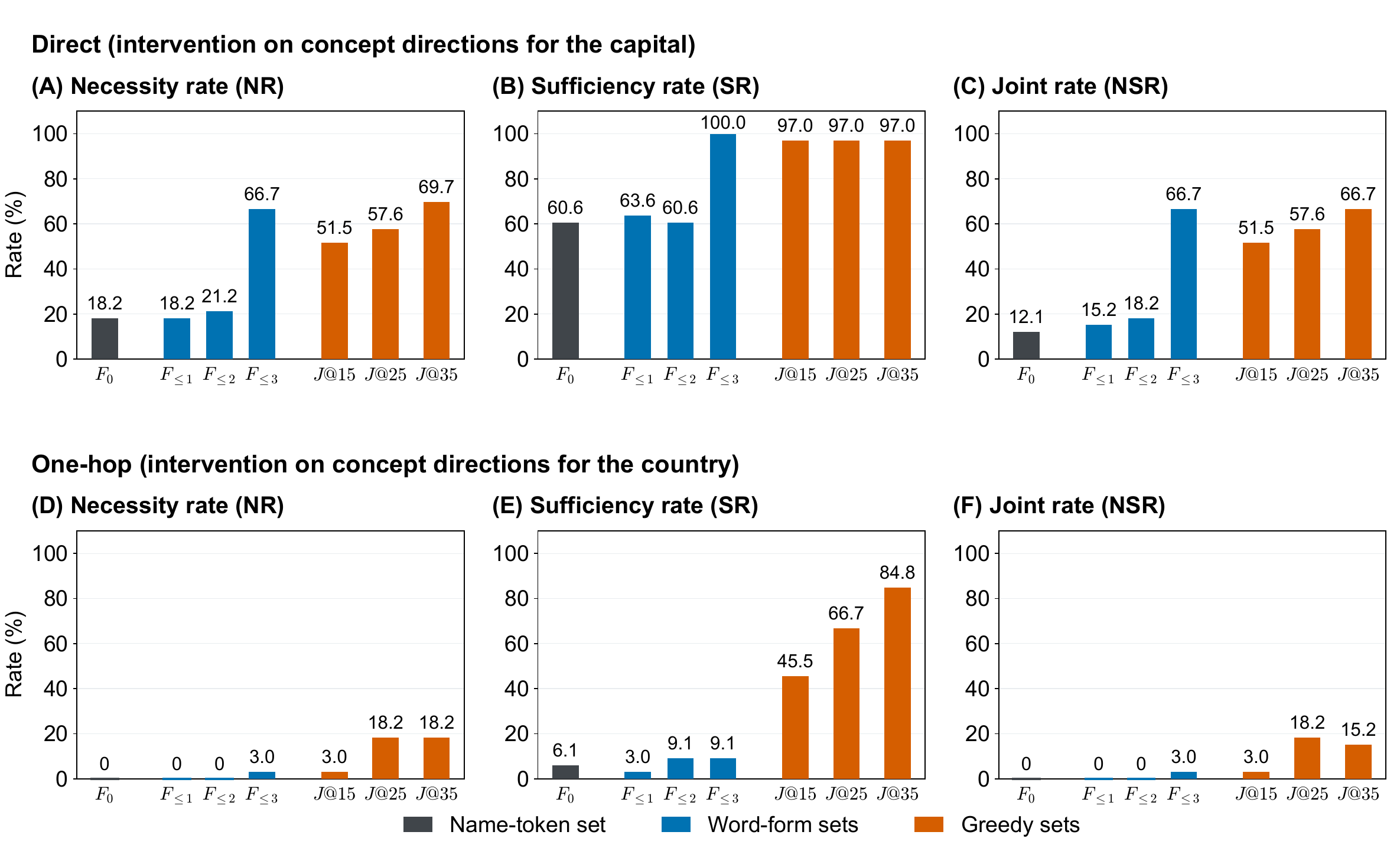}
    \caption{\textbf{Necessity and sufficiency of concept-related components.}
    Capital answers are tested using \texttt{SubRep}, with concept directions for the capital (direct, top) or the country (one-hop, bottom).
    Columns show NR, SR and NSR.}
    \label{fig:necessity-sufficiency}
\end{figure}

As illustrated in Figure~\ref{fig:necessity-sufficiency}~(A,D), NR is higher for direct interventions than for one-hop interventions across all tested direction sets.
For direct interventions, NR increases with the budget \(k\) of the greedy sets \(\greedyset{k}\), while broadening the word-form sets \(F_{\leq j}\) produces a marked increase at \(F_{\leq3}\).
All three greedy sets (\(\greedyset{15}\), \(\greedyset{25}\) and \(\greedyset{35}\)) recover nearly every answer, as shown by SR in Figure~\ref{fig:necessity-sufficiency}~(B).
The broadest word-form set, \(F_{\leq3}\), also recovers every answer.
For direct interventions using broader direction sets, NSR is close to NR, as shown in Figure~\ref{fig:necessity-sufficiency}~(C).
This means that, in most examples where the selected components are necessary for the original answer, they are also sufficient to recover the answer in the grey-image run.

For one-hop interventions, NR remains close to zero with the word-form sets \(F_{\leq j}\), and even the broader sets rarely recover the answer, as shown in Figure~\ref{fig:necessity-sufficiency}~(D,E).
The greedy sets \(\greedyset{k}\) are much more effective at recovering answers, and recovery improves markedly with the greedy budget.
However, even \(\greedyset{25}\) and \(\greedyset{35}\) disrupt only a small minority of the original answers.
For example, one-hop \(\greedyset{35}\) interventions recover about 85\% of answers, but the selected components are both necessary and sufficient in only 15\% of examples, as shown in Figure~\ref{fig:necessity-sufficiency}~(E,F).

Generally speaking, the selected concept-related components can support downstream computation.
As the reasoning chain extends from direct to one-hop, the selected components show lower NR and SR, even with broader direction sets.
For one-hop tasks, most recovered answers do not also pass the necessity test, suggesting that other components may also support the original answers.

\subsection{Causal Effects of Concepts across Various Downstream Reasoning Tasks}
\label{sec:cross-task-reuse}

In this part, we examine how an accessed concept can serve as shared intermediate information for different downstream tasks.
In a one-hop VQA task (\(I \rightarrow c_1 \rightarrow c_2 \rightarrow A\)), we keep the input image \(I\) and the initial concept \(c_1\) fixed, and vary the downstream question to ask about different concepts \(c_2\) associated with \(c_1\).
Here, \(A\) is the model's answer.

\input{cards/cross-task-reuse}

Specifically, as Example~\ref{ex:cross-task-reuse} illustrates, we extend the downstream task from identifying the capital to questions about the three-letter country code (ISO-3), currency code and other associated information.
We then change the country-related components from \textit{Ireland} to \textit{Tunisia} using \(\greedyset{25}+\texttt{SubRep}\) on Qwen3-VL-8B, and observe the changes in the model's answers across these tasks.
We find that the capital answer changes from ``Dublin'' to ``Tunis'', and the currency code changes from ``EUR'' to ``TND''.
Four of the six answers change to their targets, but the domain and calling-code answers remain non-target outputs.

For the quantitative analysis, we build 51 eligible source-to-target pairs using the same 50 national-flag images from Noto Color Emoji, with tasks covering the country name, ISO-3 code, capital, currency code, continent, internet domain and calling code.
We apply \texttt{SubRep} in all decoder layers of Qwen3-VL-8B, using the name-token set \(F_0\), broader word-form sets \(F_{\leq j}\) and greedy sets \(\greedyset{k}\), with \texttt{FullRep} as a reference.
Downstream-answer tokens are excluded during greedy selection.
We measure target-answer rate to evaluate intervention success, as in Section~\ref{sec:counterfactual-interventions}.

\Needspace{24\baselineskip}
\begin{wrapfigure}{r}{0.48\linewidth}
\centering
\includegraphics[width=\linewidth]{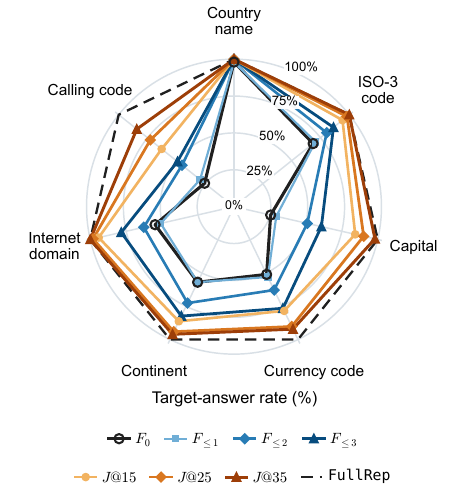}
\captionsetup{type=figure,font=small,justification=raggedright,singlelinecheck=false}
\caption{\textbf{The same concept directions influence answers across tasks.}
    We compare target-answer rates for \texttt{SubRep} using different concept direction sets, with \texttt{FullRep} as a control.}
\label{fig:cross-task-reuse}
\end{wrapfigure}

As illustrated in Figure~\ref{fig:cross-task-reuse}, direct interventions succeed in nearly every case for all direction sets, whereas the success of one-hop interventions varies more.
For the one-hop tasks, \(F_0\) and \(F_{\leq1}\) give similar target-answer rates, while the broader word-form sets improve control.
All three greedy sets give higher target-answer rates than the word-form sets on these tasks.

The larger greedy sets redirect most answers across the one-hop tasks, with further improvements as the greedy budget increases.
Calling codes remain less reliably redirected even with the largest greedy set, while \texttt{FullRep} reaches every target on every task.

Generally speaking, these experiments show that the same concept directions can influence answers across different downstream tasks, providing functional evidence for the broad availability of concept information described by GWT~\citep{baars2007architectural}.
The stronger control obtained with broader direction sets suggests that they capture more of the concept-related information used across these tasks.
\finishwrap

\subsection{Causal Effects of Concepts on Multi-Hop Reasoning}
\label{sec:multi-hop-interventions}
In this part, we further extend the direct and one-hop comparisons in Section~\ref{sec:counterfactual-interventions} to multi-hop tasks.
In these questions, the input identifies an initial concept \(c_1\), which is linked through an intermediate concept \(c_2\) to a final concept \(c_3\) asked about in the question.
We compare interventions on \(c_1\)-related components alone with joint interventions on \(c_1\)- and \(c_2\)-related components, and measure the resulting changes in the final answer \(A\).

\input{cards/multi-hop-interventions}

Example~\ref{ex:multi-hop-interventions-capital-province} shows an intervention for a question about the administrative division containing the capital of the country shown by a flag.
The source relation links the country \emph{Canada} (\(c_1\)) to its capital \emph{Ottawa} (\(c_2\)) and then \emph{Ontario} (\(c_3\)), while the target relation links \emph{Greece} to \emph{Athens} and then \emph{Attica}.
With the Canadian flag and question unchanged, a \(\greedyset{25}+\texttt{SubRep}\) intervention on country-related components changes the answer from ``Ottawa, Ontario'' to ``Athens, Attica''.

Example~\ref{ex:multi-hop-interventions-capital-airport} shows an unsuccessful case, where we apply the same type of intervention to a question about the International Air Transport Association (IATA) code of an international passenger airport serving the capital.
With country-related components changed from \emph{Denmark} to \emph{Jamaica}, the answer changes from ``CPH'' to ``SIN'', missing the target airport code ``KIN''.

For the quantitative analysis, we use 40 eligible multi-hop cases from the same national-flag image set, covering six types of questions about the capital.
These questions concern administrative divisions, rivers, airport codes, railway stations, national legislatures and metro opening years.
We follow the same intervention procedure as in Example~\ref{ex:multi-hop-interventions} and compare \(F_0\), \(F_{\leq2}\) and \(\greedyset{25}\) via \texttt{SubRep} under the two intervention conditions (\(c_1\) and \(c_1+c_2\)).
We use the concept directions for \(c_1\) alone or the combined direction sets for \(c_1\) and \(c_2\) to replace the corresponding state components.
Final-answer tokens are excluded during greedy selection.
We include \texttt{FullRep} as a control and use the same eligible cases for all comparisons.

\begin{figure}[t]
    \centering
    \includegraphics[width=\linewidth]{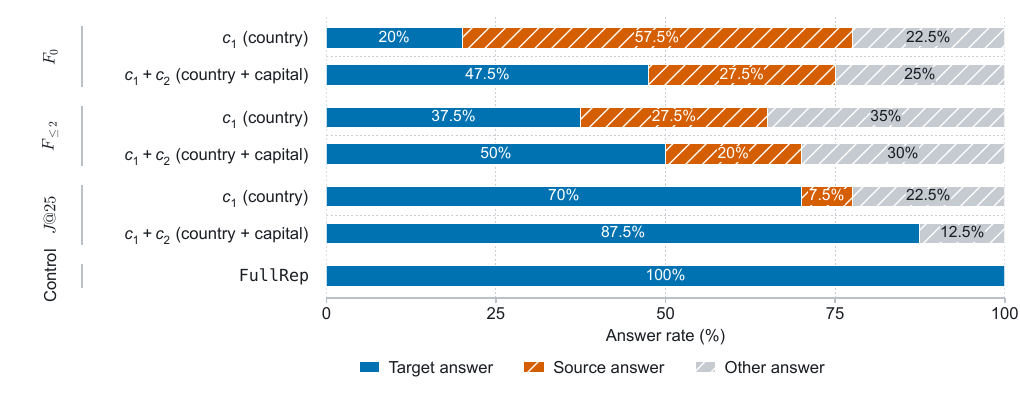}
    \caption{\textbf{Country and capital interventions in multi-hop questions.}
    We compare final-answer outcomes between country-only interventions (\(c_1\)) and combined country and capital interventions (\(c_1+c_2\)) using \texttt{SubRep}, with \texttt{FullRep} as a control.}
    \label{fig:multi-hop-composition}
\end{figure}

Figure~\ref{fig:multi-hop-composition} reports the proportions of target final answers, source final answers and other answers, based on the complete generated answers.
The experiment shows that multi-hop interventions on \(c_1\) alone can redirect the final answer.
Broadening the word-form set from \(F_0\) to \(F_{\leq2}\) increases the share of target final answers, but \(\greedyset{25}\) is more effective than either word-form set and leaves fewer answers at the source fact.

In addition, joint interventions (\(c_1+c_2\)) increase the target-answer rate for every direction set compared with interventions on \(c_1\) alone.
As a control, \texttt{FullRep} replaces the whole internal state and reaches every target answer.
In these multi-hop tasks, interventions on \(c_1\)-related components can therefore redirect the final answer, and additional interventions on \(c_2\)-related components improve control.

To sum up, our experiments show that the concept information accessed through MC-Lens is not merely shown on a ``display board'', but also participates in downstream computation and reasoning.
These effects across direct, one-hop and multi-hop tasks provide functional evidence for the downstream availability of concept information described by GWT~\citep{baars2007architectural}.

\section[Global Workspace Structure and Function across VLMs]{Global Workspace Structure and Function across VLMs}
\label{sec:workspace-organisation}

Previous sections have shown that multimodal concepts can be read from intermediate states, selectively accessed in response to queries, and used in downstream computations, providing evidence for the functions described by GWT~\citep{baars2005workspace}.
In this section, we investigate the structural and functional organisation of global workspaces in VLMs.

\subsection{Shared Workspace Structure across Model Families}
\label{sec:layerwise-organisation}

In this part, we investigate shared global workspace structures by comparing J-space geometry and prediction agreement across Qwen3-VL (2B, 8B, and 32B)~\citep{bai2025qwen3vl}, DeepSeek-VL2 (Tiny, Small, and full)~\citep{wu2024deepseekvl2}, and Gemma~3 (4B, 12B, and 27B)~\citep{gemmateam2025gemma3}.
Vocabulary-direction geometry and effective dimensionality describe the organisation of the directions used for concept readout at each layer.
Prediction agreement measures whether the model's final predicted token is already readable from an intermediate state.

\begin{figure}[p]
    \centering
    \includegraphics[width=\linewidth,height=0.82\textheight,keepaspectratio]{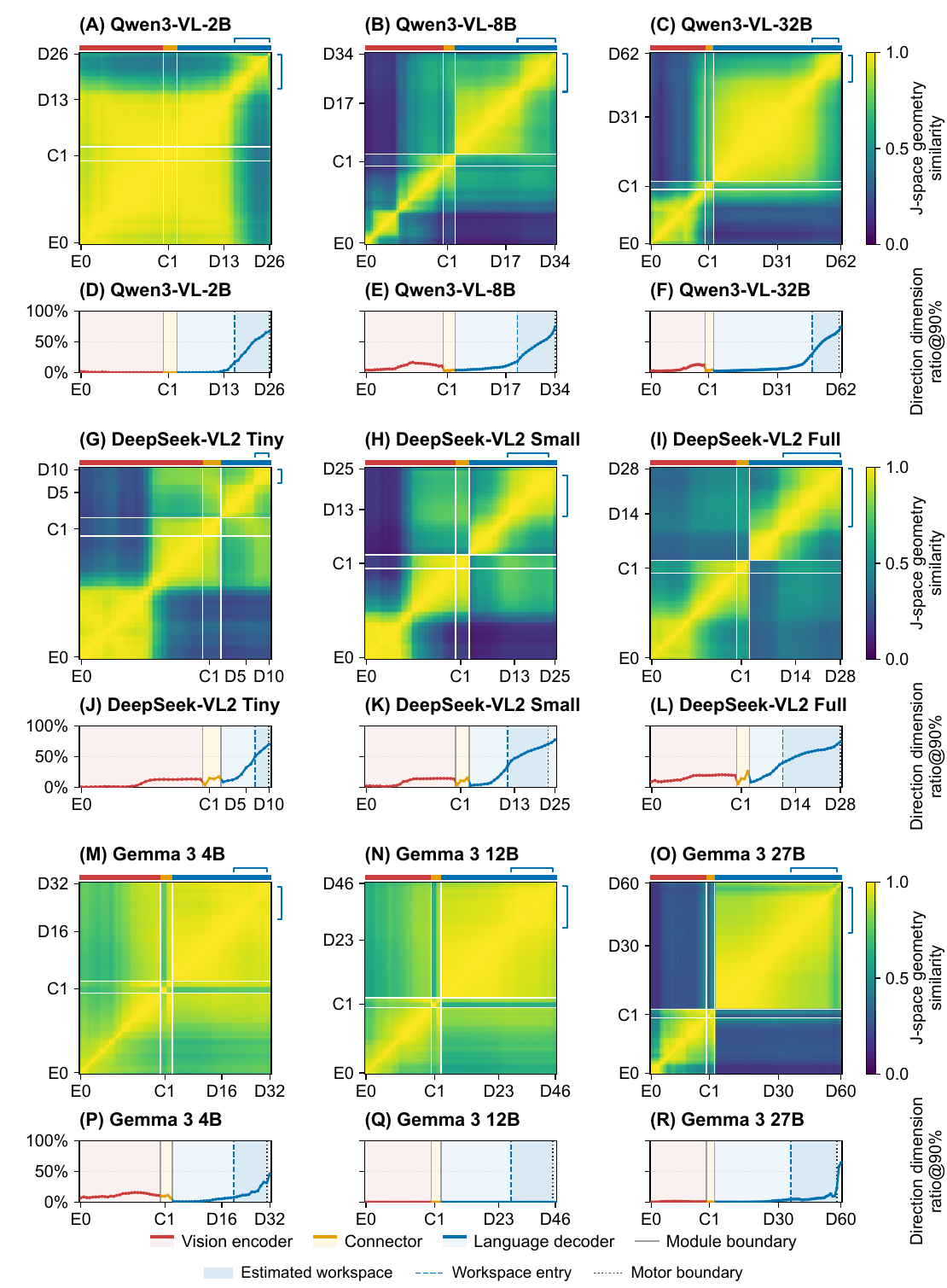}
    \caption{\textbf{J-space geometry across nine VLMs.}
    Heatmaps compare vocabulary-direction geometry across layers and components using CKA.
    Curves show the fraction of state dimensions needed to retain 90\% of the energy in the centred vocabulary directions.
    Blue brackets and darker blue shading mark the estimated workspace.}
    \label{fig:cross-model-direction-geometry}
\end{figure}

\paragraph{J-space geometry.}

We first compare the layerwise organisation of vocabulary directions across models, following the J-Lens analysis~\citep{gurnee2026verbalizable}.
We use linear centred kernel alignment (CKA)~\citep{kornblith2019similarity} to measure how similar the relationships among these directions are between layers.
Higher CKA appears as brighter blocks and indicates more similar direction relationships.
We also evaluate the spread of vocabulary directions using Direction dimension ratio@90\%.
This is the fraction of state dimensions needed to retain 90\% of the energy in the centred vocabulary directions.
A higher ratio indicates that these directions are spread over more state dimensions.

Among these models, we also estimate the workspace intervals marked in the figures from direction geometry and dimensionality, together with the layerwise readout profiles (Appendix~\ref{app:workspace-boundaries}).

In Qwen and DeepSeek, early decoder layers have relatively similar direction geometries.
At later layers, their geometries become more differentiated, as seen in the decoder blocks of \mbox{Figure~\ref{fig:cross-model-direction-geometry}~(A-C,G-I)}.
Gemma retains high similarity across much more of the decoder, as shown by the CKA matrices in Figure~\ref{fig:cross-model-direction-geometry}~(M-O).
The separation between depth regions resembles the sensory, workspace and motor organisation described by \citet{gurnee2026verbalizable}.
These regions are associated with input processing, the selection and use of intermediate concepts, and preparation of the output, respectively.
Appendix~\ref{app:lens-fit-size-comparison} shows that the broad layerwise organisation in Qwen3-VL-8B is preserved across fitting-set sizes.

In Qwen and DeepSeek, the dimension-ratio curves in Figure~\ref{fig:cross-model-direction-geometry}~(D-F,J-L) rise more sharply around the estimated workspace entry.
The start of this rise indicates the transition into the estimated workspace more clearly than the broad regions of high similarity in the CKA matrices.
As processing moves through this interval, vocabulary directions spread over more state dimensions.
In Gemma, the dimension ratios remain low through much of the decoder: 4B and 27B rise mainly near the output, while 12B remains low throughout, as shown in Figure~\ref{fig:cross-model-direction-geometry}~(P-R).

Across models, encoder dimension ratios remain relatively low.
The vision encoders differ in their layerwise direction geometry, both across model families and within the same family.
In some encoders, a boundary in the CKA matrix is accompanied by an increase in the dimension ratio, as in DeepSeek-VL2 Small (Figure~\ref{fig:cross-model-direction-geometry}~(H,K)).
In others, such as Gemma~3 12B, the CKA matrix shows less distinct layer separation and the dimension ratio stays low (Figure~\ref{fig:cross-model-direction-geometry}~(N,Q)).

\paragraph{Prediction agreement.}

Direction geometry describes the fitted lens, but does not tell us when the eventual output token becomes accessible for a particular input.
We therefore compare the pre-answer readout at each decoder layer with the model's own final prediction on 100 image-question inputs from A-OKVQA~\citep{schwenk2022aokvqa}, following the layerwise agreement analysis of the J-Lens study~\citep{gurnee2026verbalizable}.
For each input, we take the model's highest-scoring token at the final layer and check whether it appears among the top eight tokens read out from each intermediate layer.
Agreement@8 is the percentage of inputs that satisfy this condition at a given layer.
Higher values indicate that the final predicted token is more consistently accessible from the intermediate readout.

\begin{figure}[tbp]
    \centering
    \includegraphics[width=\linewidth]{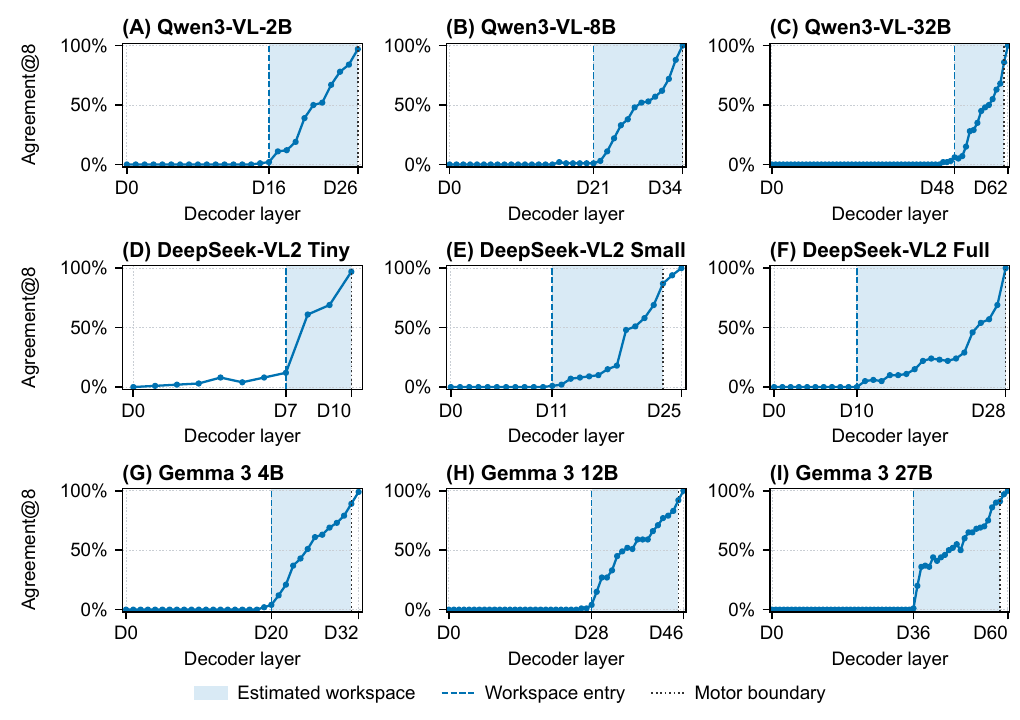}
    \caption{\textbf{Prediction agreement across nine VLMs.}
    Agreement@8 is the percentage of inputs for which the model's highest-scoring token at the final layer appears among the top eight tokens in the intermediate readout.}
    \label{fig:cross-model-prediction-agreement}
\end{figure}

The Agreement@8 curves in Figure~\ref{fig:cross-model-prediction-agreement} remain low at early depths and rise mainly within the estimated workspace intervals across all nine models.
Gemma also shows a sharp rise in Agreement@8, although its decoder direction geometry is less differentiated (Figures~\ref{fig:cross-model-direction-geometry}~(M-O) and~\ref{fig:cross-model-prediction-agreement}~(G-I)).
This highlights the need to consider multiple indicators when estimating workspace boundaries.

Across these models, the final predicted token becomes more readable within the estimated workspace, often alongside changes in direction geometry.

\subsection{Shared Workspace Functions across Model Families}
\label{sec:cross-model-workspace-functions}

In this part, we investigate shared global workspace functions across VLMs with different layerwise organisations.
We compare concept accessibility, query-dependent selection and downstream use across the nine models.

\paragraph{Visual concept accessibility.}

\begin{figure}[p]
    \centering
    \includegraphics[width=\linewidth]{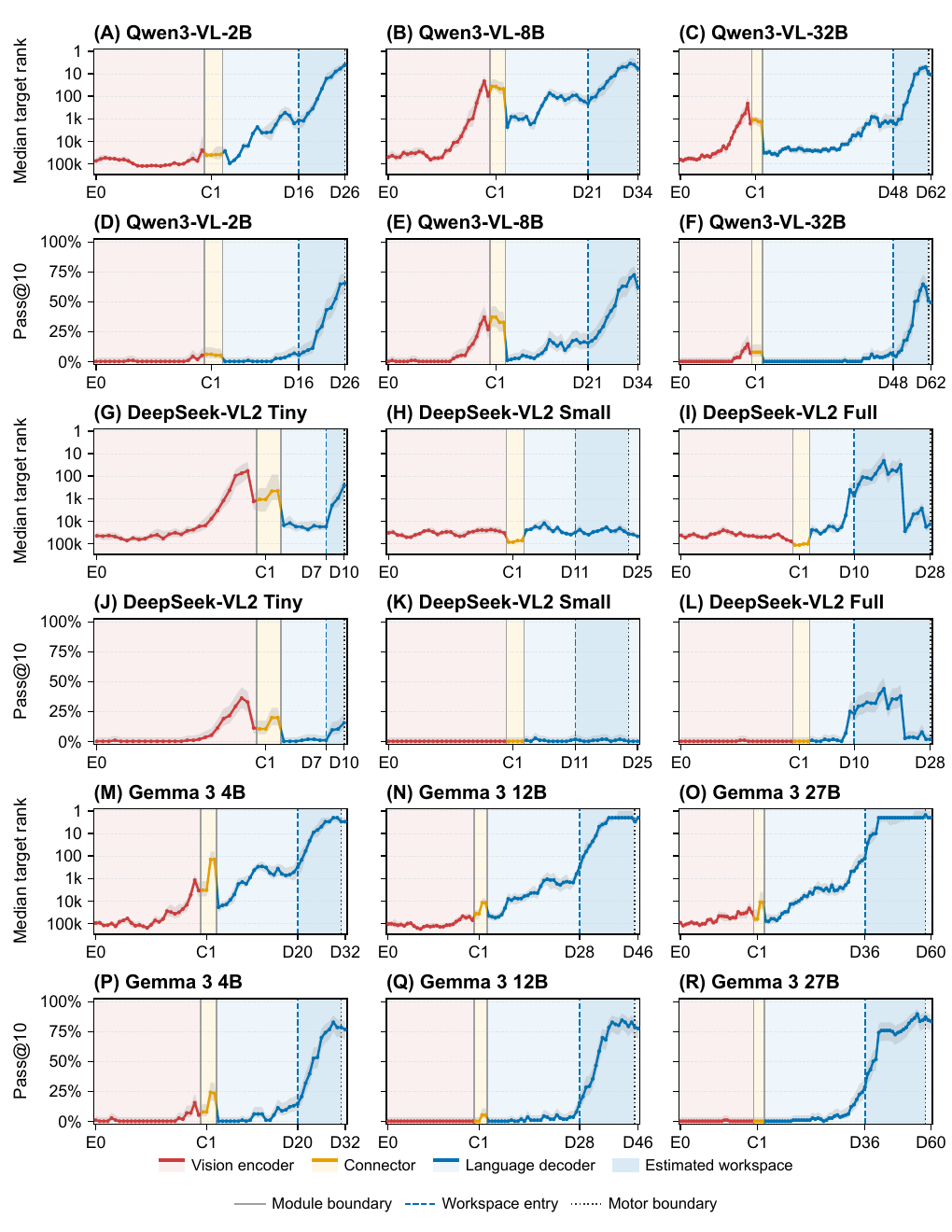}
    \caption{\textbf{Object-name readout across nine VLMs.}
    Median target rank and \(\operatorname{Pass}@10\) track readability from centred mean visual states without a question.
    Grey bands show 95\% confidence intervals.}
    \label{fig:cross-model-visual-readout}
\end{figure}

We first compare concept accessibility from visual states across network depth without a specific text query.
We conduct the object readout experiments of Section~\ref{sec:visual-readout} across nine models.
The median-rank and \(\operatorname{Pass}@10\) curves in Figure~\ref{fig:cross-model-visual-readout} track changes in object-name readability.

Across most models, object-name readout improves remarkably within the estimated workspace.
In some models, including Qwen8B and 32B, readability already increases towards the end of the encoder, then falls at the start of the decoder before rising again.
In Qwen and Gemma, decoder readouts improve within the estimated workspace and exceed those in the encoder, as seen in the rank and \(\operatorname{Pass}@10\) curves in Figure~\ref{fig:cross-model-visual-readout}~(A-F,M-R).
This improvement within the estimated workspace is shared despite differences in direction geometry.\looseness=-1

The main exceptions are DeepSeek Small and Full, shown in Figure~\ref{fig:cross-model-visual-readout}~(H,I,K,L).
Small remains weak throughout, while Full improves within the estimated workspace but falls sharply in the final few decoder layers.

\paragraph{Query-dependent selective access.}

We next compare concept accessibility and query-dependent selection across the nine models, following the settings in Sections~\ref{sec:multimodal-readout-selection} and~\ref{sec:question-dependent-selection}.
We measure \(\operatorname{Pass}@10\) for concept accessibility and selective sensitivity for query-dependent selection.
Higher \(\operatorname{Pass}@10\) indicates better readability of the target answer token, while higher selective sensitivity indicates more consistent selection under all three questions about each image.\looseness=-1

\begin{figure}[p]
    \centering
    \includegraphics[width=\linewidth]{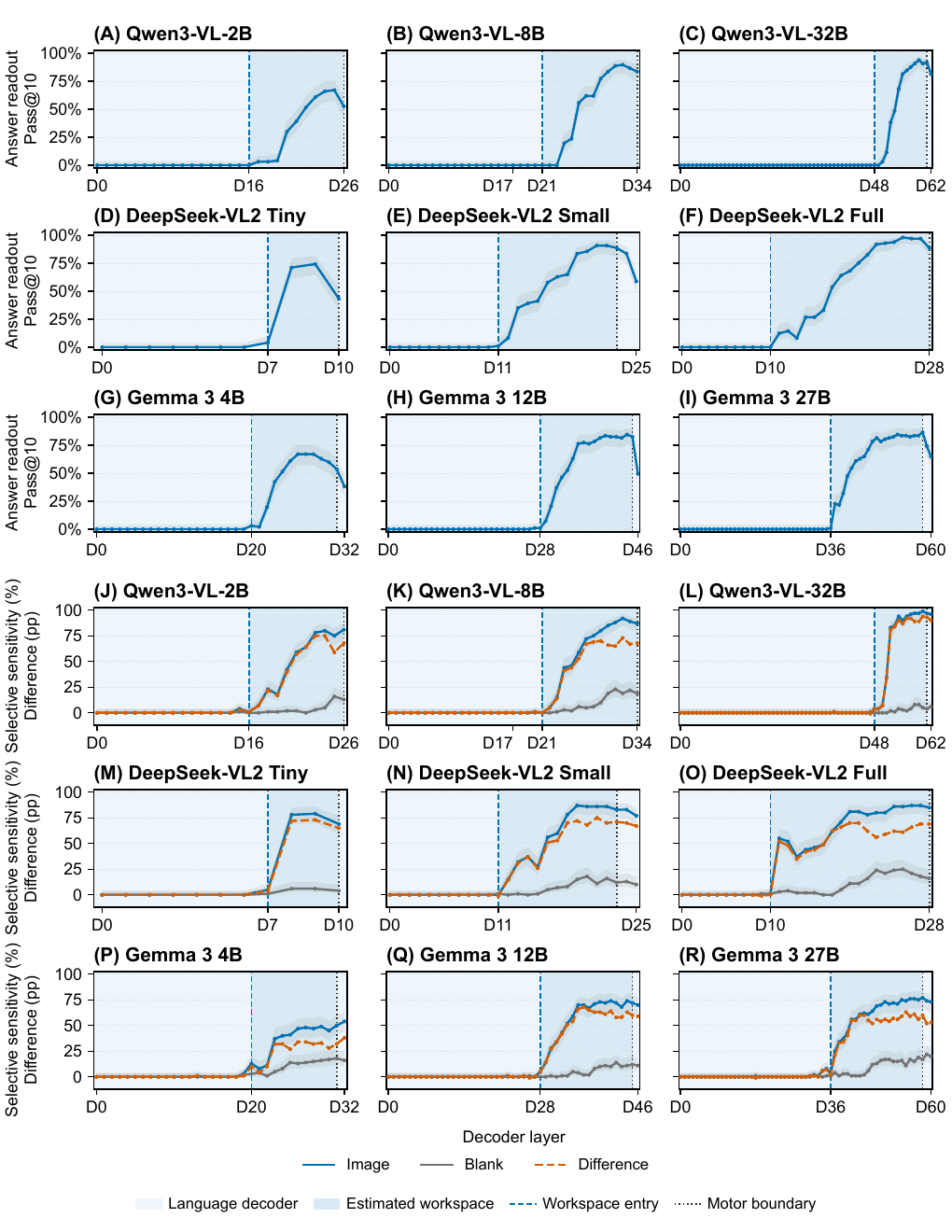}
    \caption{\textbf{Answer readout and query-dependent selection across nine VLMs.}
    \mbox{\textbf{(A-I)}} Answer-token \(\operatorname{Pass}@10\) with image evidence.
    \mbox{\textbf{(J-R)}} Selective sensitivity, the percentage of images with successful selection under all three questions at the same decoder layer.
    Difference is the image-minus-blank gap in percentage points.
    Grey bands show 95\% confidence intervals.}
    \label{fig:cross-model-readout-selection-depth}
\end{figure}

Across all three model families, answer readout remains low in early decoder layers and increases mainly within the estimated workspace, as shown in Figure~\ref{fig:cross-model-readout-selection-depth}~(A-I).
Query-dependent selection also improves within this region, as shown in panels (J-R).
Image input produces a much larger increase in selection than a blank image, indicating that the questions select information supplied by the scene.

\begin{figure}[tbp]
    \centering
    \includegraphics[width=\linewidth]{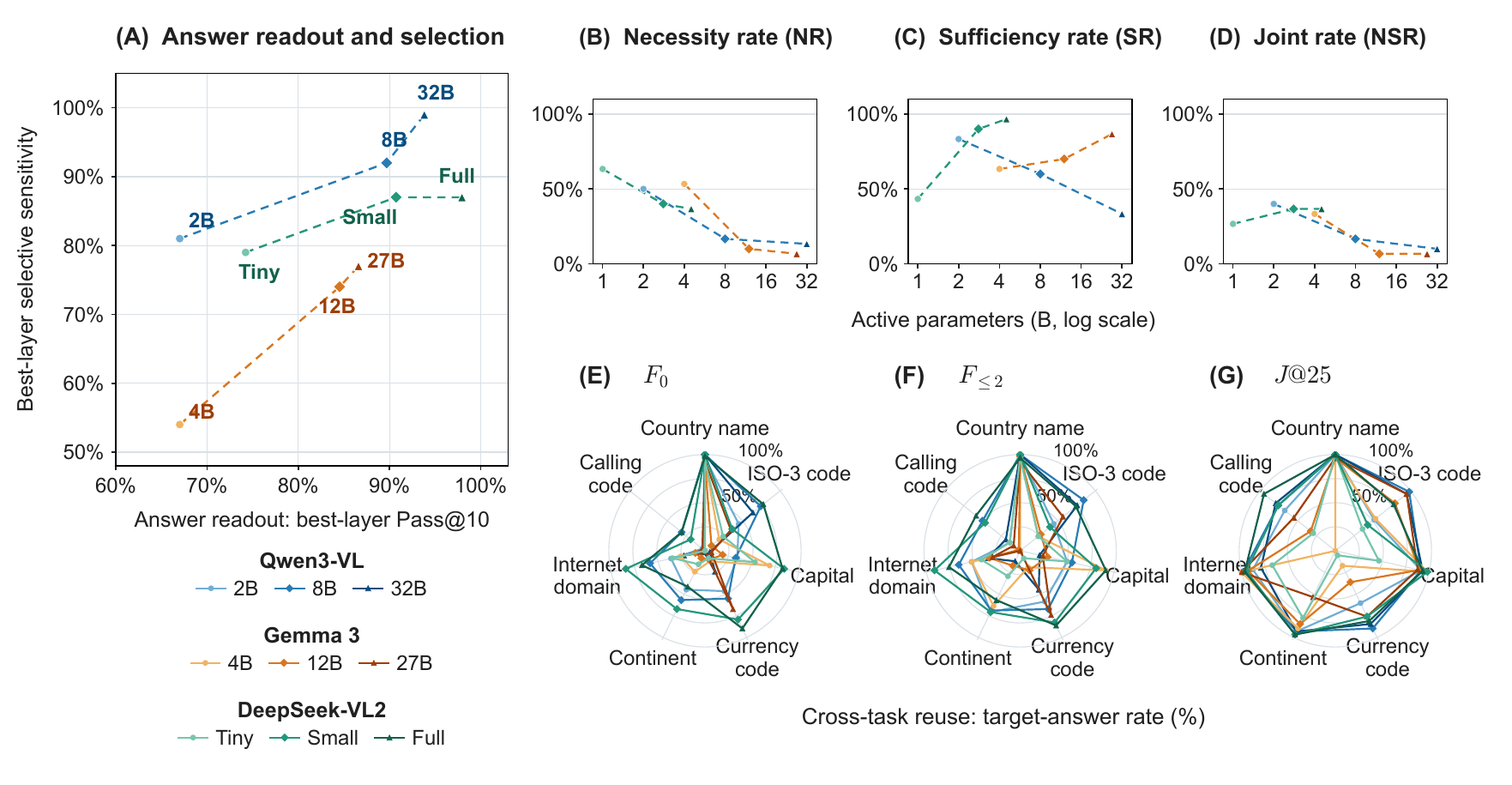}
    \caption{\textbf{Readout, selection and downstream effects across nine VLMs.}
    \textbf{(A)} Best-layer answer-token \(\operatorname{Pass}@10\) and selective sensitivity.
    \textbf{(B-D)} NR, SR and NSR for capital answers, using \(\greedyset{25}+\texttt{SubRep}\) on country-related components.
    \textbf{(E-G)} Cross-task target-answer rates using \texttt{SubRep} with three concept direction sets for the country.
    In \textbf{(A-D)}, dashed lines connect models within each family.}
    \label{fig:cross-model-workspace-properties}
\end{figure}

We then jointly analyse each model's best-layer answer readout and query-dependent selection in Figure~\ref{fig:cross-model-workspace-properties}~(A).\footnote{The measurements use different image sets and may peak at different layers.}
Points towards the upper right indicate stronger answer readout and more reliable selection, while those towards the lower left indicate weaker performance on both.
The within-family trajectories in Figure~\ref{fig:cross-model-workspace-properties}~(A) move towards the upper right as model size increases in Qwen and Gemma, combining stronger answer readout with more reliable query-dependent selection.
DeepSeek also moves up and right from Tiny to Small, but the move from Small to Full is mainly rightward, with better readout and the same selection peak.
Qwen lies above Gemma at comparable readout levels, indicating that similar answer accessibility can accompany different selection success.\looseness=-1

Across model families, answer readout and query-dependent selection both strengthen within the estimated workspace.
Larger models generally achieve stronger answer readout and more reliable selection, but models with similar answer readout can differ in selection success.

\paragraph{Necessity and sufficiency analysis.}

We next examine the necessity and sufficiency of selected intermediate concept components for downstream reasoning, following the experimental setting in Section~\ref{sec:necessity-sufficiency}.
All nine models are required to answer capital-identification questions on national-flag images from Noto Color Emoji.
We apply one-hop interventions to the country-related state components along the concept directions of the greedy set \(\greedyset{25}\).
As in Section~\ref{sec:necessity-sufficiency}, both tests apply the \(\operatorname{do}\) intervention for \texttt{SubRep} in Equation~\ref{eq:do-subrep}.
Following the definitions in Section~\ref{sec:necessity-sufficiency}, higher NR indicates greater dependence on the accessed concept information in downstream computation.
Higher SR indicates that this information is more often sufficient to produce the correct downstream answer.
Higher NSR indicates that the accessed concept information is more often both necessary and sufficient for the correct downstream answer.

Across all three families, NR decreases as model size increases, indicating that fewer original answers depend on the selected country-related components (Figure~\ref{fig:cross-model-workspace-properties}~(B)).
However, SR increases with model size in Gemma and DeepSeek but decreases in Qwen (Figure~\ref{fig:cross-model-workspace-properties}~(C)).
Thus, lower necessity in larger models is a shared pattern, while changes in sufficiency differ across model families.

In every model, NSR remains below SR (Figure~\ref{fig:cross-model-workspace-properties}~(C,D)).
This means that the selected components can be sufficient to recover the correct answer without being necessary for the original answer, with this pattern particularly pronounced in the larger Gemma and DeepSeek models.

\paragraph{Shared concept directions for multiple downstream tasks.}

We finally extend the cross-task intervention experiment of Section~\ref{sec:cross-task-reuse} to all nine models to examine the downstream use of the same concept directions across tasks.
Within each model, we keep the concept directions for the country fixed across the seven questions.
We compare \(F_0\), \(F_{\leq2}\) and \(\greedyset{25}\) using target-answer~rates.

The \(\greedyset{25}\) profiles in Figure~\ref{fig:cross-model-workspace-properties}~(G) generally extend further outwards than the word-form profiles in panels (E) and (F).
Expanding from \(F_0\) to \(F_{\leq2}\) increases the mean target-answer rate across the seven tasks in every model, and \(\greedyset{25}\) raises it further.
With \(\greedyset{25}\), larger models generally achieve stronger control, although increasing model size does not always improve the target-answer rate.
These experiments therefore show that, within each model, the same concept directions can influence answers across different downstream tasks.
The stronger control obtained with direction sets extending beyond concept names suggests that they capture more of the concept-related information used across these tasks.

Generally speaking, these VLMs share patterns of concept accessibility and query-dependent selection within the estimated workspace.
The accessed concept information can also contribute to different downstream computations.
These shared functions are observed across models with different internal workspace structures.
These findings provide functional evidence for the selective access and broad availability of concept information described by GWT.

\section[Cross-Modal Concept Alignment and Selective Access during Learning]{Cross-Modal Concept Alignment and Selective Access during Learning}
\label{sec:alignment-learning}

Our experiments so far have examined concept accessibility and downstream effects in pre-trained VLMs with well-developed cross-modal alignment.
In this section, we aim to investigate how cross-modal concept alignment and a selective access mechanism develop during training.
To build a simple experimental setting, we connect a frozen CLIP vision encoder~\citep{radford2021clip} to a frozen Gemma language decoder~\citep{gemmateam2024gemma2} through a randomly initialised multilayer perceptron (MLP) connector.
We train the connector on an image-caption prediction task for vision-language alignment, while keeping the vision encoder and language decoder frozen, as in the feature-alignment stage of LLaVA~\citep{liu2023visual} (Appendix~\ref{app:alignment-training}).
By examining different training checkpoints, we can track the development of cross-modal concept alignment and selective access.

\subsection[Tracking Cross-Modal Concept Alignment during Training]{Tracking Cross-Modal Concept Alignment during Training}
\label{sec:alignment-measurements}

In this part, we use MC-Lens to analyse cross-modal concept alignment during VLM training.
We develop two metrics, J-state cosine and reconstruction coverage (ReconCov), to measure cross-modal concept correspondence.
These metrics use concept directions to focus the analysis on information associated with a specific concept.
We then evaluate how well these metrics track cross-modal concept alignment during training.

\begin{figure}[t]
    \centering
    \includegraphics[width=\linewidth]{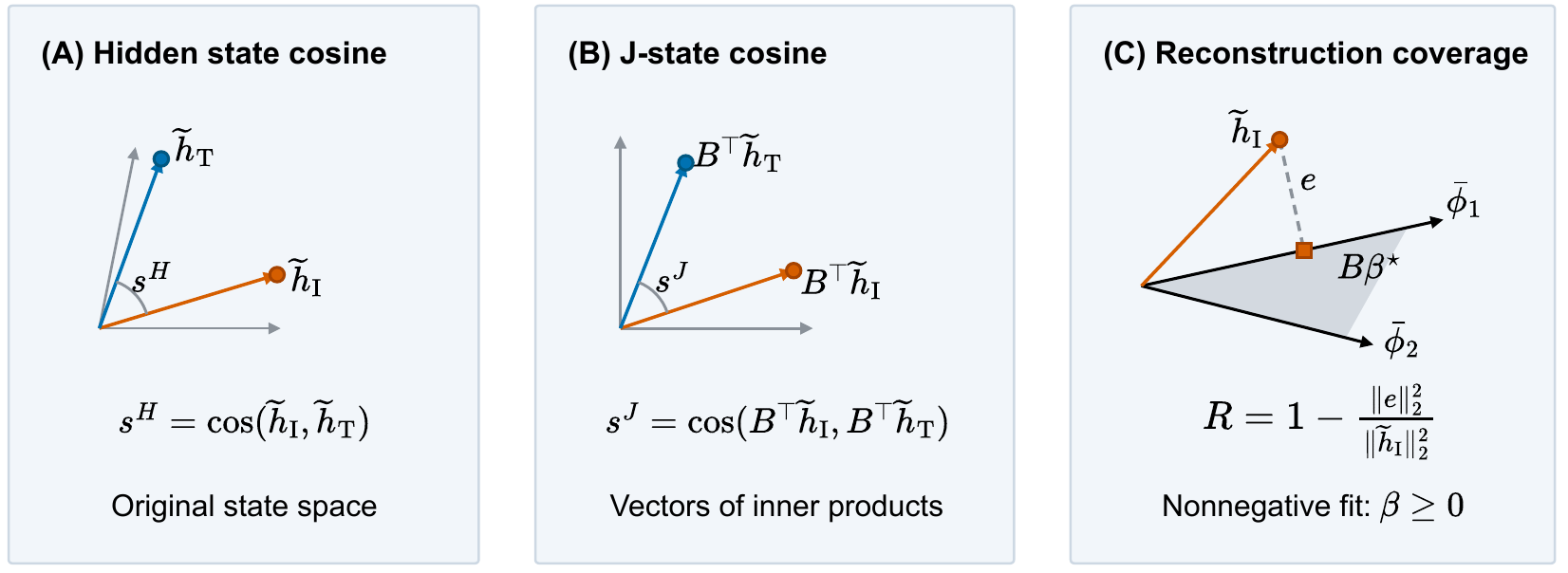}
    \caption{\textbf{Measuring cross-modal concept correspondence.}
    \textbf{(A)} Hidden state cosine.
    \textbf{(B)} J-state cosine.
    \textbf{(C)} Reconstruction coverage (ReconCov, \(R\)).
    Here, \(\widetilde h_{\mathrm I}\) and \(\widetilde h_{\mathrm T}\) are the mean image and text states at the same decoder layer, centred against their respective reference means, and \(B\) contains unit concept directions.}
    \label{fig:alignment-measurements}
\end{figure}

\paragraph{J-state cosine.}
J-state cosine measures the similarity between image and text states through their inner products with the same concept directions, as illustrated in Figure~\ref{fig:alignment-measurements}~(B).
Specifically, let \(\widetilde h_{\mathrm I}\) and \(\widetilde h_{\mathrm T}\) be the centred mean states for an image input and a text input naming the concept, respectively.\footnote{States are averaged over the input positions at the same decoder layer. We follow the reference centring in Section~\ref{sec:vocabulary-readout}.}
For each state, we take its inner product with every column of \(B\), the matrix of unit concept directions defined in Section~\ref{sec:concept-direction-sets}.
We compute the cosine between the resulting vectors \(B^\top\widetilde h_{\mathrm I}\) and \(B^\top\widetilde h_{\mathrm T}\):
\begin{equation}
    s^J=\cos\!\left(B^\top\widetilde h_{\mathrm I},B^\top\widetilde h_{\mathrm T}\right).
    \label{eq:concept-response-cosine}
\end{equation}
\begin{samepage}
The resulting J-state cosine ranges from \(-1\) to \(1\).
Values closer to \(1\) indicate more similar patterns of inner products, values near \(0\) indicate orthogonal patterns, and values near \(-1\) indicate opposing patterns.
For comparison, we apply hidden state cosine as a baseline~\citep{shukor2024implicit}, which compares the centred states directly without selecting concept directions in J-space:
\begin{equation}
    s^H=\cos\!\left(\widetilde h_{\mathrm I},\widetilde h_{\mathrm T}\right).
    \label{eq:hidden-state-cosine}
\end{equation}

\end{samepage}

\paragraph{Reconstruction coverage.}
A high J-state cosine does not necessarily mean that these concept directions can accurately reconstruct the image state.
We therefore define ReconCov, a reconstruction-based metric that measures the extent to which these directions can reconstruct the centred image state.
For example, we use greedy concept directions selected from text describing \emph{kayak} to reconstruct the centred state of a kayak image.
Figure~\ref{fig:alignment-measurements}~(C) illustrates this reconstruction.
We find the nonnegative coefficients that minimise the squared reconstruction error:
\begin{equation}
    \beta^\star=\underset{\beta\geq0}{\operatorname{argmin}}\,
        \left\|\widetilde h_{\mathrm I}-B\beta\right\|_2^2.
    \label{eq:alignment-reconstruction}
\end{equation}
We call \(B\beta^\star\) the J-space component of \(\widetilde h_{\mathrm I}\) on this fixed direction set, following J-Lens~\citep{gurnee2026verbalizable}.
We define ReconCov as
\begin{equation}
    R=1-\Bigl(\left\|\widetilde h_{\mathrm I}-B\beta^\star\right\|_2^2
        \,/\,\left\|\widetilde h_{\mathrm I}\right\|_2^2\Bigr).
    \label{eq:alignment-reconstruction-coverage}
\end{equation}
Here, ReconCov is one minus the squared reconstruction error \(\left\|\widetilde h_{\mathrm I}-B\beta^\star\right\|_2^2\) divided by the squared norm of the centred image state \(\|\widetilde h_{\mathrm I}\|_2^2\).
A ReconCov near \(1\) means an almost exact reconstruction, while a ReconCov near \(0\) means that the directions capture little of the state.

\Needspace{6\baselineskip}
\paragraph{Concept discrimination.}
The proposed J-state cosine and ReconCov measure different aspects of cross-modal concept correspondence, so their raw scores cannot be compared directly to determine which metric performs better.
In this part, we want to compare their ability to distinguish images of a given concept from images of other concepts.

\begin{samepage}
Specifically, for each concept, we calculate every image's score for that same concept using each metric.
We then compare the scores of images depicting this concept with those of images depicting other concepts.
For example, when evaluating \emph{kayak}, kayak images should receive higher scores for this concept than images of other objects, such as blenders and cellphones.
If a metric gives similarly high scores to both groups, it does not distinguish the concept well.
We use changes in this concept discrimination during training to evaluate the metrics' ability to reflect the development of cross-modal concept alignment.
\par
\end{samepage}

We measure this discrimination using the area under the receiver operating characteristic curve (AUC)~\citep{fawcett2006introduction}: how often an image of the corresponding concept scores above an image of another concept.
Let \(\mathcal P_c\) and \(\mathcal N_c\) be the images of concept \(c\) and the images of other concepts, respectively, and let \(s(I,c)\) be the score assigned to image \(I\) for concept \(c\).
Then
\begin{equation}
    \operatorname{AUC}_c
    =\frac{1}{|\mathcal P_c|\,|\mathcal N_c|}
    \sum_{I^+\in\mathcal P_c}\sum_{I^-\in\mathcal N_c}
    \Bigl[\mathbf 1\{s(I^+,c)>s(I^-,c)\}
    +\tfrac12\mathbf 1\{s(I^+,c)=s(I^-,c)\}\Bigr],
    \label{eq:concept-discrimination-auc}
\end{equation}
where \(\mathbf 1\) equals \(1\) when its condition holds and \(0\) otherwise.
An AUC of \(1\) indicates perfect separation, while an AUC of \(0.5\) indicates chance ranking.
We calculate AUC separately for each concept in the evaluated concept set \(\mathcal C\).
We define the concept discrimination score as the mean AUC across concepts, \(\frac{1}{|\mathcal C|}\sum_{c\in\mathcal C}\operatorname{AUC}_c\), with equal weight for each concept.
This evaluation applies to J-state cosine, ReconCov and the hidden state cosine baseline.

\paragraph{Evaluation during training.}
In this part, we evaluate concept discrimination at different training checkpoints.
We use the object-readout set from Section~\ref{sec:visual-readout}, with four images for each of 32 concepts.
We evaluate the proposed metrics, J-state cosine and ReconCov, using the word-form set \(F_{\leq2}\) and the greedy sets \(\greedyset{k}\) selected from separate development text, and compare all four combinations with hidden state cosine.
We hold the layer-specific MC-Lens mappings fitted at the final checkpoint and the concept direction sets fixed across training.
At each checkpoint, we average each image's score for a concept across decoder layers, before calculating AUC and the concept discrimination score.

\suppressfloats[t]
\begin{figure}[htbp]
    \centering
    \includegraphics[width=\linewidth]{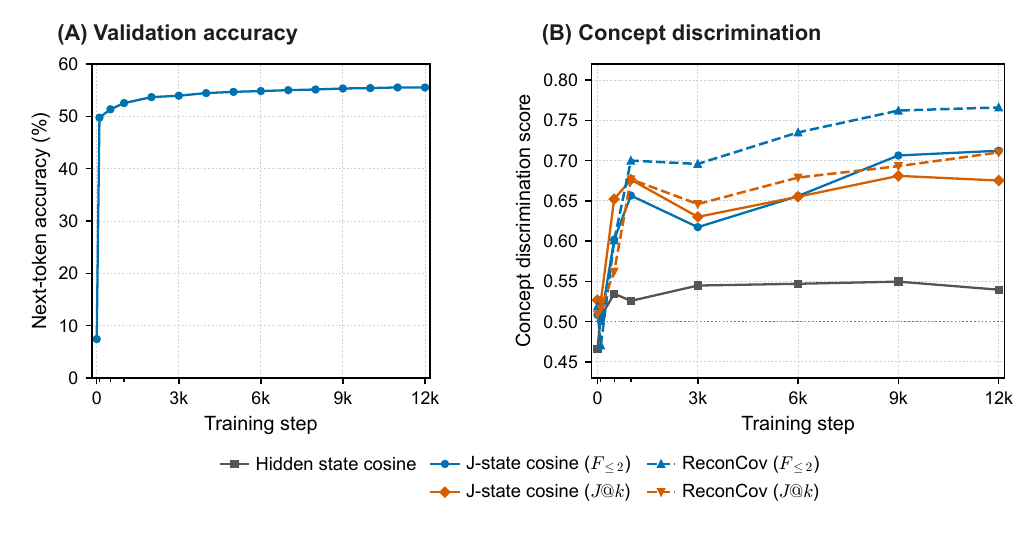}
    \caption{\textbf{Caption prediction and concept discrimination during training.}
    \textbf{(A)} Held-out caption next-token accuracy, used as a control for progress on the training task.
    \textbf{(B)} Concept discrimination scores, used to compare the metrics' ability to track the development of cross-modal concept alignment during training.}
    \label{fig:alignment-training}
\end{figure}

As shown in Figure~\ref{fig:alignment-training}~(A), caption next-token accuracy improves rapidly early in training and then more gradually.
This shows that the trained connector increasingly supports caption prediction from the image and the preceding caption tokens.
All five settings initially show near-chance concept discrimination, as shown in Figure~\ref{fig:alignment-training}~(B).
As training progresses, both proposed metrics distinguish images of the corresponding concept from other images more reliably, while hidden state cosine remains close to chance.

At later checkpoints, ReconCov distinguishes images of the corresponding concept from other images better than J-state cosine for either direction set.
For ReconCov, the word-form set \(F_{\leq2}\) consistently provides better concept discrimination than the greedy sets \(\greedyset{k}\) after early training.
For J-state cosine, \(\greedyset{k}\) provides better concept discrimination early in training, but \(F_{\leq2}\) performs better at later checkpoints.

These results show that our proposed metrics can track the development of cross-modal concept correspondence during training.
They also provide better concept discrimination than hidden state cosine, making the development of concept correspondence more apparent.

\input{cards/learning-query-selectivity}

\subsection[Development of Selective Access during Training]{Development of Selective Access during Training}
\label{sec:selective-access-development}
Our earlier experiments showed selective access to visual concepts in trained VLMs.
Here, we use checkpoints from the same controlled training run to examine how this ability develops during vision-language alignment.

Example~\ref{ex:learning-query-selectivity} pairs a fixed image of a cup on the left and a book on the right with questions asking for either object.
At each checkpoint, we read both object names from the pre-answer decoder state and record their best ranks across complete single-token word forms.
We compare these ranks under the two questions to examine question-dependent selection.

Early in training, both object names become more readable under either query, but both queries favour \emph{book}.
After \(1{,}000\) steps, question-dependent differences become clearer: the left query makes \emph{cup} more readable and \emph{book} less readable, while the right query continues to favour \emph{book} and makes \emph{cup} less readable.
At the later checkpoints, each query favours its requested object.

In this example, concept accessibility improves early in training, before reliable selective access develops.
The later question-dependent changes suggest that selective access develops during training.

\section{Discussion}
\label{sec:discussion}

\noindent\textbf{Concept readouts beyond verbalisation.}
In this paper, we propose MC-Lens, which allows us to read verbalised concepts from intermediate visual and language states.
However, reading a concept ``name'' gives only a limited description of the corresponding concept representation.
In this paper, we extend this description to include word-form sets \(F_{\leq j}\) and greedy sets \(\greedyset{k}\).
Counterfactual experiments also show that interventions on intermediate states using a set of concept directions change one-hop and multi-hop answers more reliably than those using only the vocabulary direction of a single name token.

These extensions broaden the description of a concept, but they still use the model's language vocabulary.
For example, object names and attribute words give only a partial description of an object's shape, texture and other visual details.

This motivates future work on concept readouts that express concept information directly through multimodal representations, alongside or instead of words.
For example, images, video and audio could provide readouts of content that is difficult to verbalise.
Such readouts could help us study the representation, selection and use of this information in subsequent computation.

\noindent\textbf{Multimodal concept processing and GWT.}
With our proposed MC-Lens, we examine the relationship between multimodal concept processing in VLMs and GWT~\citep{baars2005workspace}.
Our experiments show that multimodal concept information can be selectively accessed in intermediate states and can influence downstream computation across tasks.
These patterns appear across model families and provide functional evidence for the selective access and downstream availability of information described by GWT.

However, our results do not yet identify the mechanisms for broadcasting selected multimodal information to different downstream processes, which is a central property of GWT~\citep{baars2005workspace}.
Future work could trace selected information through downstream computations and test the pathways supporting its use.
As GWT seeks to explain conscious access, studying these mechanisms in LLMs and VLMs could contribute to research on machine consciousness~\citep{dehaene2017consciousness,butlin2023consciousness}.

\noindent\textbf{J-space constraints for cross-modal alignment.}
In Section~\ref{sec:alignment-measurements}, we use J-space metrics to measure the development of concept-wise vision-language alignment during training.
These metrics could also serve as additional constraints during training.

For example, an auxiliary loss could encourage image and text states expressing the same concept to respond similarly along its concept directions at the same decoder layer.
Future work could combine this objective with token prediction and test its effects on concept alignment, query-dependent selection and downstream task performance.

\section{Related Work}
\label{sec:related-work}

\subsection{Global Workspace Theory}
\label{sec:global-workspace-theory}

In an experiment on human visual attention, observers watched a video of two teams passing basketballs and counted the passes made by one team.
During the video, a person in a gorilla suit walked among the players.
Surprisingly, many observers failed to notice the gorilla~\citep{simons1999gorillas}.
Global workspace theory offers an explanation for such failures of conscious perception~\citep{baars2005workspace}.
In GWT, specialised processors carry out operations such as detecting motion and recognising objects.
These processors work in parallel, and their outputs compete for access to a shared workspace with limited capacity.
The workspace briefly holds the selected content and broadcasts it to all processors across the whole system, including those involved in memory, language and action~\citep{baars2007architectural}.
Global workspace theory associates conscious access with this broadcast.
In the ``basketball and gorilla'' example, the counting task gives greater priority to information about the attended team's passes in the competition for workspace access.
Information about the gorilla may lose this competition and remain outside the workspace.
Without being broadcast, the gorilla information remains unavailable for conscious report, and the observer fails to notice the gorilla.

The global neuronal workspace model~\citep{dehaene1998neuronal} explains GWT's selection and broadcast mechanisms in terms of neural activity in the human brain.
This model proposes a network of neurons distributed across several cortical regions and linked by long-range connections.
The network receives signals from specialised processors and sends feedback to amplify or suppress their activity.
\citet{dehaene2001towards} elaborated the conditions for conscious access.
They emphasised that a representation must be amplified and sustained through interactions with workspace neurons to become available to multiple processes, including perceptual categorisation, long-term memory, evaluation and intentional action.
\citet{dehaene2003neuronal} further simulated two visual targets presented in quick succession.
The first target could prevent the second from triggering \emph{ignition}, a sudden transition from an early sensory response to sustained activity across the workspace.
The model predicted that the second target would tend to be either unseen or clearly seen, consistent with the two peaks observed in human observers' visibility ratings~\citep{dehaene2003neuronal}.

Global workspace theory has also helped advance research on machine consciousness.
\citet{blum2022conscious} formalised GWT's selection and broadcast mechanisms in the Conscious Turing Machine (CTM), an explicit computational model of consciousness.
In this model, long-term memory (LTM) consists of many specialised processors, each with its own memory.
A processor submits information as a chunk, a short message containing its content and an assigned importance.
Short-term memory (STM) serves as the shared workspace and holds one chunk at a time.
The Up Tree selects a winning chunk through successive pairwise competitions, and chunks with higher competition scores are more likely to advance.
The winning chunk enters STM, and the Down Tree broadcasts it to all LTM processors.
The authors also used the ``basketball and gorilla'' example~\citep{simons1999gorillas} to illustrate this process.
The instruction to count one team's passes is first broadcast to all processors.
The visual processors then give greater priority to information about that team.
The gorilla chunk has little chance of entering STM and being broadcast, so the CTM fails to consciously notice the gorilla~\citep{blum2022conscious}.
The CTM gives GWT a precise computational formulation, with explicit rules for competition, storage and broadcast.

Beyond formal accounts of consciousness, GWT has also inspired neural network architectures that use shared workspaces to coordinate information processing.
\citet{goyal2022workspace} added a few shared memory slots to Transformers to coordinate information across input positions.
Representations at different positions compete for write access, and each position reads the selected information to update its state.
This design improved performance on visual reasoning tasks.
\citet{vanrullen2021deepgwt} proposed a shared latent workspace to connect separately trained vision, language and other networks.
The workspace would translate selected information from one network into representations usable by the other networks, implementing broadcast across modalities.
To reduce dependence on paired image-text data, \citet{devillers2025workspace} trained mappings between frozen vision and language networks through a shared workspace.
They also trained the mappings to reconstruct each representation after translation to the other modality and back, which allowed learning from unpaired data and reduced the number of paired examples needed for alignment and translation.
\citet{fountas2026workspace} built an explicit workspace, the Global Key-Value Workspace, into pretrained LLMs.
Their method selects components from the key-value cache, writes them to the workspace and broadcasts the result back to all components.
Their experiments showed improved reasoning performance, and selecting fewer components preserved much of the performance of full broadcast at lower computational cost.

Another line of research examines the emergence of workspace-like organisation through LLM training.
To study information integration, \citet{urbina2026synergistic} measured synergy between attention heads, which is information that the heads carry only in combination, not individually.
They identified a synergistic core in the middle layers of trained models, absent in randomly initialised models.
The ablation of high-synergy heads disproportionately impaired mathematical reasoning.
Reinforcement learning restricted to these heads also produced larger gains than training redundant or randomly selected heads.
\citet{gurnee2026verbalizable} developed J-Lens to investigate global workspace structure and function in LLMs.
They compared J-space geometry across layers using CKA and examined concept readouts and representational capacity.
These analyses identified an intermediate band of layers with coherent concept representations and limited capacity.
Within this band, MLP layers amplified J-space directions far more strongly than random directions.
A subset of attention heads relayed J-space content across token positions.
Both findings are consistent with GWT's broadcast mechanism.
To study concept availability during reasoning, \citet{chun2026manifold} measured how easily the model's internal states could be separated according to the value of an intermediate variable.
The states were easy to separate while the model computed the variable and again when a later step used it.
In between, they were much harder to separate, although trained probes could still recover the variable's value.
The authors interpreted this pattern as the allocation of limited representational capacity to the current computation.
In this paper, we extend this line of work to visual concepts in VLMs, examining selective access to these concepts under different questions and their availability to multiple downstream computations.

\subsection{Mechanistic Interpretability}
\label{sec:vocabulary-readout-interventions}

A model's final answer provides only a limited view of its internal concept representations.
To study these representations, researchers have developed methods for reading information from intermediate states.
\citet{alain2016understanding} introduced linear classifier probes, training a separate classifier on each layer's activations while keeping the original network fixed.
To discover features beyond predefined categories, \citet{bricken2023monosemanticity} used a sparse autoencoder to decompose activations in a one-layer language model.
This revealed features for specific text patterns, and activating selected features steered generation towards the corresponding patterns.
Another way to read an intermediate state is to express its content through the model's own vocabulary.
The Logit Lens~\citep{nostalgebraist2020logitlens} applies the model's output mapping directly to each layer's hidden states to track token predictions across layers.
The Tuned Lens~\citep{belrose2023tunedlens} first applies a learned transformation specific to each layer.
Its readouts more closely match the model's final token probabilities.
Patchscopes~\citep{ghandeharioun2024patchscopes} inserts a hidden state into a decoding prompt and lets the model generate a description of its content.

Vocabulary readouts can also be constructed from the model's Jacobians, which describe how changes in intermediate states affect its output.
\citet{hernandez2024linearity} proposed the Attribute Lens to read attributes from entity representations.
They constructed a mapping by averaging Jacobians over examples of the same relation, such as musician-instrument pairs, allowing \emph{trumpet} to be recovered from the representation of \emph{Miles Davis}.
\citet{gurnee2026verbalizable} applied the same averaging principle across prompts and token positions to construct J-Lens for reading concepts across contexts.
J-Lens maps intermediate states into the final layer's representation space and applies the model's output mapping, revealing intermediate concepts omitted from answers in multi-hop reasoning.
To further reduce noise in the averaged Jacobian, \citet{yan2026shorthorizons} retained only the pairs of token positions with the largest Jacobian norms.
Keeping the top 20\% improved intermediate-concept recall on five of six tasks in Qwen3-8B.
To improve readouts in early layers, \citet{blank2026rlens} introduced R-Lens, replacing ordinary gradients with coefficients computed using layer-wise relevance propagation.
This reduced irrelevant tokens in the readouts and revealed some intermediate concepts earlier than J-Lens.

The original J-Lens ranks individual vocabulary tokens~\citep{gurnee2026verbalizable}.
However, a concept's name may span several tokens, and its representation may contain information beyond a single name token.
To recover complete concept names, \citet{gong2026multitoken} introduced Concept Lens, using token readouts and the input context to propose candidates.
It obtains each candidate's representation from a fixed prompt and scores it against the original hidden state.
These representations support both multi-token concept readout and concept swaps.
In VLMs, a visual concept can also be expressed through different names and related semantic content.
Our proposed MC-Lens therefore constructs a set of vocabulary directions for each concept.
Word-form selection collects alternative forms of its name, while greedy selection chooses directions that capture its representation, including directions for related concepts.
We use these sets to read visual concepts and test their roles in downstream computation through counterfactual interventions.

Reading a concept from an intermediate state does not establish that the model uses it in downstream computation.
\citet{yang2024latentmultihop} used gradients to estimate whether stronger recall of an intermediate entity would support the second step of two-hop reasoning.
Across the tested relations, they found substantial evidence for entity recall and moderate evidence for its subsequent use.
\citet{gurnee2026verbalizable} tested intermediate concepts by swapping their J-Lens coordinates with those of alternative concepts.
These swaps redirected two-hop answers and, in separate experiments, changed answers about different attributes of the same entity.
\citet{zahedi2026microlens} compared concept swaps at different token positions in two-hop capital questions.
The intermediate entity, a state or country, was readable at both the city-name and final prompt positions, but swapping it at the city name redirected answers more often.
To study the use of external evidence, \citet{kawada2026evidence} split the hidden state into the component reconstructed with J-Lens and the remaining component.
They swapped each component between runs that accepted a supplied answer and runs that did not.
In Llama and Qwen, swapping the reconstructed components preserved the rate of accepting a supplied answer, while swapping the remaining components reduced it.
In this paper, we use counterfactual interventions to examine the role of visual concepts in VLM reasoning.
We compare name-token directions with broader concept direction sets and test the necessity and sufficiency of the selected information for correct answers.
We also examine whether the same concept directions influence multiple downstream tasks and successive reasoning steps.

Mechanistic interpretability studies have also examined how VLMs represent visual information and use it to answer questions.
\citet{fu2025hidden} compared VLM answers with direct readouts from visual encoder features on the same tasks.
These readouts achieved better task performance, and the relevant visual information remained accessible in the connector and decoder.
To trace how visual information reaches the answer, \citet{salazar2026pathways} used activation patching between paired examples.
They found that open-ended object recognition relied on image tokens, while spatial judgements used visual information transferred to question tokens.
\citet{krojer2026latentlens} introduced LatentLens to interpret visual tokens by comparing them with contextualised text representations.
It retrieved descriptions of image regions even in early layers, where direct vocabulary projection was less informative.
The authors also found that, during connector training, visual representations became interpretable first near the language model's input and later in deeper layers.

In this paper, we use our proposed MC-Lens to examine the development of cross-modal concept alignment and selective access during controlled vision-language alignment training.
Our readouts show that concept accessibility improves early in training, while reliable selective access according to the question develops later.

\section{Conclusion}
\label{sec:conclusion}

In this paper, we introduce MC-Lens to access concept information from intermediate visual and language states in a shared semantic space.
Using MC-Lens, we examine concept accessibility and selective access, the role of concepts in downstream reasoning, the development of cross-modal alignment during training, and shared workspace functions across VLMs.
Our work provides an experimental basis for studying GWT in multimodal artificial systems.
It also opens a way to investigate how GWT can guide improvements in cross-modal alignment.
Further investigation of the mechanisms that make selected information available to different downstream processes could deepen our understanding of workspace functions and contribute to research on machine consciousness.

\FloatBarrier
\bibliographystyle{plainnat}
\begingroup
\addtolength{\bibsep}{-2.7pt}
\interlinepenalty=5000
\bibliography{references}
\endgroup

\manuscriptappendix{Appendix of MC-Lens}

\setcounter{figure}{0}
\renewcommand{\thefigure}{S\arabic{figure}}

\setcounter{table}{0}
\renewcommand{\thetable}{S\arabic{table}}

\setcounter{equation}{0}
\renewcommand{\theequation}{S\arabic{equation}}

\renewcommand{\theHfigure}{S\arabic{figure}}
\renewcommand{\theHtable}{S\arabic{table}}
\renewcommand{\theHequation}{S\arabic{equation}}

\section{Implementation and Experimental Details}
\label{app:implementation-details}

\subsection{Lens Fitting}
\label{app:lens-fitting}

We fit the lenses using multimodal examples from A-OKVQA~\citep{schwenk2022aokvqa}.
Following the J-Lens reference implementation~\citep{gurnee2026verbalizable}, we exclude padding, the first 16 sequence positions and the final sequence position from the decoder position sets defined in Section~\ref{sec:multimodal-lenses}.
These exclusions apply to the intermediate decoder positions for L-Lens and the final-layer decoder positions for both lenses.
We number Transformer blocks from zero and extract states after each block.

We also fit lenses to intermediate states within the connector, including states before and after its linear transformations and activation functions.
We follow Equation~\ref{eq:multimodal-jacobian} with \(m=C\).

\subsection{Greedy Selection for Interventions}
\label{app:greedy-intervention-directions}

We use greedy selection to construct concept direction sets for both state reconstruction (Section~\ref{sec:readout-reconstruction}) and counterfactual interventions (Section~\ref{sec:downstream-computation}).
In the procedure described in Section~\ref{sec:concept-direction-sets}, each iteration selects a vocabulary direction, refits the coefficients of all selected directions by NNLS, and updates the remaining difference.
For interventions, we approximate the NNLS refit to reduce the cost of repeated coefficient fitting.
At each iteration, we solve unconstrained least squares and set negative coefficients to zero before updating the remaining difference.

\subsection{Counterfactual Intervention Operators}
\label{app:intervention-operators}

We define four operators for editing decoder states to examine the effects of concept information on downstream answers.
The source and target runs use inputs expressing the source and target concepts under the same question.
At an edited layer \(\ell\), let \(h_\ell^s\) be the current source state and \(h_\ell^t\) the unedited target state at the same layer and position.
We write \(h_\ell'\) for the edited source state.
When several layers are edited, the updates are applied in order of depth.

For a vocabulary token \(w\), the raw vector is \(\phi_{\ell,w}=J_\ell^\top u_w\), where \(J_\ell\) is the decoder lens and \(u_w\) is the token's unembedding vector.
Its unit direction is \(\bar\phi_{\ell,w}=\phi_{\ell,w}/\|\phi_{\ell,w}\|_2\).
The superscript \(+\) denotes the Moore-Penrose pseudoinverse.

\paragraph{\texttt{PairSwap}.}
\texttt{PairSwap} exchanges the source and target name-token coordinates within the source state, following \citet{gurnee2026verbalizable}.
Let \(s\) and \(t\) index the \(F_0\) name tokens for the source and target concepts, respectively.
We use the raw vectors \(\phi_{\ell,s}\) and \(\phi_{\ell,t}\), retaining their original lengths.
The least-squares coordinates of \(h_\ell^s\) along these two vectors are
\begin{equation}\label{eq:pairswap-coordinates}
    \begin{bmatrix}a_s\\a_t\end{bmatrix}
    =
    \begin{bmatrix}\phi_{\ell,s}&\phi_{\ell,t}\end{bmatrix}^{+}h_\ell^s.
\end{equation}
The coordinates \(a_s\) and \(a_t\) may be negative.
We exchange them and map the coordinate difference back into the state space:
\begin{equation}\label{eq:pairswap-update}
    h_\ell'
    =h_\ell^s+
    \begin{bmatrix}\phi_{\ell,s}&\phi_{\ell,t}\end{bmatrix}
    \left(
    \begin{bmatrix}a_t\\a_s\end{bmatrix}
    -\begin{bmatrix}a_s\\a_t\end{bmatrix}
    \right).
\end{equation}
The update preserves the component of \(h_\ell^s\) orthogonal to the two directions.
It exactly exchanges the coordinates when the directions are linearly independent.

\paragraph{\texttt{SubRep}.}
\texttt{SubRep} replaces the source state's component in a selected subspace with the corresponding component of the target state.
Let \(\mathcal W_{\ell,s}\) and \(\mathcal W_{\ell,t}\) be the token sets selected for the source and target concepts by word-form selection or greedy selection.
We combine their tokens as \(\{w_1,\ldots,w_n\}=\mathcal W_{\ell,s}\cup\mathcal W_{\ell,t}\).
The unit directions form the columns of \(B_\ell\), and \(P_\ell\) projects onto their span:
\begin{equation}\label{eq:appendix-subrep-projector}
    B_\ell=
    \begin{bmatrix}\bar\phi_{\ell,w_1}&\cdots&\bar\phi_{\ell,w_n}\end{bmatrix},
    \qquad P_\ell=B_\ell B_\ell^{+}.
\end{equation}
The complete state update is
\begin{equation}\label{eq:appendix-subrep-update}
    h_\ell'=(I-P_\ell)h_\ell^s+P_\ell h_\ell^t,
\end{equation}
where \(I\) is the identity matrix.
The update retains the source component orthogonal to the selected subspace and takes the component within that subspace from the matched target run.

\paragraph{\texttt{AbIn}.}
\texttt{AbIn} removes the source concept component and injects a scaled mean of the target concept directions.
For ablation, we collect the unit directions indexed by \(\mathcal W_{\ell,s}\) as the columns of \(B_\ell^s\).
The projector onto their span is
\begin{equation}\label{eq:abin-source-projector}
    P_\ell^s=B_\ell^s(B_\ell^s)^+.
\end{equation}
For injection, we average the unit directions indexed by \(\mathcal W_{\ell,t}\):
\begin{equation}\label{eq:abin-target-direction}
    v_\ell^t=
    \frac{1}{|\mathcal W_{\ell,t}|}
    \sum_{w\in\mathcal W_{\ell,t}}\bar\phi_{\ell,w}.
\end{equation}
We use \(v_\ell^t\) without further normalisation.
To set the injection scale, let \(h_\ell^{s,0}\) denote the unedited source state at layer \(\ell\) and at the last position of the current prefix.
We take the norm of its projection onto the source span:
\begin{equation}\label{eq:abin-reference-scale}
    r_\ell=\|P_\ell^s h_\ell^{s,0}\|_2.
\end{equation}
The complete state update is
\begin{equation}\label{eq:abin-state-update}
    h_\ell'=(I-P_\ell^s)h_\ell^s+\alpha r_\ell v_\ell^t.
\end{equation}
The first term removes the source concept component, and the second adds the mean target direction with strength controlled by \(\alpha\).
We use \(\alpha=0.5\) for all \texttt{AbIn} comparisons in Figure~\ref{fig:intervention-direction-operator}.

\paragraph{\texttt{FullRep}.}
\texttt{FullRep} replaces the entire source state with the unedited target state at the same layer and position:
\begin{equation}\label{eq:fullrep-state-update}
    h_\ell'=h_\ell^t.
\end{equation}
We use this operator as a control for replacing the full state, including components outside the selected concept directions.

\subsection{Estimating Workspace Boundaries}
\label{app:workspace-boundaries}

We manually estimate the workspace intervals shown in Section~\ref{sec:workspace-organisation} by examining structural and readout changes across decoder layers.
For Qwen and DeepSeek, changes in vocabulary-direction geometry and effective dimensionality help identify the transition into the workspace.
For Gemma, where these structural changes are less distinct, the entry estimates rely more on prediction agreement, concept readout and query-dependent selection.

We place the end of the workspace at the transition towards the final output stage, where prediction agreement approaches its final-layer level.
The readout and selection profiles used to choose these boundaries do not provide independent validation of their locations.

\subsection{Vision-Language Alignment Training}
\label{app:alignment-training}

We paired CLIP ViT-L/14-336 with Gemma-2-9B-it.
We trained the gated MLP connector, the added visual-token embeddings and the visual padding embeddings for 12,000 steps on image-caption pairs from PixMo-Cap~\citep{deitke2025molmo}.
We evaluated caption next-token accuracy on held-out image-caption pairs, conditioning each prediction on the image and the preceding tokens of the reference caption.
A prediction was counted as correct if it matched the corresponding reference token.

\FloatBarrier

\section{Supplementary Comparisons}
\label{app:supplementary-comparisons}
\subsection{Stability of J-Space Geometry across Fitting-Set Sizes}
\label{app:lens-fit-size-comparison}

We examine whether the layerwise J-space organisation observed in Section~\ref{sec:layerwise-organisation} changes with the amount of data used to fit L-Lens.
We fit L-Lens for Qwen3-VL-8B using 1, 5, 10, 100 and 1,000 prompts and compare vocabulary-direction geometry across decoder layers using CKA.

\begin{figure}[htbp]
    \centering
    \includegraphics[width=\linewidth]{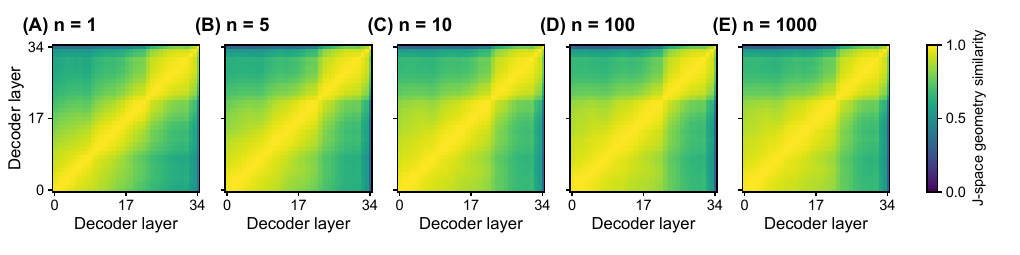}
    \caption{\textbf{Decoder J-space geometry across fitting-set sizes.}
    Each panel compares vocabulary-direction geometry across decoder layers of Qwen3-VL-8B using CKA, with L-Lens fitted on \(n\) prompts.}
    \label{fig:lens-fit-size-geometry}
\end{figure}

Figure~\ref{fig:lens-fit-size-geometry} shows that the broad distinction between early and late decoder layers is visible even with a single fitting prompt.
As the fitting set grows, the region of high similarity among early layers becomes clearer.
With 100 prompts, the pattern already closely resembles that obtained with the 1,000 prompts used in the main analysis.
The broad layerwise organisation is therefore preserved across the tested fitting-set sizes.

\subsection{Effects of Aggregation and Reference Centring on Visual Readout}
\label{app:visual-readout-aggregation-centring}

We examine how aggregation across image-token positions and reference centring affect whole-image object readout.
Using the images from Section~\ref{sec:visual-readout}, we compare reading the mean image-token state with reading each position separately and retaining each vocabulary token's highest score across positions.
We evaluate both approaches with and without reference centring.

\begin{table}[htbp]
    \centering
    \small
    \caption{\textbf{Object-name readout with and without reference centring.}
    Both readouts use L-Lens at decoder layer D32 of Qwen3-VL-8B on the same 128 images without a text question.
    \(\operatorname{Pass}@10\) is the percentage of images for which at least one complete single-token word form of the target object's name ranks among the top ten tokens.}
    \label{tab:visual-readout-aggregation-centring}
    \begin{tabular}{lrr}
    \toprule
    Decoder readout & Uncentred & Centred \\
    \midrule
    Mean image-token state & 0.8\% & 69.5\% \\
    Highest score per token across positions & 6.3\% & 1.6\% \\
    \bottomrule
    \end{tabular}
\end{table}

Without reference centring, both approaches yield low \(\operatorname{Pass}@10\) (Table~\ref{tab:visual-readout-aggregation-centring}).
Centring substantially improves object-name readout from the mean state, but reduces \(\operatorname{Pass}@10\) when each vocabulary token's highest score is taken across positions.
These results support using the centred mean state for whole-image object readout, as in Section~\ref{sec:visual-readout}.

\subsection{Effects of Intervention Depth and Window Width}
\label{app:intervention-depth-windows}

We examine how intervention depth and window width affect the direct and one-hop tasks in Section~\ref{sec:counterfactual-interventions}.
We apply \texttt{SubRep} with the greedy set \(\greedyset{25}\) to country-related components within windows of 3, 5 and 9 consecutive decoder layers.
For each width, we vary the final edited layer and measure the target-answer rate.
Comparing widths at the same final edited layer tests the effect of extending the intervention to earlier layers.

\begin{figure}[tbp]
    \centering
    \includegraphics[width=\linewidth]{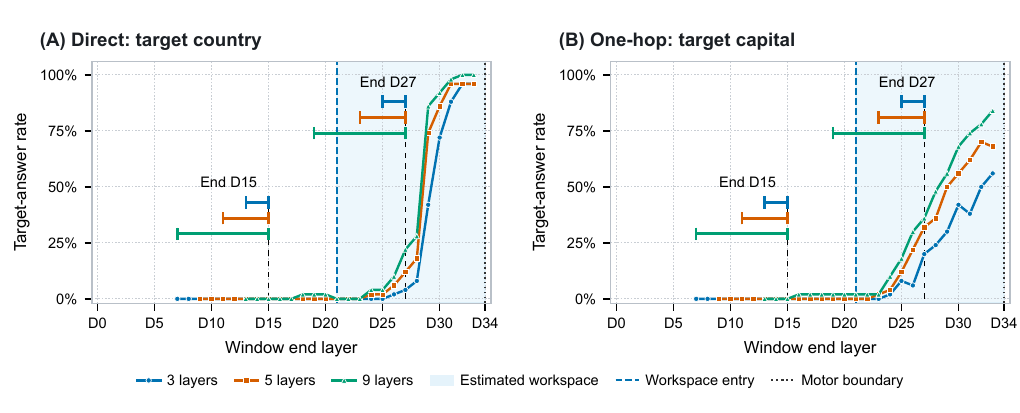}
    \caption{\textbf{Intervention depth and width affect downstream answers.}
    Curves show target-answer rates for \textbf{(A)} direct and \textbf{(B)} one-hop tasks under \(\greedyset{25}+\texttt{SubRep}\) on 50 source-to-target flag pairs in Qwen3-VL-8B.
    Each colour denotes a window width, and the horizontal axis gives the final edited layer.
    The coloured horizontal bars mark the span from the first to the last edited layer.}
    \label{fig:intervention-depth-window}
\end{figure}

Figure~\ref{fig:intervention-depth-window} shows that target-answer rates remain low at early window positions and rise mainly as the window end moves through the estimated workspace.
For the direct task, wider windows produce an earlier rise, but all widths reach similarly high rates at late layers.
For the one-hop task, wider windows retain a clear advantage even at late layers.
At a fixed final edited layer near the end of the decoder, wider windows therefore provide a larger benefit for the one-hop task.

\manuscriptendappendix
\end{document}